\documentclass[10pt,twocolumn,letterpaper]{article}

\usepackage{iccv}
\usepackage{times}
\usepackage{epsfig}
\usepackage{graphicx}
\usepackage{amsmath}
\usepackage{amssymb}
\usepackage{multirow}
\usepackage[table,xcdraw]{xcolor}

\usepackage[pagebackref=true,breaklinks=true,letterpaper=true,colorlinks,bookmarks=false]{hyperref}

\usepackage{booktabs}

\usepackage[accsupp]{axessibility}

\iccvfinalcopy 

\def\iccvPaperID{2758} 

\begin{document}

\title{High-Resolution Document Shadow Removal via A Large-Scale Real-World Dataset and A Frequency-Aware Shadow Erasing Net}

\author{Zinuo Li\thanks{Equal contributions} \qquad Xuhang Chen\footnotemark[1] \qquad Chi-Man Pun\thanks{Corresponding authors} \qquad Xiaodong Cun\footnotemark[2]
\\
University of Macau     
}


\hypersetup{
    colorlinks = true,
    linkbordercolor = {white},
    citecolor=blue
}

\newcommand{\xiaodong}[1]{{\color{blue}{[xiaodong: #1]}}}

\maketitle
\ificcvfinal\thispagestyle{empty}\fi

\begin{abstract}
Shadows often occur when we capture the documents with casual equipment, which influences the visual quality and readability of the digital copies. Different from the algorithms for natural shadow removal, the algorithms in document shadow removal need to preserve the details of fonts and figures in high-resolution input. Previous works ignore this problem and remove the shadows via approximate attention and small datasets, which might not work in real-world situations. We handle high-resolution document shadow removal directly via a larger-scale real-world dataset and a carefully designed frequency-aware network. As for the dataset, we acquire over 7k couples of high-resolution~(2462 $\times$ 3699) images of real-world document pairs with various samples under different lighting circumstances, which is 10 times larger than existing datasets. As for the design of the network, we decouple the high-resolution images in the frequency domain, where the low-frequency details and high-frequency boundaries can be effectively learned via the carefully designed network structure. Powered by our network and dataset, the proposed method clearly shows a better performance than previous methods in terms of visual quality and numerical results. The code, models, and dataset are available at \url{https://github.com/CXH-Research/DocShadow-SD7K}.
\end{abstract}

\begin{figure}[t]
    \begin{minipage}[b]{1.0\linewidth}
        \begin{minipage}[b]{.32\linewidth}
            \centering
            \centerline{\includegraphics[width=\linewidth]{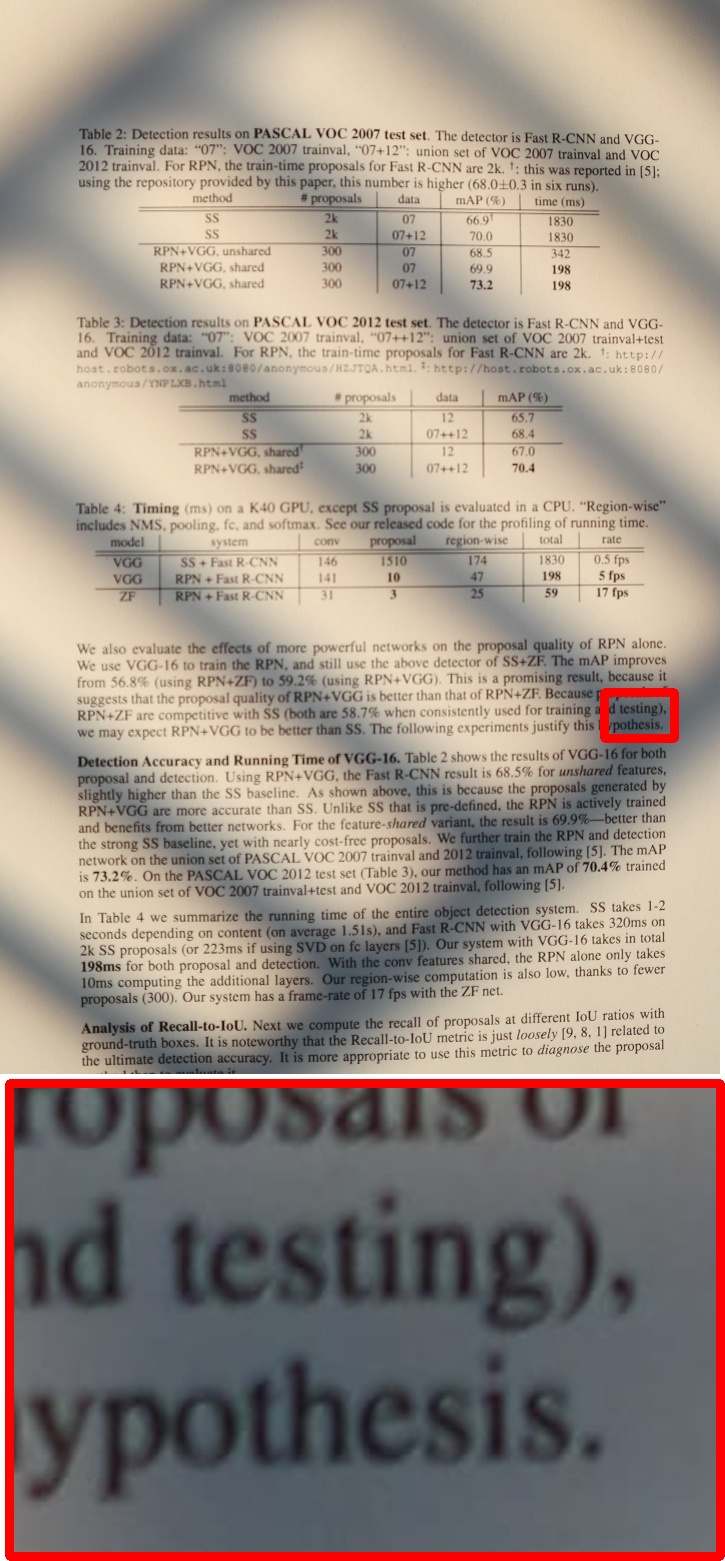}}
            \centerline{(a) Input}\medskip
        \end{minipage}
        \hfill
        \begin{minipage}[b]{.32\linewidth}
            \centering
            \centerline{\includegraphics[width=\linewidth]{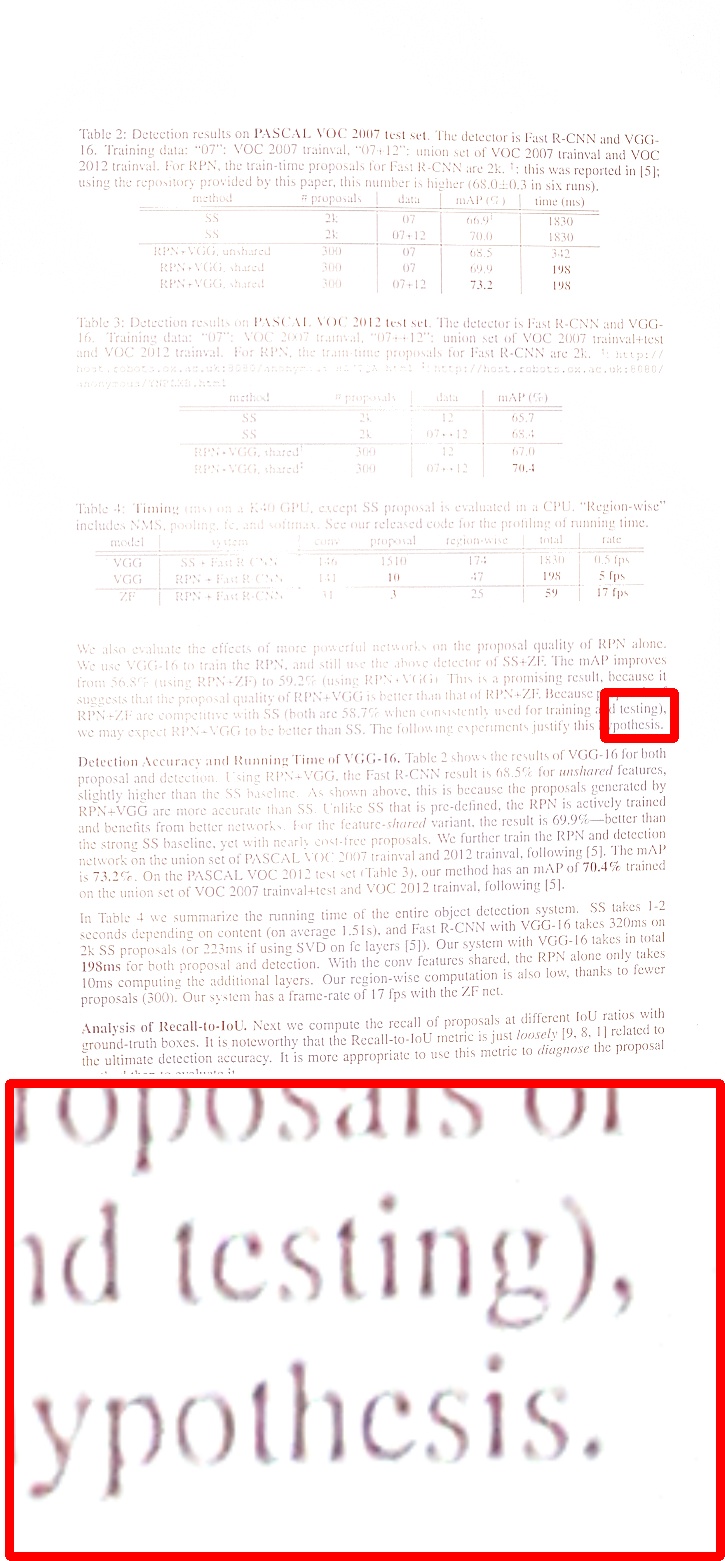}}
            \centerline{(b) Shah \etal~\cite{Shah2018AnIA}}\medskip
        \end{minipage}
        \hfill
        \begin{minipage}[b]{0.32\linewidth}
            \centering
            \centerline{\includegraphics[width=\linewidth]{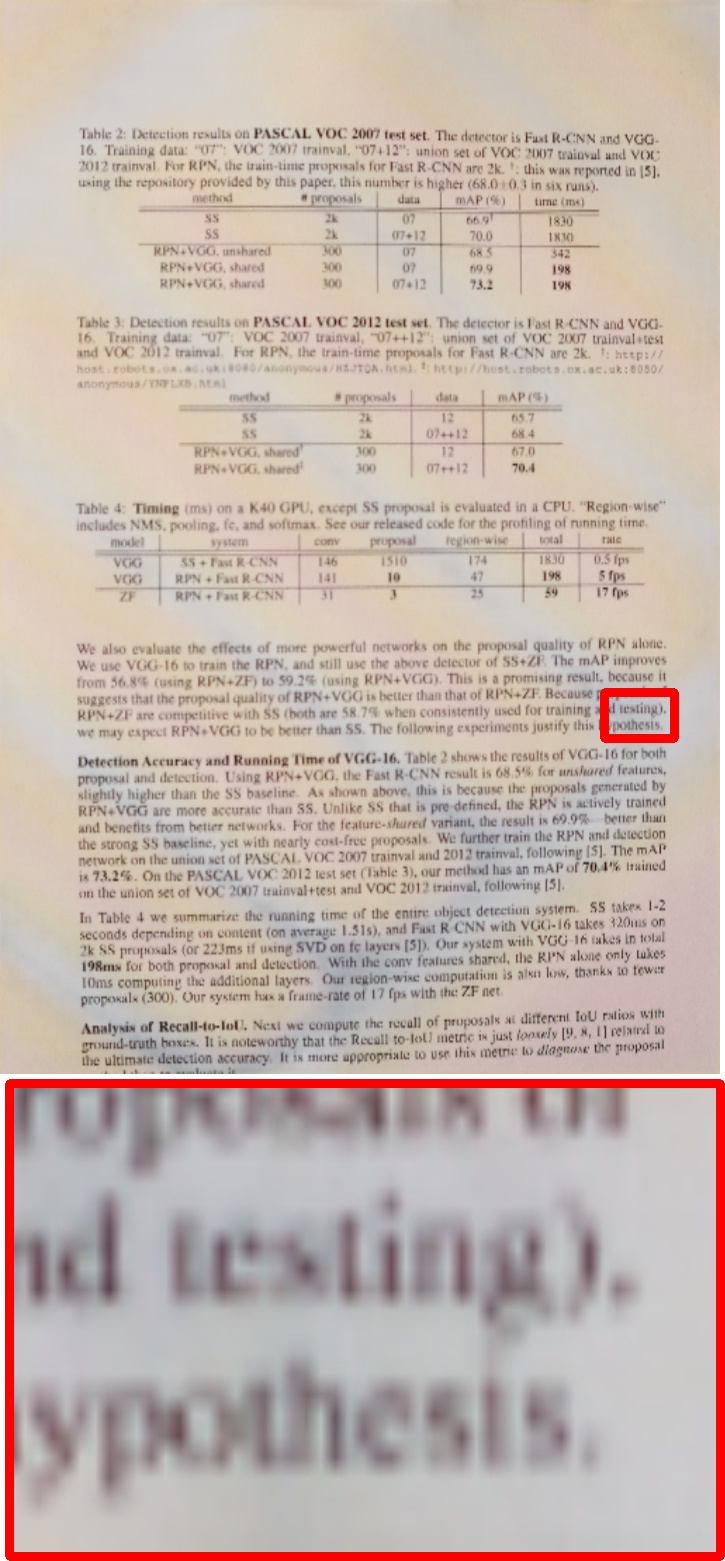}}
            \centerline{(c) Fu \etal~\cite{fu2021auto}}\medskip
        \end{minipage}
    \end{minipage}
    \begin{minipage}[b]{1.0\linewidth}
        \begin{minipage}[b]{.32\linewidth}
            \centering
            \centerline{\includegraphics[width=\linewidth]{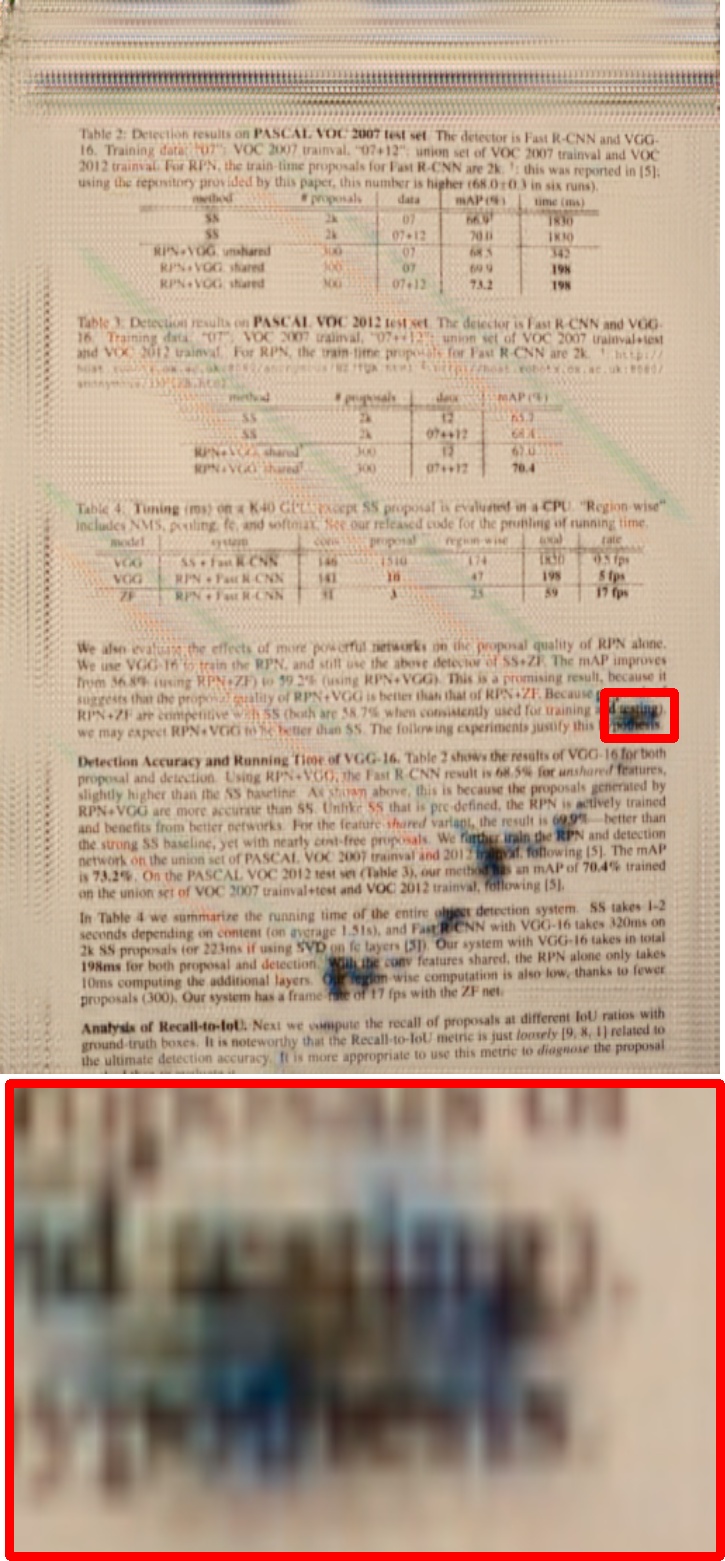}}
            \centerline{(d) Liu \etal~\cite{liu2021shadow}}\medskip
        \end{minipage}
        \hfill
        \begin{minipage}[b]{.32\linewidth}
            \centering
            \centerline{\includegraphics[width=\linewidth]{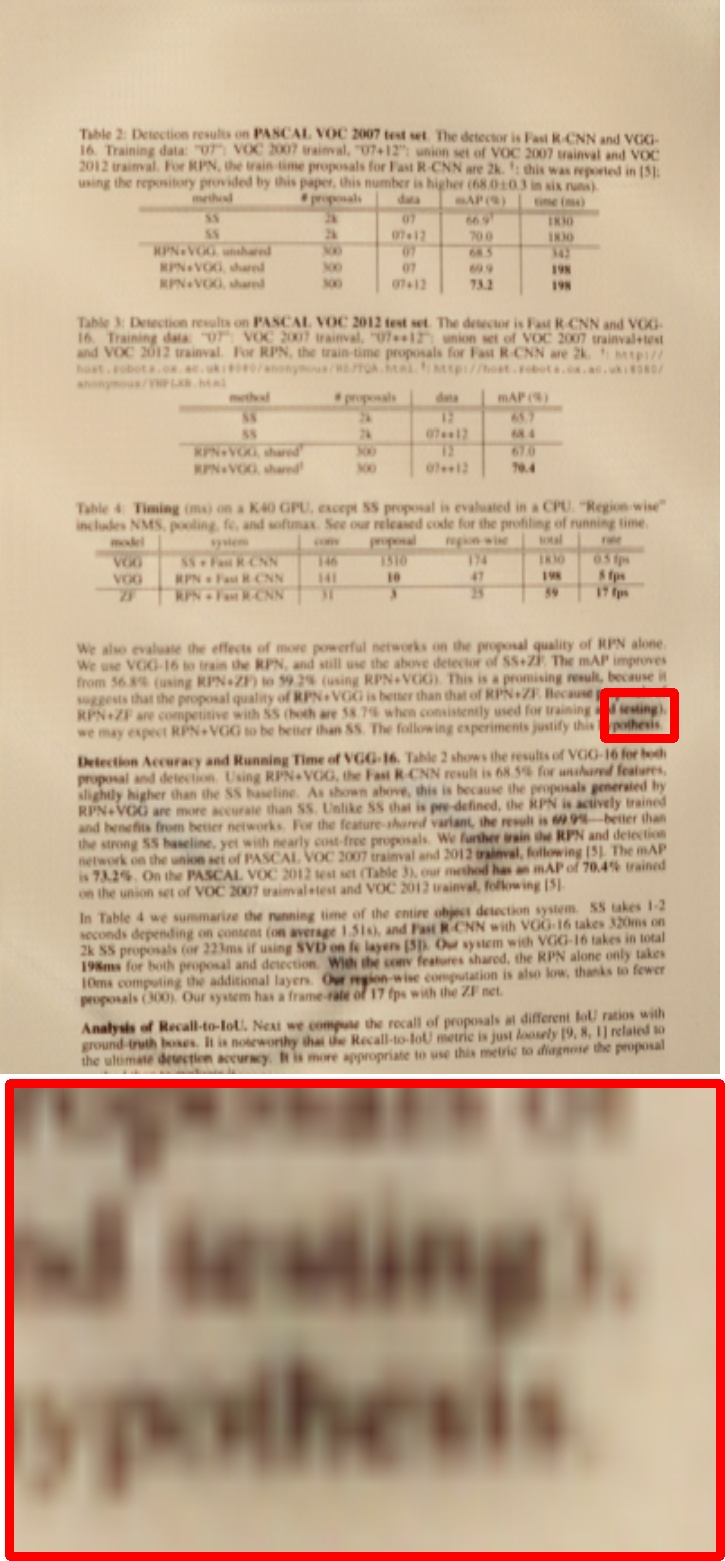}}
            \centerline{(e) Hu \etal~\cite{hu2019mask}}\medskip
        \end{minipage}
        \hfill
        \begin{minipage}[b]{0.32\linewidth}
            \centering
            \centerline{\includegraphics[width=\linewidth]{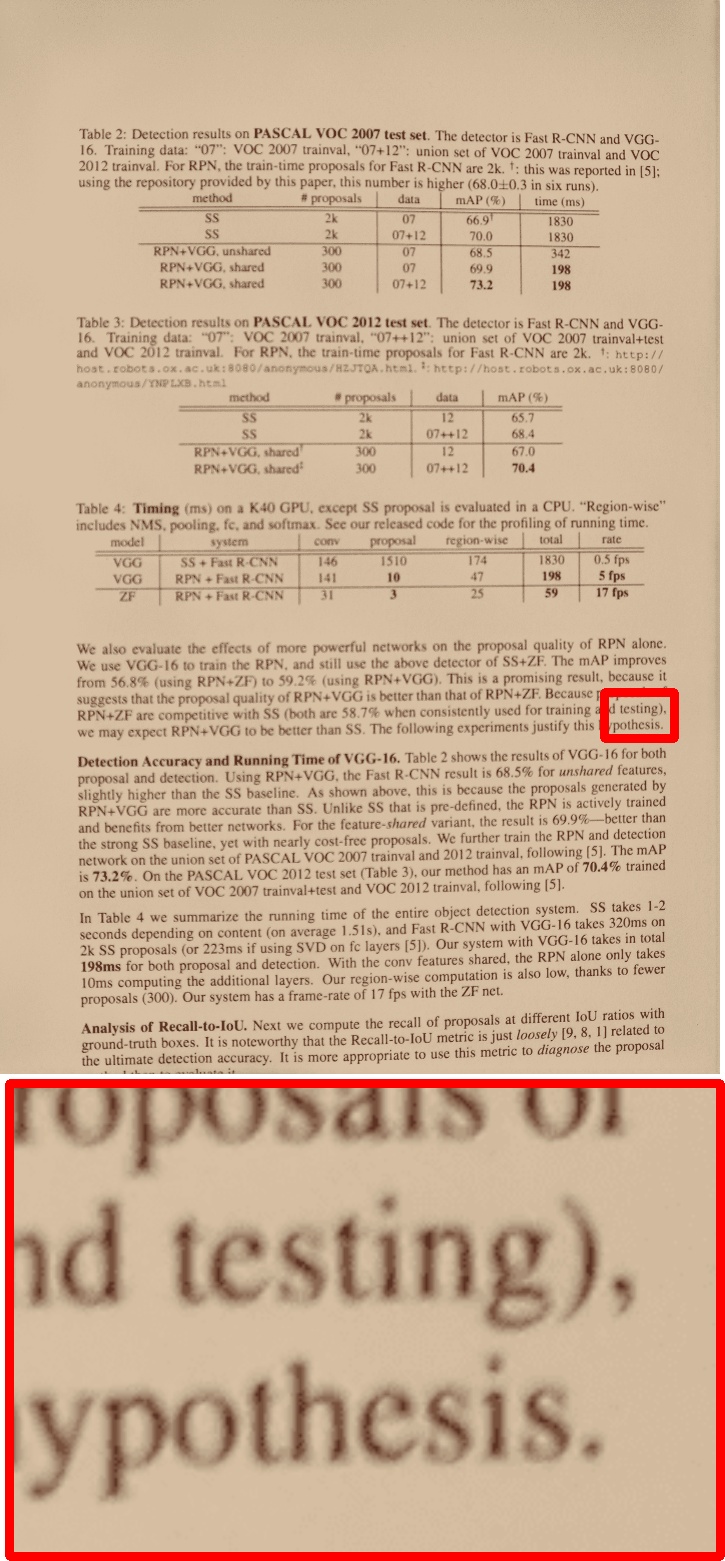}}
            \centerline{(f) Ours}\medskip
        \end{minipage}
    \end{minipage}
    \vspace{-1.5em}
    \caption{
    Visual results of input document shadow image~(a),  classic method~(b), supervised method~(c), weakly-supervised method~(d),  unsupervised method~(e) and ours~(f). Our model removes the shadow while preserving the original document's content and aspect ratio.
    } 
    \vspace{-1em}
    \label{fig:intro}
\end{figure}

\section{Introduction}

When occluders obstruct the light source, shadows are cast on the paper. These shadows reduce the readability and hinder subsequent text-related intelligent tasks such as document enhancement, optical character recognition~(OCR), key information extraction, and semantic entity labeling~\cite{kligler2018document,xu2020layoutlm,shi2016end}, \etc. 

Removing the cast shadow of the digital copy has a long need in the real world and also there are a lot of methods, including the traditional methods that applied physics-based illumination models~\cite{jung2018water, wang2019effective, Wang2020ShadowRO}. In reality, these approaches are significantly constrained due to the fact that the assumptions utilized in illumination models~\cite{Shah2018AnIA} are often too rigid to account for real-world shadow. Thus, they may incorrectly compute the exposure and reflectance as shown in Fig.~\ref{fig:intro}.

Recent methods apply deep learning for this task since it has achieved remarkable progress in natural shadow removal~\cite{cun2020towards} and other computer vision tasks~\cite{he2015spatial,he2016deep,wang2022uformer}. 
However, several obstacles make it challenging to transfer previous natural shadow removal methods to the document domain. 
On the one hand, in contrast to natural shadow removal, documents are presented by high-resolution images in the real world, which needs to preserve fine-grained information such as fonts and figures. 
However, the majority of document shadow removal methods~\cite{lin2020bedsr,chen2023shadocnet} are only designed for relatively low-resolution images and is hard to handle high-resolution situations directly since they are relying on the low-resolution approximation of attention for guidance. 
On the other hand, there are no large-scale and high-resolution datasets designed specifically for document shadow removal at this time. 
Consequently, given limited image pairs for training, existing methods usually result in different artifacts~(as shown in Fig.~\ref{fig:intro}~(c)). Although weakly-supervised~\cite{Le_2020_ECCV} and unsupervised~\cite{hu2019mask,liu2021shadow} shadow removal methods alleviate the requirements for data, they hold a strong assumption of the statistical similarity between shadowed and unshadowed domains. 
As illustrated in Fig.~\ref{fig:intro} (d) (e), once the training domain is very different from the test, these methods create hallucination contents and suffer from unstable results.




To solve the above problems, we first offer a large-scale real-world shadow dataset, Shadowed Document 7K~(SD7K), which contains more than 7k real-world and high-resolution shadow and shadow-free document image pairs with also the manually annotated shadow masks for related research, \eg, shadow detection~\cite{cun2020towards}. Compared to existing document shadow removal datasets, SD7K includes various document types, \eg, manga, papers, and figures. Also, different from previous work which only considers one specific light environment, we use three distinct light sources to increase the diversity of the training samples. 
Besides the dataset, we also propose a robust network for high-resolution image processing specifically, named Frequency-aware Shadow Erasing Net~(FSENet). In detail, inspired by the laplacian pyramid~\cite{burt1987laplacian,liang2021high}, we divide the document image into two frequency portions, each with its own domain-specific properties. The low-frequency component is primarily responsible for illuminations or colors while high-frequency components are more related to content details. As for the low-frequency deshading module, we involve the transformer~\cite{dosovitskiy2020image} based network to model the global illuminations alternation and a high-frequency restoration module based on cascaded aggregations for adaptive pixel enhancement. For high-frequency modules, we involve several dilated convolution-based texture recovery modules to restore the refined details similar to \cite{cun2020towards}.
With the aforementioned techniques, our model achieves the state-of-the-art performance of previous methods on two common small benchmarks and the proposed SD7K. 

We summarize the main contribution as follows:
\begin{itemize}
\item We provide SD7K, a large-scale real-world dataset consisting of high-resolution shadow and the associated shadow-free images under various illumination conditions.
\item We propose FSENet, a frequency-aware network with a carefully designed network structure to handle high-resolution document shadows.
\item Both qualitative and quantitative results on all available public datasets indicate that the proposed FSENet performs favorably against state-of-the-art methods.
\end{itemize}

\section{Related Work}

\noindent\textbf{Document Shadow Dataset.}
The collection of genuine shadow images and their corresponding shadow-free counterparts is a challenging task due to the varying lighting conditions. As a result, existing datasets are limited in size. The Bako dataset~\cite{bako2016removing} comprises outdated, inaccessible images in tiny quantities. Similarly, Jung's dataset~\cite{jung2018water} is limited in size and favors light shadows. Kligler's dataset~\cite{kligler2018document} emphasizes colorful symbols, which do not convey a significant amount of information. Among these datasets, the RDSRD dataset~\cite{lin2020bedsr}, while still modest in size, has a better quality in comparison. Different from the existing datasets, we propose a high-resolution and large-scale dataset SD7K that addresses the aforementioned problem, which can facilitate the research in document shadow removal.

\begin{table*}[t]
\centering
\resizebox{\textwidth}{!}{
\begin{tabular}{l|l|l|l|l|l|l}
\toprule
Dataset       & Triplets \#  & Entities \# & Mean Resolution & Document Type  & With Mask & Light Temperature Control \\ \hline
Bako~\cite{bako2016removing}       & 81 & N/A & N/A   & Text          & No & No      \\
Kligler~\cite{kligler2018document}      & 300 & 25 & $2424\times2013$  & Text           & No & No      \\
Jung~\cite{jung2018water}          & 87 & 87  & $602\times860$   & Text     & No & No     \\
RDSRD~\cite{lin2020bedsr}    &  540 & N/A  & N/A    & Text     & Yes & N/A 
\\ \hline
\textbf{SD7K} & 7620 & 350+ & $2462\times3699$ & Text and picture & Yes & Different lamps \& sunlight   \\ \bottomrule
\end{tabular}
}
\vspace{0.3em}
\caption{
Comparison of SD7K with other real-world datasets. The proposed dataset covers a large number of document shadow images and it is the only dataset that satisfies all important data features about document shadow currently. 
}
\label{datasetcomp}
\end{table*}

\noindent\textbf{Natural Shadow Removal.}
Methods for natural shadow removal might be informative and applicable to document images. On the one hand, GAN-based methods such as ST-CGAN~\cite{wang2018stacked}, use two stacked conditional GANs to jointly detect and eliminate shadows. Mask-ShadowGAN~\cite{hu2019mask} extend the paired GAN to unpaired setting via re-formulated cycle-consistency constraints. To fully utilize the dataset, DHAN~\cite{cun2020towards} synthesize pseudo shadow images and learned border artifacts-free images via a dual hierarchical aggregation network. SG-ShadowNet~\cite{wan2022style} studies the importance of normalization and learns the style representation of the non-shadow region to harmonize the shadow and shadow-free parts. 
On the other hand, CNN-based methods such as SP+M Net~\cite{le2020shadow} and SP+M+I Net~\cite{le2021physics} utilize the physical linear transformation model to enhance the shadow region and image decomposition to reconstruct the shadow-free image. AEF-Net~\cite{fu2021auto} predicts the exposure parameters of the shadowed regions and fused a series of overexposed shadow images with the original shadow image to restore the shadow region. ShadowFormer~\cite{guo2023shadowformer} is a transformer-based network to exploit the context correlation between shadow and non-shadow regions.

\noindent\textbf{Document Image Shadow Removal.}
Traditional document shadow removal methods usually demand prior knowledge regarding shadow and document context. This includes estimating the background color~\cite{wang2019effective, Wang2020ShadowRO} and modeling shading and lighting effects~\cite{jung2018water,Shah2018AnIA}. Deep learning-based methods are also designed to remove the shadows of documents. Currently, BEDSR-Net~\cite{lin2020bedsr} stands as the state-of-the-art model in document shadow removal. It predicts global background color and encodes shadow position as an attention map, allowing for the effective removal of shadows. However, as discussed in our introduction, it needs to predict the approximate attention maps from the feature map, which is not suitable for the high resolution inputs.

\section{SD7K Dataset}

\begin{figure}[t]
    \begin{minipage}[b]{1.0\linewidth}
        \begin{minipage}[b]{.50\linewidth}
            \centering
            \centerline{\includegraphics[width=\linewidth]{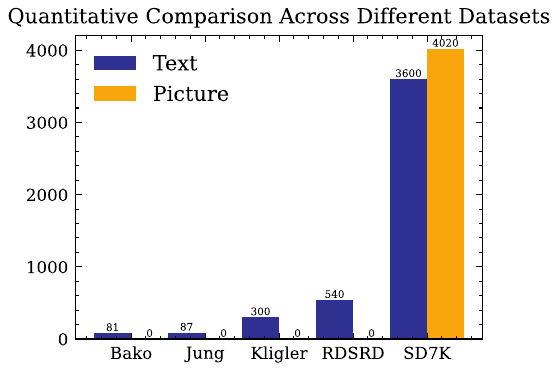}}
            \centerline{(a) Quantitative comparison}\smallskip
        \end{minipage}
        \hfill
        \begin{minipage}[b]{0.49\linewidth}
            \centering
            \centerline{\includegraphics[width=\linewidth]{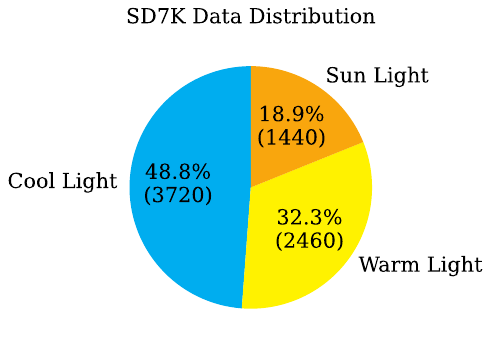}}
            \centerline{(b) Light distribution}\smallskip
        \end{minipage}
    \end{minipage}
    \caption{Data distribution of SD7K and quantitative comparison across all document shadow datasets.}
    \vspace{-1em}
    \label{fig:dataset}
\end{figure}

As shown in Tab.~\ref{datasetcomp} and Fig.~\ref{fig:dataset}~(a), there are only a limited number of datasets~\cite{jung2018water,lin2020bedsr,bako2016removing,kligler2018document} available for document shadow removal, which is typically small and comprises only hundreds of real-world samples. To train a network on such-scale datasets, synthesizing training samples via computer graphic engines is required~\cite{lin2020bedsr}. However, the rendered samples still show different distributions from the real samples, making it difficult in real-world applications.


To overcome this limitation, we introduce a massive real-world dataset named SD7K for document shadow removal in terms of the occluder entities, the numbers, the resolution, the light source, \etc. Below, we give the details of how to collect and process the dataset. 


\begin{figure}[ht]
    \begin{minipage}[b]{1.0\linewidth}
        \begin{minipage}[b]{.32\linewidth}
            \centering
            \centerline{\includegraphics[width=\linewidth, height=1.5\linewidth]{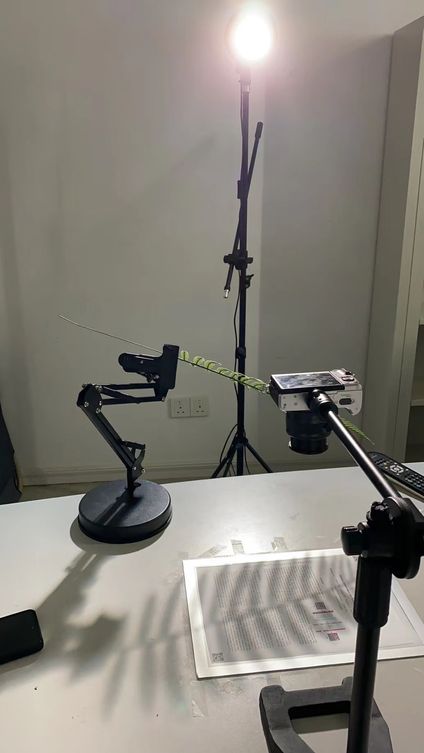}}
            \centerline{(a)}\medskip
        \end{minipage}
        \hfill
        \begin{minipage}[b]{0.32\linewidth}
            \centering
            \centerline{\includegraphics[width=\linewidth, height=1.5\linewidth]{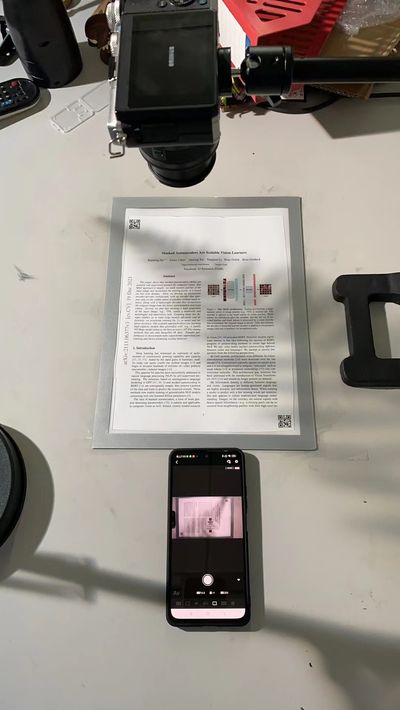}}
            \centerline{(b)}\medskip
        \end{minipage}
        \hfill
        \begin{minipage}[b]{0.32\linewidth}
            \centering
            \centerline{\includegraphics[width=\linewidth, height=1.5\linewidth]{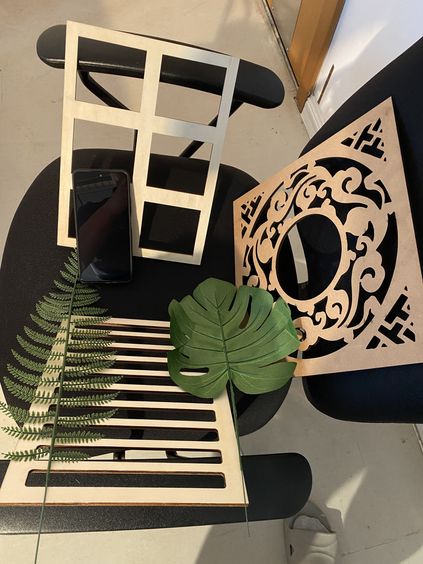}}
            \centerline{(c)}\medskip
        \end{minipage}
    \end{minipage}
    \vspace{-1.5em}
    \caption{(a) our data acquisition setup for constructing the dataset. (b) a remote phone control shutter where the user only needs to click a button to capture a pair of shadow-free/shadow images. (c) some showcases of our occluders.}
    \vspace{-0.5em}
    \label{camera}
\end{figure}


First, we use a Canon EOS M6 camera to capture printed document images from real-world scenarios. To capture paired shadow and shadow-free images, we make all the light close and leave one light source. All the settings of the camera are also fixed, including focal length, aperture, exposure, and ISO, to avoid the influence of the camera algorithm. 
Then, we use a camera stand to hold the camera and make it shoot on the paper vertically as shown in Fig.~\ref{camera}~(a). To capture the paired shadow and shadow-free image without involving the changes caused by the manual presses, as in Fig.~\ref{camera}~(b), we use a remote control to capture the photo between the shadowed and shadow-free paper by removing the occluder. We also take the pairs as short as possible to avoid the light difference between captures.

To further enhance diversity, we alternate the positions of the document and occluder after capturing a few batches of documents. 
Overall, we use over 30 types of occluders along with more than 350 documents to contribute to the dataset, some of the occluders are shown in Fig.~\ref{camera}~(c). These occluders have the shape of both regular and irregular forms, which provides adequate coverage for various situations, we calculate the shadow area percentage of the whole dataset, which is $41.38\% \pm 13.58\%$, indicating that it is sufficiently divergent for our shadow removal task.

We also consider the fluctuations in sunlight intensity and random placement of shadows, which will be significantly influenced by the light variance. To simulate the results in different light sources, we introduce three types of light sources in our dataset for collection, including regular sunlight, a stationary 80W LED yellow light having a color temperature of 3000K, and a stationary 80W LED white light having a color temperature of 6000K. The light distribution of the whole dataset is shown in Fig.~\ref{fig:dataset}~(b).

\begin{figure}[t]
    \begin{minipage}[b]{1.0\linewidth}
        \begin{minipage}[b]{.32\linewidth}
            \centering
            \centerline{\includegraphics[width=\linewidth, height=1.5\linewidth]
            {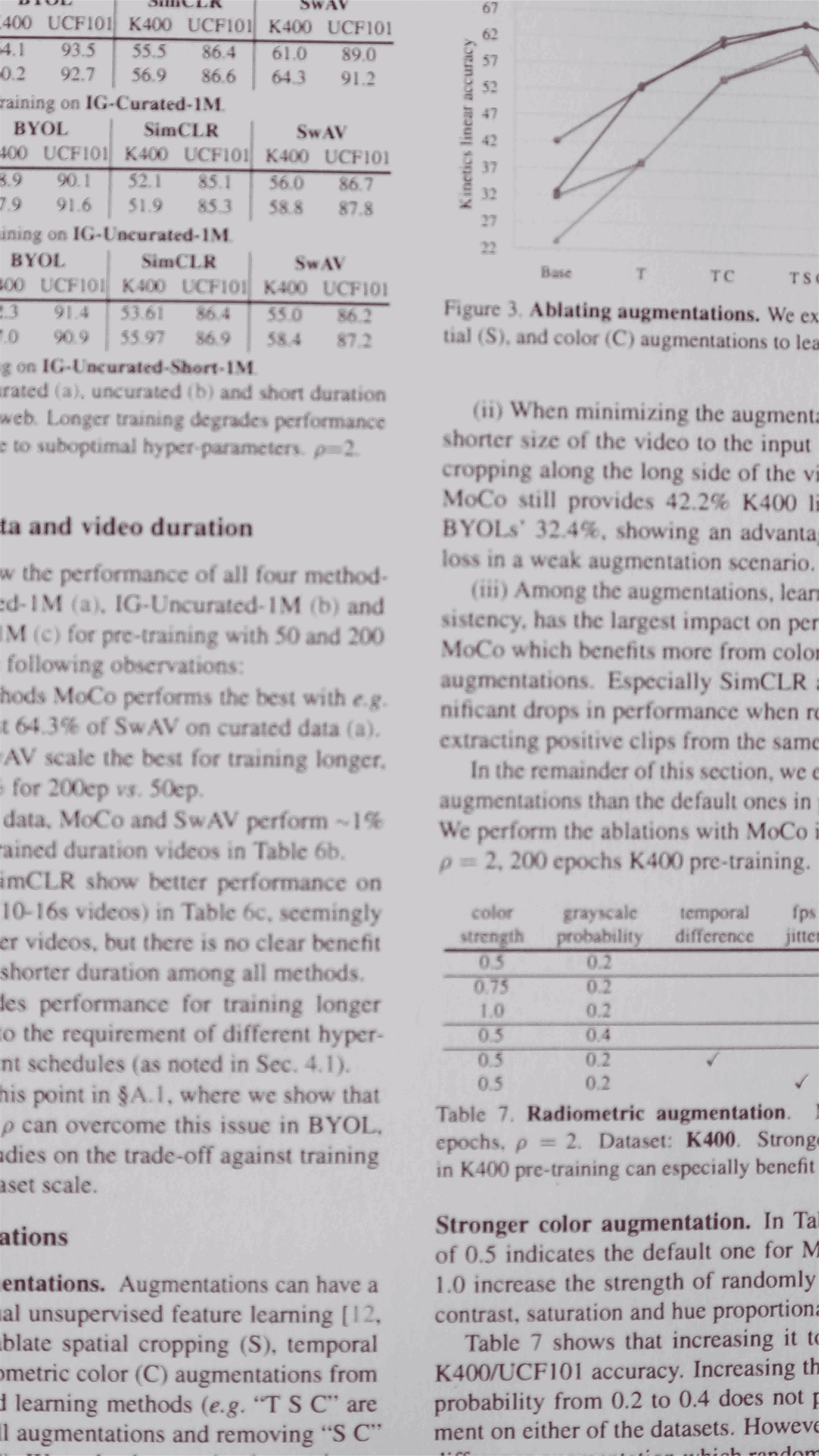}}
        \end{minipage}
        \hfill
        \begin{minipage}[b]{0.32\linewidth}
            \centering
            \centerline{\includegraphics[width=\linewidth, height=1.5\linewidth]
            {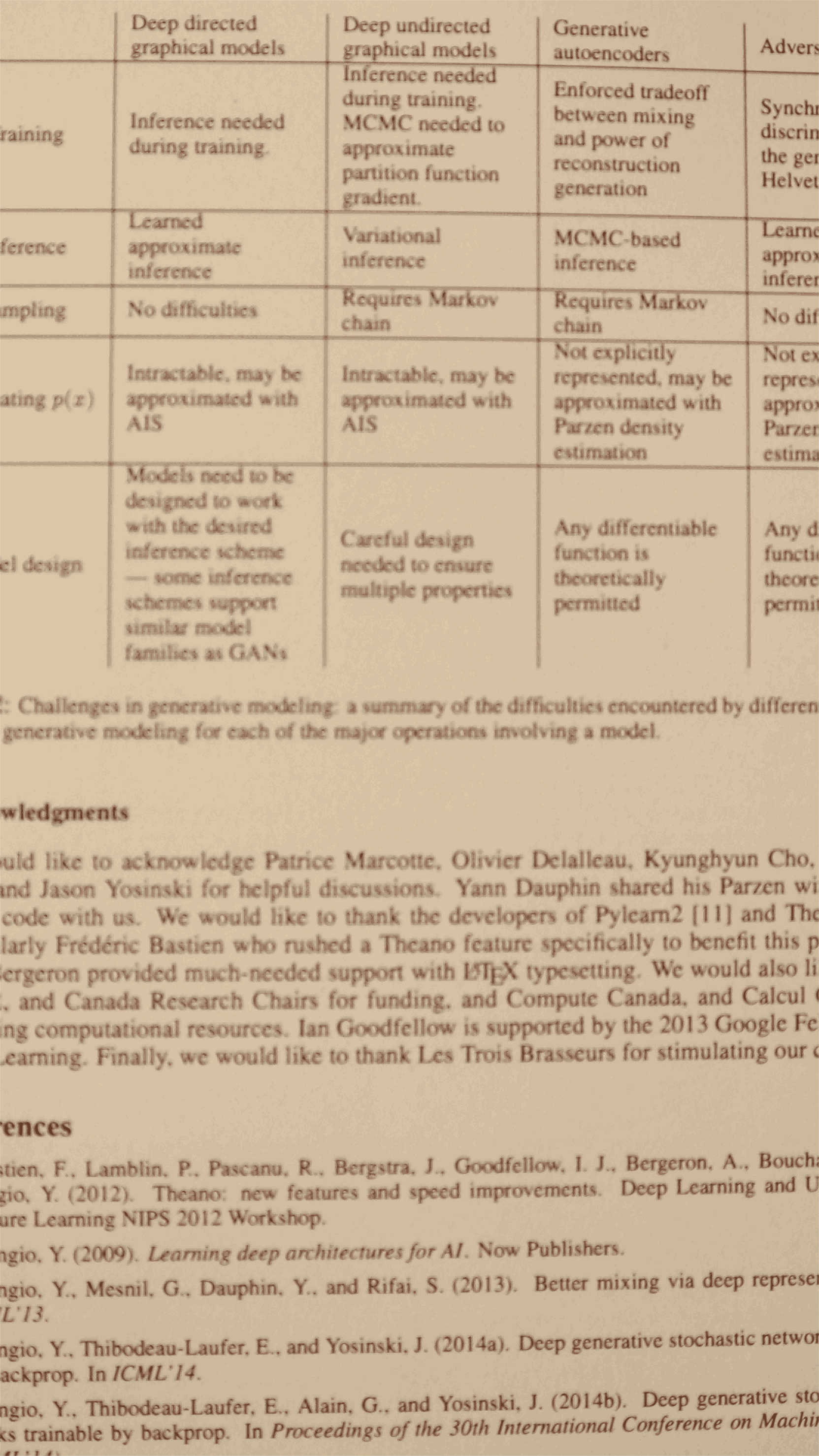}}
        \end{minipage}
        \hfill
        \begin{minipage}[b]{0.32\linewidth}
            \centering
            \centerline{\includegraphics[width=\linewidth, height=1.5\linewidth]
            {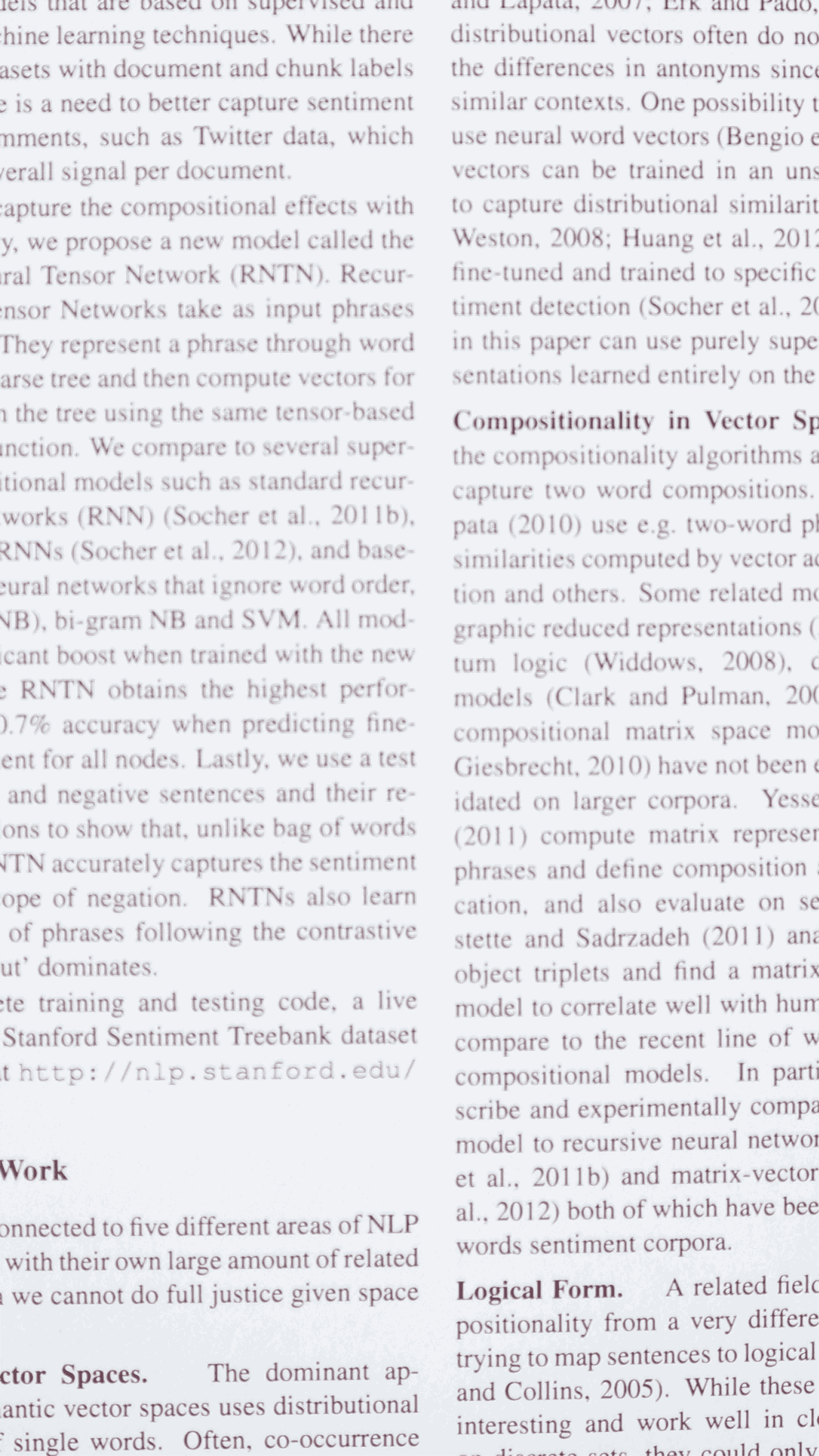}}
        \end{minipage}
    \end{minipage}
    
    \vspace{2.5pt}
    
    \begin{minipage}[b]{1.0\linewidth}
        \begin{minipage}[b]{.32\linewidth}
            \centering
            \centerline{\includegraphics[width=\linewidth, height=1.5\linewidth]
            {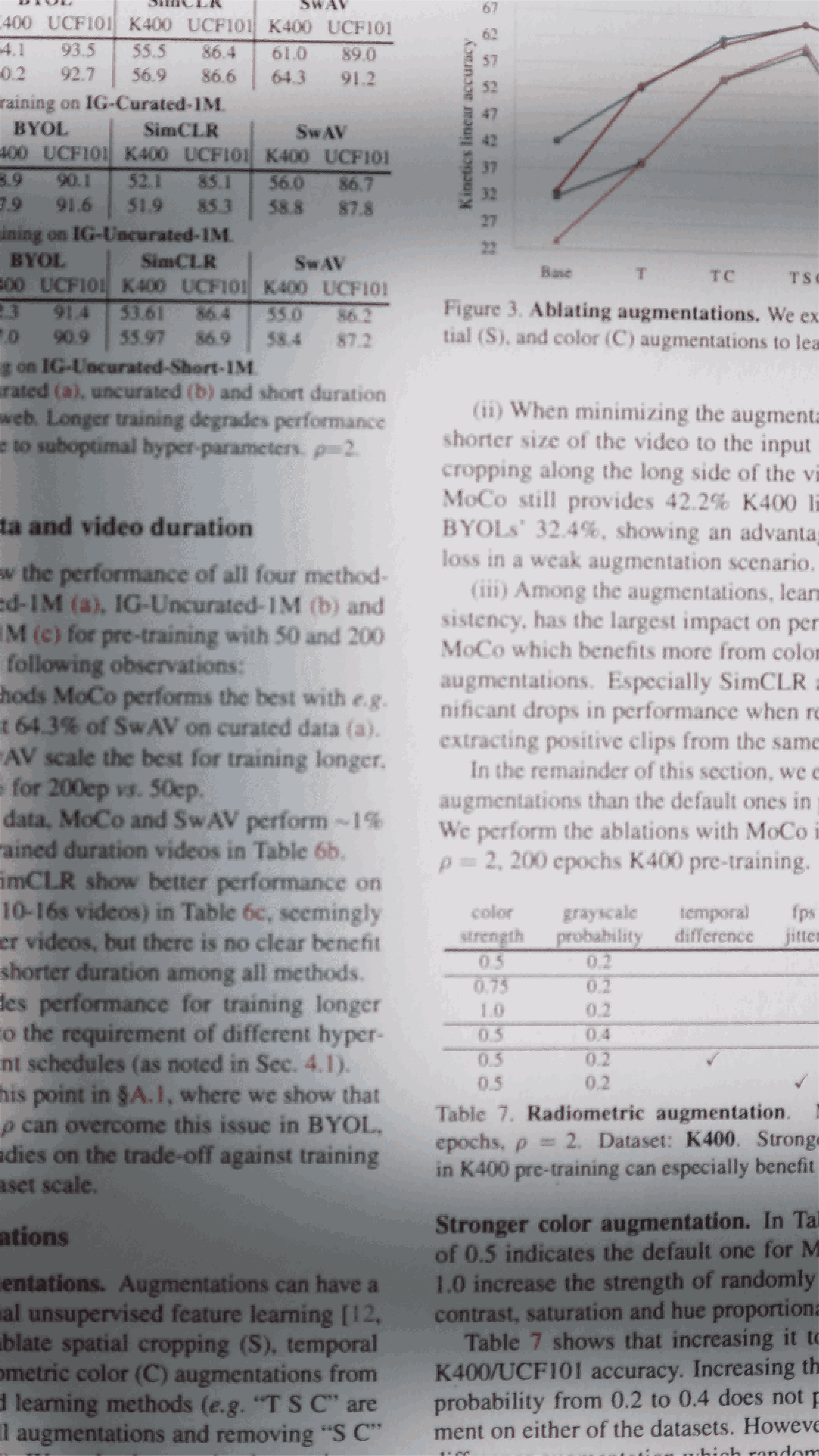}}
            \centerline{Cool light}\medskip
        \end{minipage}
        \hfill
        \begin{minipage}[b]{0.32\linewidth}
            \centering
            \centerline{\includegraphics[width=\linewidth, height=1.5\linewidth]
            {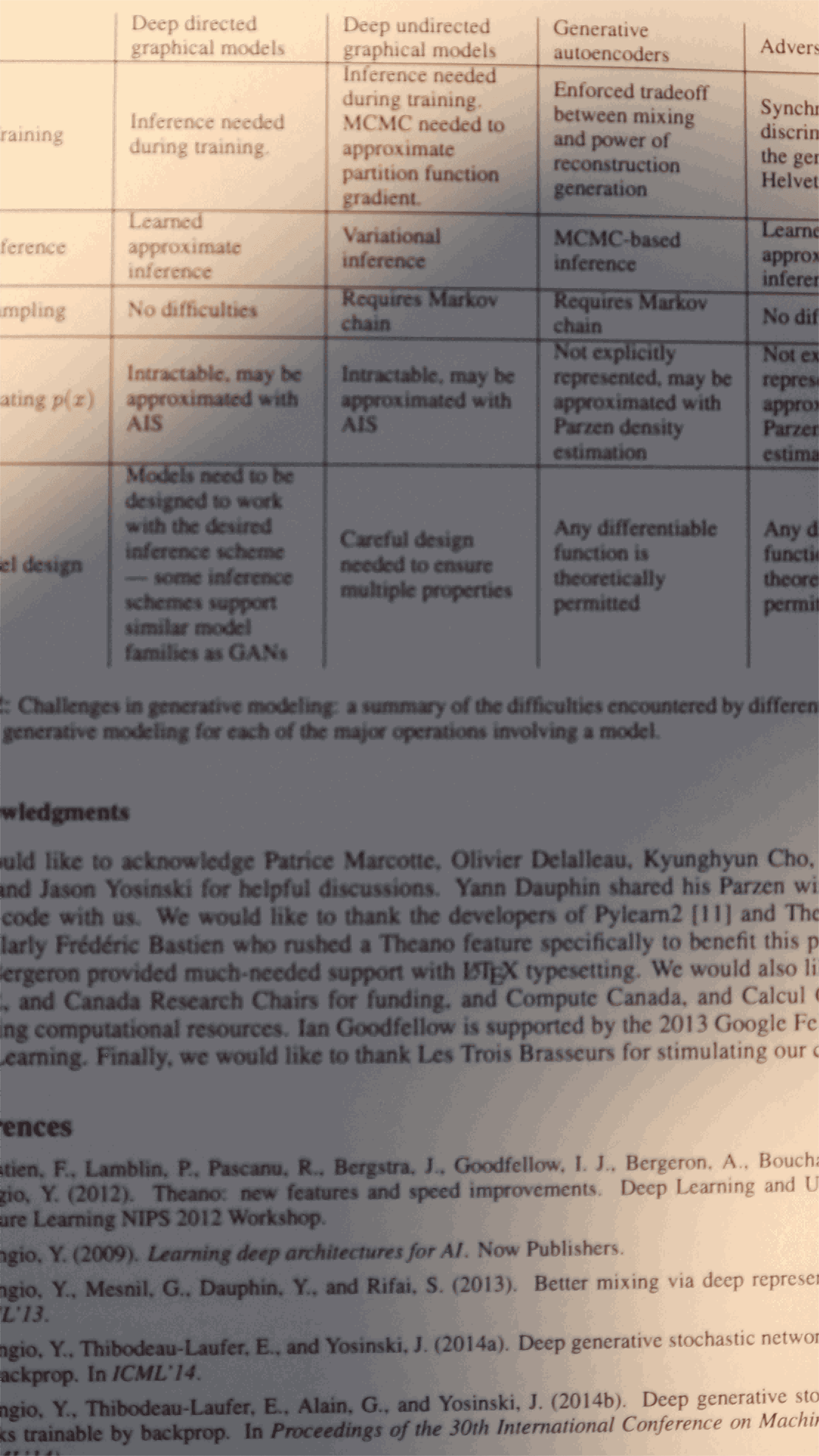}}
            \centerline{Warm light}\medskip
        \end{minipage}
        \hfill
        \begin{minipage}[b]{0.32\linewidth}
            \centering
            \centerline{\includegraphics[width=\linewidth, height=1.5\linewidth]
            {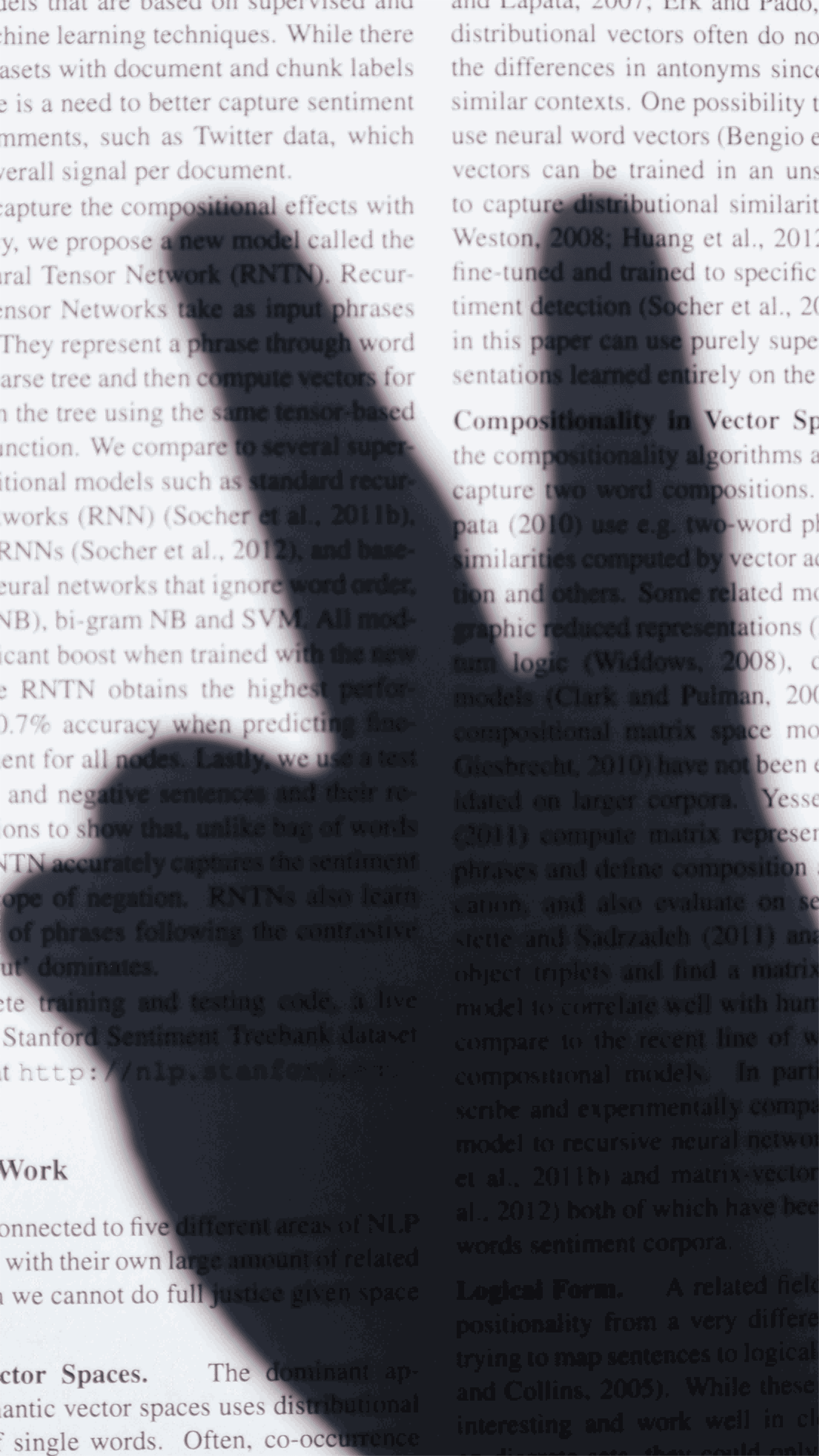}}
            \centerline{Sunlight}\medskip
        \end{minipage}
    \end{minipage}
    \vspace{-1.5em}
    \caption{Example shadow and shadow-free images from SD7K.}
    \vspace{-1em}
    \label{Act-Shadow}
\end{figure}

For data cleaning purposes, we first remove images with noticeable motion distortion or lack of focus. Since some photo still contains outliers and the context is duplicated, we then review the photographs in groups and remove them. Finally, we crop the areas outside of the margins of the document, resulting in a total of 7620 samples, where some of the samples are shown in Fig.~\ref{Act-Shadow}. We also use a threshold-based method to extract the mask of the training pair since the shadow mask will also benefit related tasks.

\section{Frequency-Aware Shadow Erasing Net}
Since real-world document shadow removal needs huge GPU memory resources, therefore our network initially divides the input image into different frequency components using the Laplacian Pyramid~(LP) similar to \cite{liang2021high} to reduce the computational budgets. As shown in Fig.~\ref{fig:method}, we propose a powerful frequency-aware shadow removal network named FSENet by designing different modules to handle the high-frequency and low-frequency computation individually.  After training, the multi-frequency components are mixed to produce the final results. In the rest of this section, we first give the details of the Laplacian pyramid in Sec.~\ref{sec:lp}. Then, in Sec.~\ref{sec:low} and Sec.~\ref{sec:high}, we give the details of the low-frequency deshading module and high-frequency restoration module to precisely modify the color details and the high-frequency details individually. Finally, we give the loss functions in Sec.~\ref{sec:loss}.

\subsection{Laplacian Pyramid and Method Overview}
\label{sec:lp}
Laplacian Pyramid~(LP)~\cite{burt1987laplacian} plays an important role in image processing. By decomposing and restoring images in a lossless manner, LP provides a rapid processing speed and is well-suited for dealing with high-resolution images. Inspired by \cite{liang2021high} for image translation, we implement LP for document shadow removal and train the network on multi-frequency.

\begin{figure*}[ht]
    \begin{minipage}[b]{1.0\linewidth}
        \includegraphics[width=\linewidth]{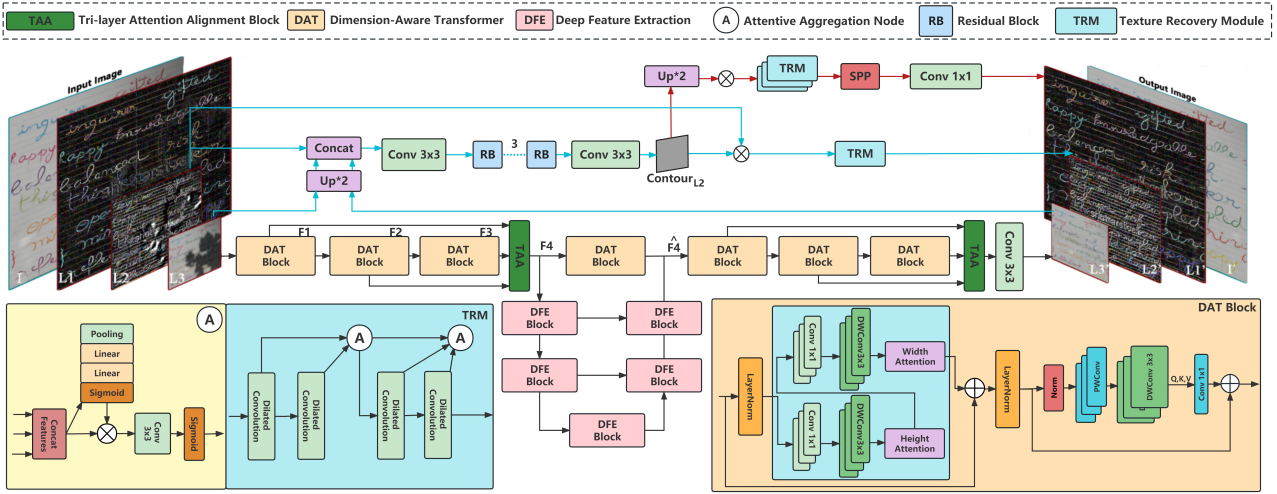}
    \end{minipage}
    \vspace{-1em}
    \caption{
    The network structure of our proposed FSENet. Following \cite{liang2021high}, given a high-resolution image $I\in \mathbb{R}^{H\times W\times 3}$, we first use Laplacian Pyramid~(\eg Depth~$D$ = 2 in this case) to decompose the images to multiple frequency components. The \textbf{Black} arrows: The low-frequency part $L_3\in \mathbb{R}^{\frac{H}{2^D}\times \frac{W}{2^D}\times C}$ are refined to $L_3^{'}\in \mathbb{R}^{\frac{H}{2^D}\times \frac{W}{2^D}\times C}$ utilizing DAT, TAA and DFE block. \textcolor{cyan}{\textbf{Cyan}} arrows: For the high-frequency part $L_2\in \mathbb{R}^{\frac{H}{2^{D-1}}\times \frac{W}{2^{D-1}}\times C}$, a contour $C_{L_2}\in \mathbb{R}^{\frac{H}{2^{D-1}}\times \frac{W}{2^{D-1}}\times 1}$ is learned to bridge the low-frequency and high-frequency features. \textcolor{red}{\textbf{Red}} arrows: For the remaining components with higher frequency, the learned contour is successively upsampled and refined using the proposed SPP and TRM. 
    }
    \label{fig:method}
\end{figure*}

In detail, given an image $I \in \mathbb{R}^{{H} \times {W} \times 3}$, the $D$ layers LP contains several high-frequency components $[L_0, ..., L_{D-1}]$ and a low-frequency component $L_D \in \mathbb{R}^{\frac{H}{2^D} \times \frac{W}{2^D} \times 3}$, where $L_D$ is $D$ times downsample of $I$. Each high-frequency component can be described as $L_{i}=I_{i}-\operatorname{PyrUp}\left(\operatorname{PyrDown}\left(I_{i}\right)\right)$, where $I_{i}$ is the $i$-th down-sampled image and PyrUp and PyrDown stand for upsampling and downsampling, respectively. 
As illustrated in Fig.~\ref{fig:method}, the $L_{1}$ and $L_{2}$ include the high-frequency component edges and textures, while the low-frequency component $ L_{3} = I_{3}$ majorly contains the color information.
For reversible reconstruction, the final $\hat{I}$ can be obtained by reversed update the equation of $ \hat{I}_{D-1} = \operatorname{PyrUp}(\hat{I}_D) + \hat{L}_{D-1} $ from $D$ to 0, where $\hat{I}_D$ is the learned low-frequency features and $\hat{L}_{D-1}$ is the learned high-frequency features from the proposed network. We give detailed instructions for each network below.

\subsection{Low-Frequency Deshading Module}
\label{sec:low}

After getting the low-frequency components, we use several transformer-based blocks to reduce the color and illumination distortions of the shadow images.
Compared to CNNs, transformers are superior at modeling non-local self-similarity and long-range dependencies. As shown in Fig.~\ref{fig:method}, we implement a transformer-based module called Dimension-Aware Transformer~(DAT) inspired by~\cite{wang2022ultra} with a lightweight UNet-like sub-network~\cite{chen2022simple} for low-frequency feature modeling.

Formally, given a low-frequency component $L_3 \in \mathbb{R}^{\frac{H}{4} \times \frac{W}{4} \times 3}$ of input image $I \in \mathbb{R}^{H \times W \times 3}$, we first feed $L_3$ into three sequential DAT blocks to extract rich features. Each DAT block is built via a sequential height-attention and width-attention to learn the feature in different aspects individually and a convolution-based layer for local feature merging, where we denote the intermediate features of each DAT block as ${F1,F2,F3}\in \mathbb{R}^{\frac{H}{4} \times \frac{W}{4} \times C }$. Please refer to the original \cite{wang2022ultra} for further details of the height-attention and width-attentions.

We then feed these features $F1, F2, F3$ into a Tri-layer Attention Alignment~(TAA) block for feature remixing. Given concatenated features $F_{\text {in}}\in \mathbb{R}^{N \times \frac{H}{4}\times \frac{W}{4} \times C} = Concat([F1, F2, F3])$, where N = 3 in this case, $F_{\text {in}}$ is first reshaped to $\hat{{F}}_{\text {in}}\in \mathbb{R}^{\frac{H}{4}\times \frac{W}{4}\times 3C}$. Then, we send $\hat{{F}_{\text {in}}}$ to a $1 \times 1$ convolution layer to aggregate pixel-level cross-channel context, followed by three $3 \times 3$ depth-wise convolution layers to generate query $\mathbf{Q}$, key $\mathbf{K}$, and value $\mathbf{V}$, individually. The $\mathbf{Q}, \mathbf{K}$ are then reshaped to $\hat{\mathbf{Q}}\in \mathbb{R}^{N \times \frac{H}{4}\frac{W}{4}C}$ and $\hat{\mathbf{K}}\in \mathbb{R}^{\frac{H}{4}\frac{W}{4}C \times N}$ respectively to calculate an attention matrix $\mathbf{A \in \mathbb{R}^{N \times N}}$. Next, we reshape the $\mathbf{V}$ to $\hat{\mathbf{V}}\in \mathbb{R}^{\frac{H}{4}\frac{W}{4}C \times N}$ to multiply $\hat{\mathbf{V}}$ and $\mathbf{A}$, and the input feature $\hat{{F}}_{\text {in}}$ is added to generate the output feature $F4$, which can be described as:
\begin{equation}
    F4=Conv_{1 \times 1}LA
(\hat{\mathbf{Q}}, \hat{\mathbf{K}}, \hat{\mathbf{V}})+\hat{F}_{\text {in}},
\label{eq:fout}
\end{equation}
where $Conv_{1 \times 1}$ represents a convolution layer with the kernel of $1 \times 1$. LA indicates Layer Attention~\cite{wang2022ultra}, which can be written as:
\begin{equation}
    LA(\hat{\mathbf{Q}}, \hat{\mathbf{K}}, \hat{\mathbf{V}})=\hat{\mathbf{V}} \operatorname{softmax}(\hat{\mathbf{Q}} \hat{\mathbf{K}} / \alpha),
\label{eq:la}
\end{equation}
where $\alpha$ is the temperature of the attention. Please refer to the supplementary material for more details of the TAA.

After getting $F4$, as shown in Fig.~\ref{fig:method}, we also use a UNet-like~\cite{chen2022simple, wang2022uformer} sub-network built by Deeper Feature Extraction~(DFE) blocks to learn multi-scale features. Each DFE block begins with a LayerNorm to stabilize the training procedure, followed by two convolutions. Then, it utilizes SimpleGate~\cite{chen2022simple} to increase the diversity of the features by dividing the features along the channel dimension into two parts and multiplying them together.
We also utilize Simplified Channel Attention~(SCA) similar to \cite{chen2022simple} to capture global information in a computationally efficient manner, which can be denoted as:
\begin{equation}
SCA(\mathbf{X})=\mathbf{X} * \mathbf{W} \operatorname{pool}(\mathbf{X}),
\label{eq:sca}
\end{equation}
where $\mathbf{W}$ represents a fully-connected layer, $\operatorname{pool}$ represents the global average pooling procedure that combines spatial data into channels, and $*$ represents a channel-wise multiplication. The skip-connections~\cite{he2016deep} are connected before the LayerNorm to the output similar to \cite{chen2022simple}. We also give a detailed figure of the DFE block in the supplementary materials.

For the refined feature $\hat{F4}$, it then passes through three transformer blocks and a TAA block to produce the enhanced features for image reconstruction using DAT and TAA blocks similar to the feature extractions of $F1, F2, F3$. Finally, a convolution layer with the kernel size $3 \times 3$ is performed on the enhanced features to produce the low-frequency component $L_3^{'} \in \mathbb{R}^{\frac{H}{4} \times \frac{W}{4} \times C}$.

\begin{figure*}[ht]
    \begin{minipage}[b]{1.0\linewidth}
        \begin{minipage}[b]{.19\linewidth}
            \centering
            \centerline{\includegraphics[width=\linewidth]{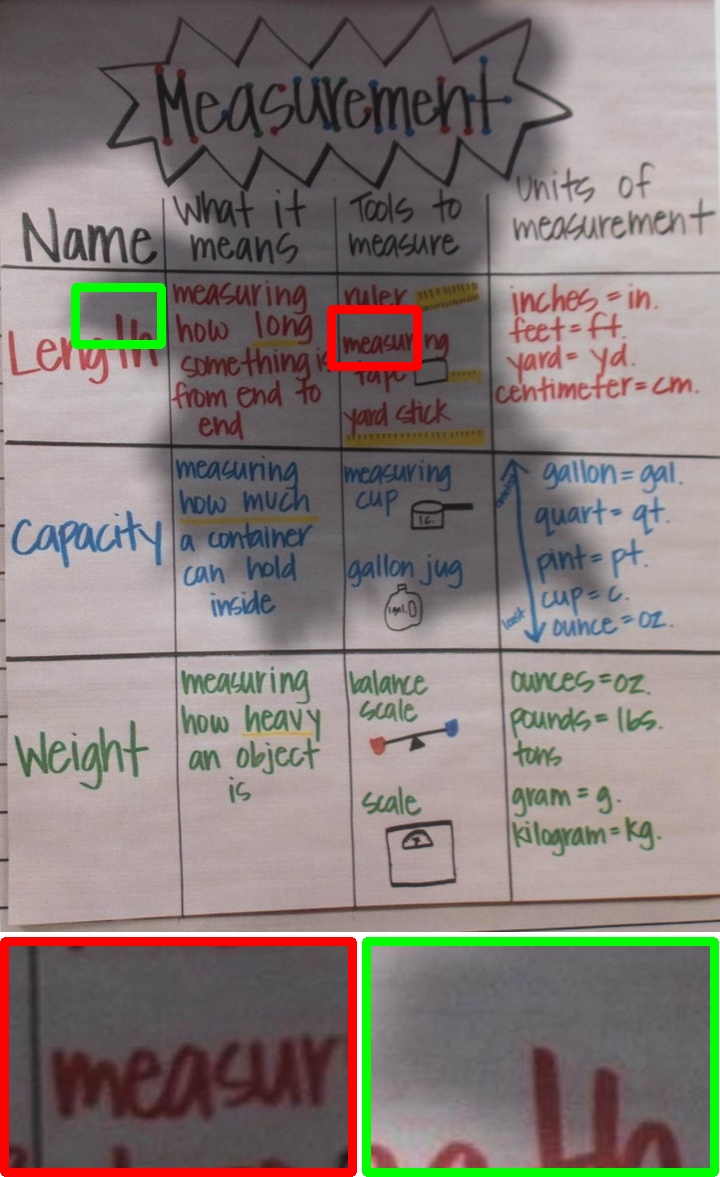}}
        \end{minipage}
        \hfill
        \begin{minipage}[b]{.19\linewidth}
            \centering
            \centerline{\includegraphics[width=\linewidth]{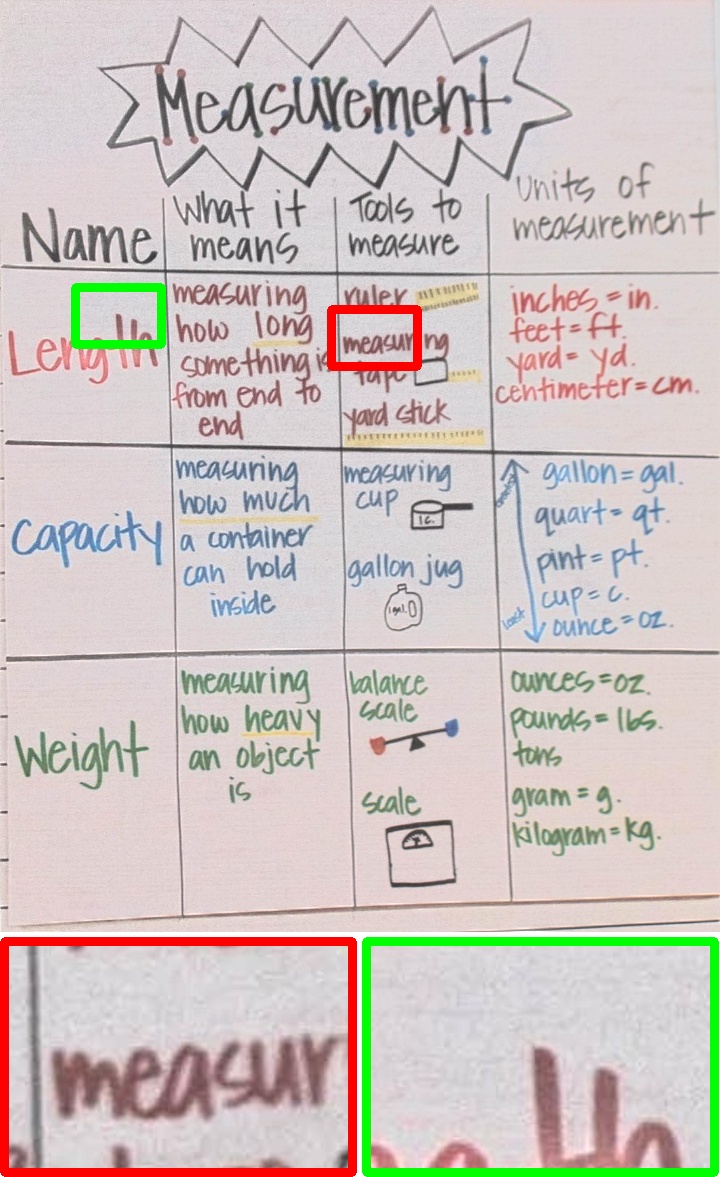
            }}
        \end{minipage}
        \hfill
        \begin{minipage}[b]{.19\linewidth}
            \centering
            \centerline{\includegraphics[width=\linewidth]{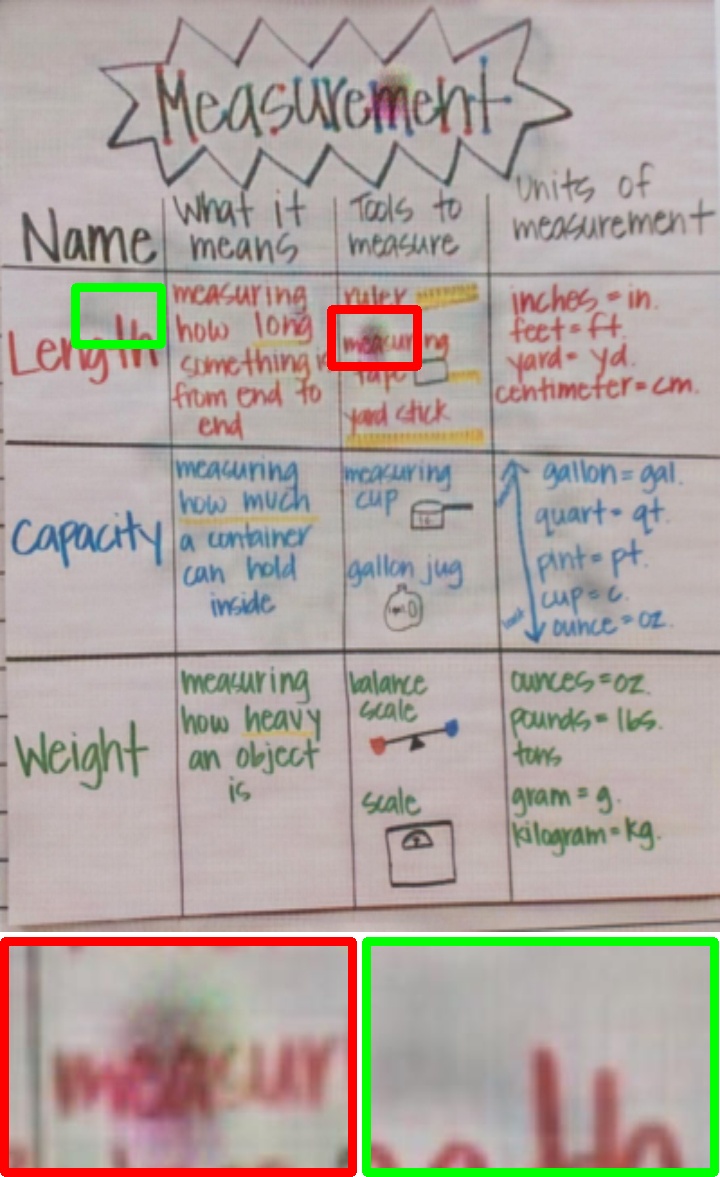}}
        \end{minipage}
        \hfill
        \begin{minipage}[b]{.19\linewidth}
            \centering
            \centerline{\includegraphics[width=\linewidth]{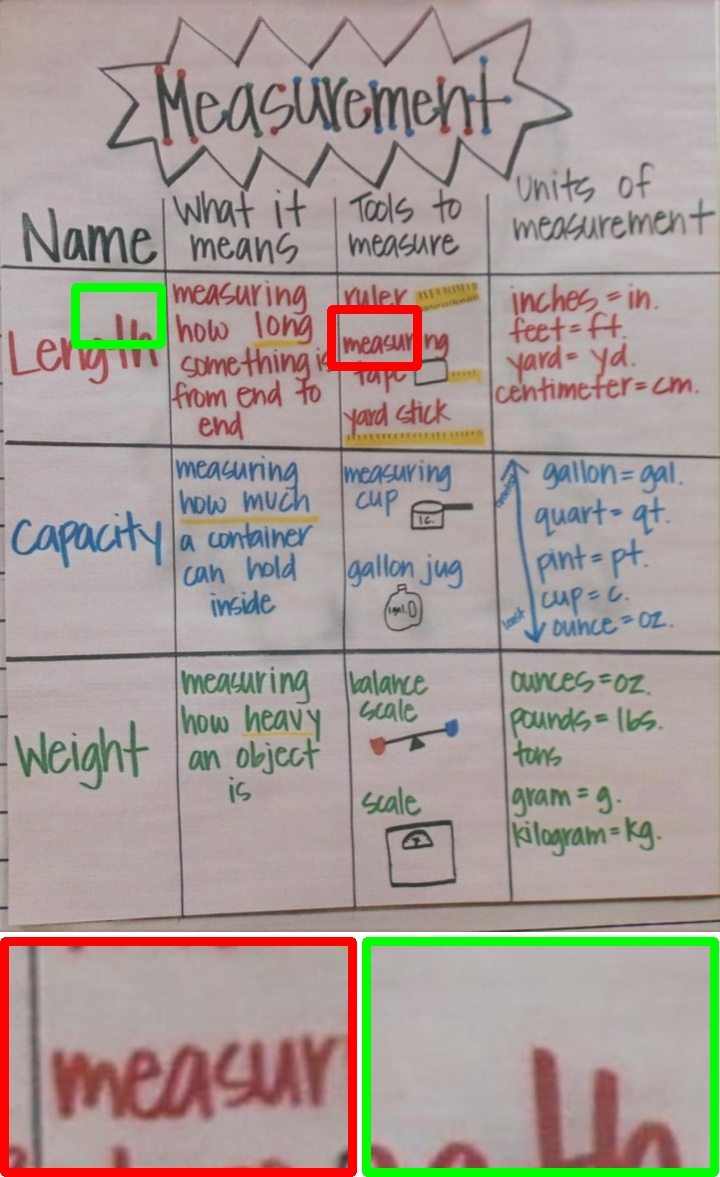}}
        \end{minipage}
        \hfill
        \begin{minipage}[b]{.19\linewidth}
            \centering
            \centerline{\includegraphics[width=\linewidth]{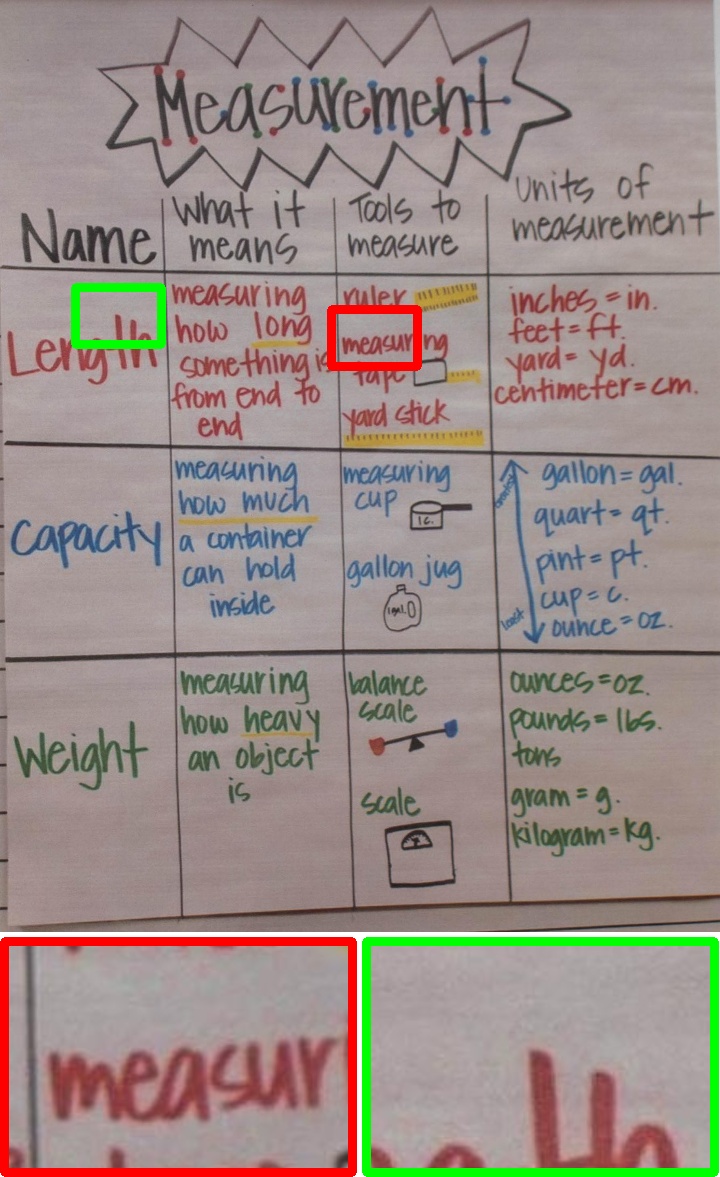}}
        \end{minipage}

        \vspace{0.15cm}
        
        \begin{minipage}[b]{.19\linewidth}
            \centering
            \centerline{\includegraphics[width=\linewidth]{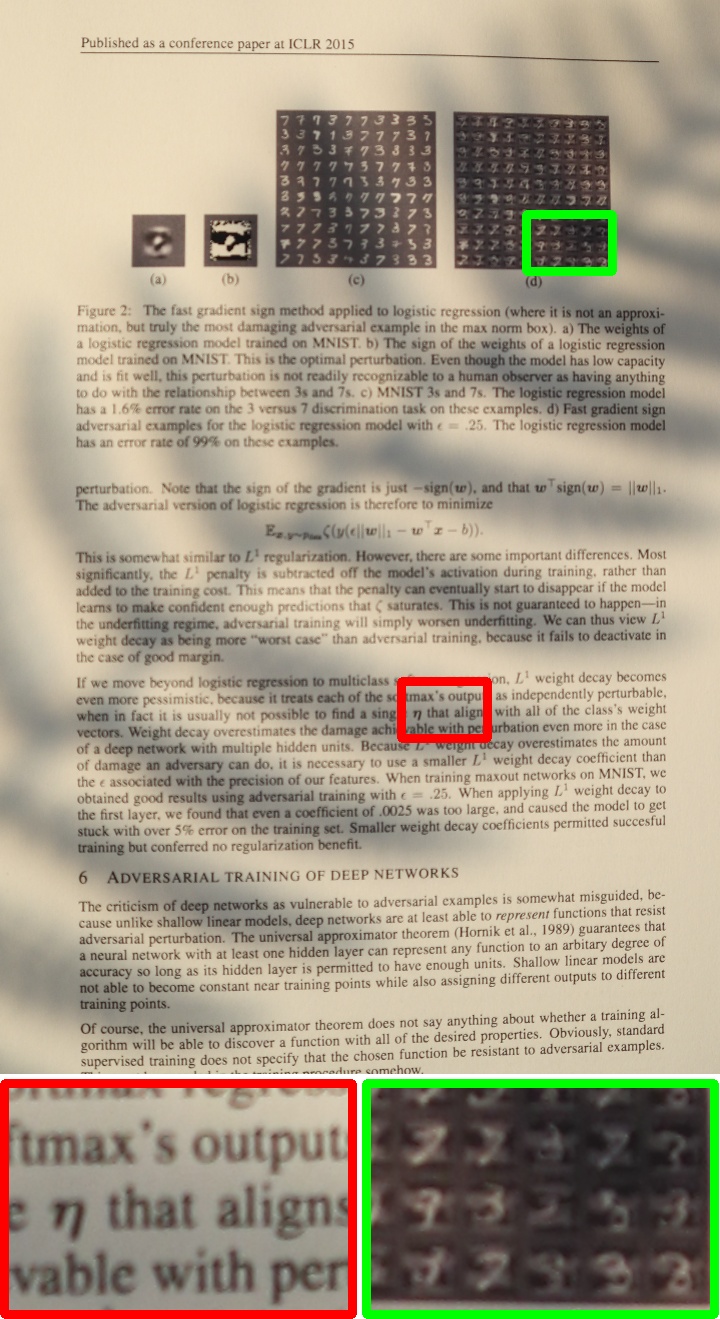}}
            \centerline{Input}\medskip
        \end{minipage}
        \hfill
        \begin{minipage}[b]{.19\linewidth}
            \centering
            \centerline{\includegraphics[width=\linewidth]{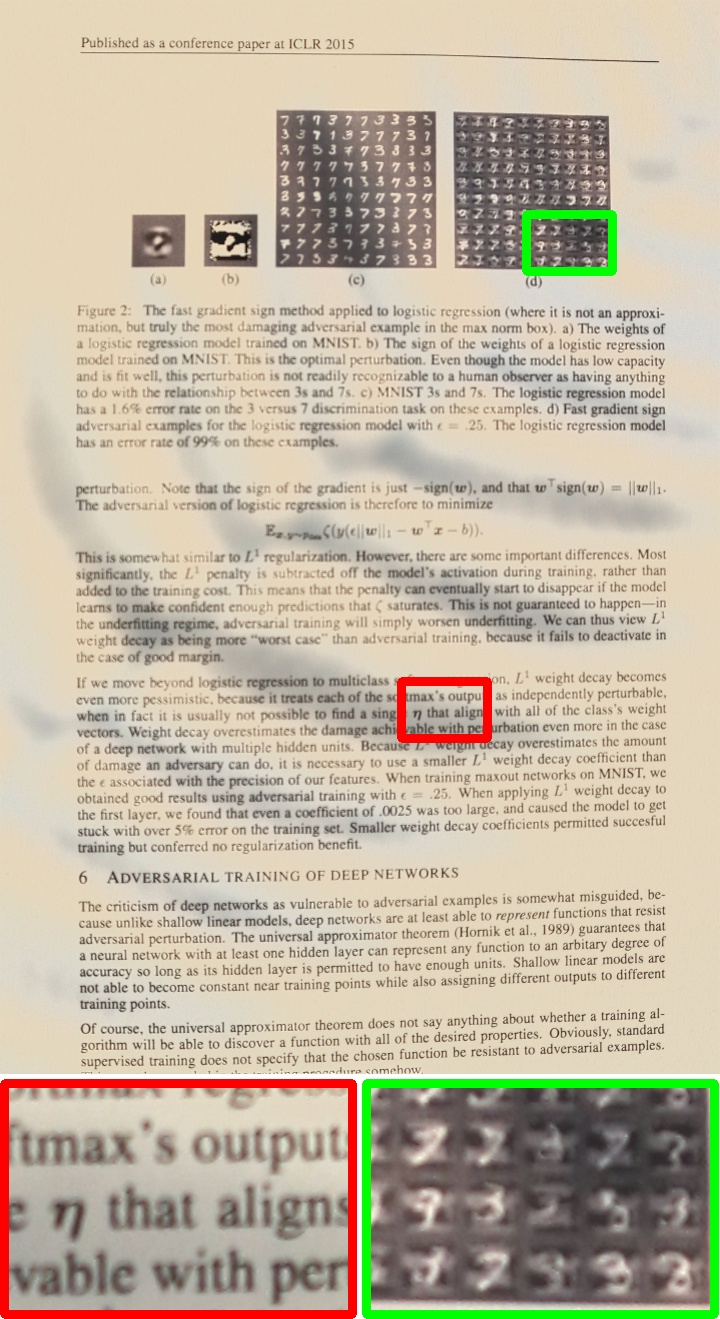
            }}
            \centerline{Jung \etal~\cite{jung2018water}}\medskip
        \end{minipage}
        \hfill
        \begin{minipage}[b]{.19\linewidth}
            \centering
            \centerline{\includegraphics[width=\linewidth]{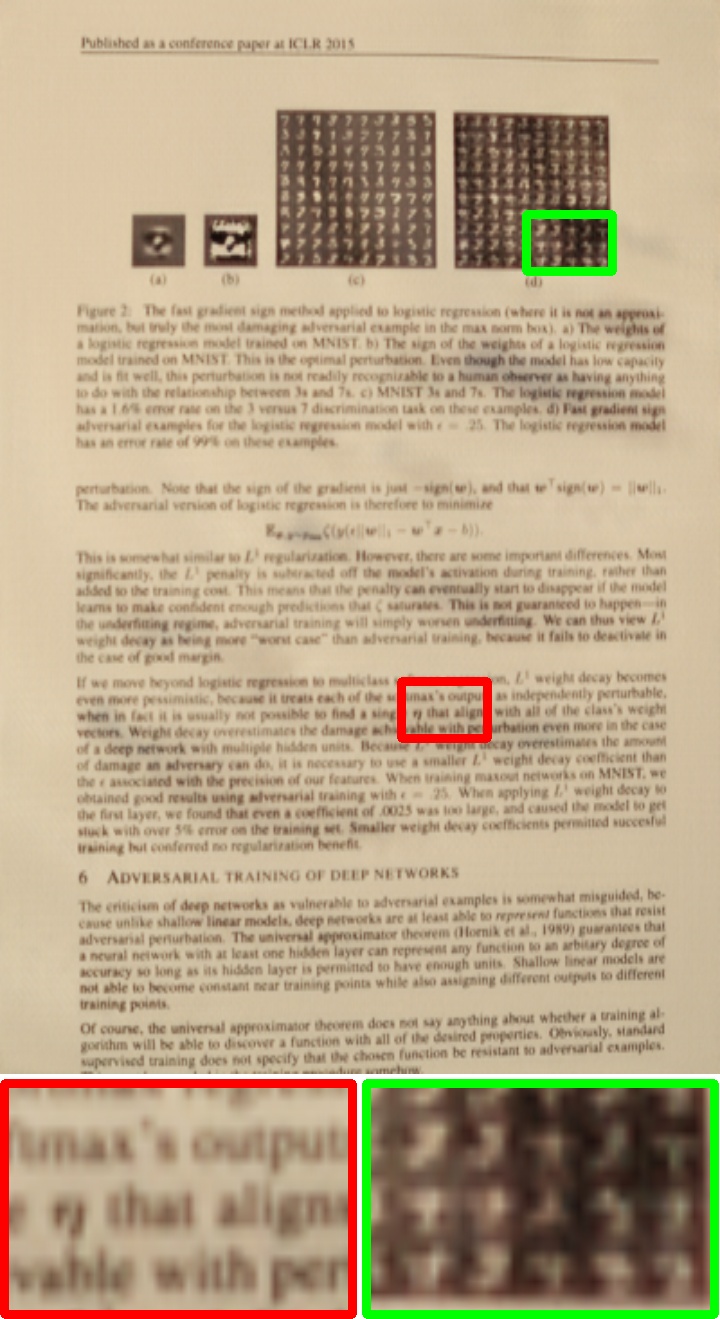}}
            \centerline{MaskShadowGAN~\cite{hu2019mask}}\medskip
        \end{minipage}
        \hfill
        \begin{minipage}[b]{.19\linewidth}
            \centering
            \centerline{\includegraphics[width=\linewidth]{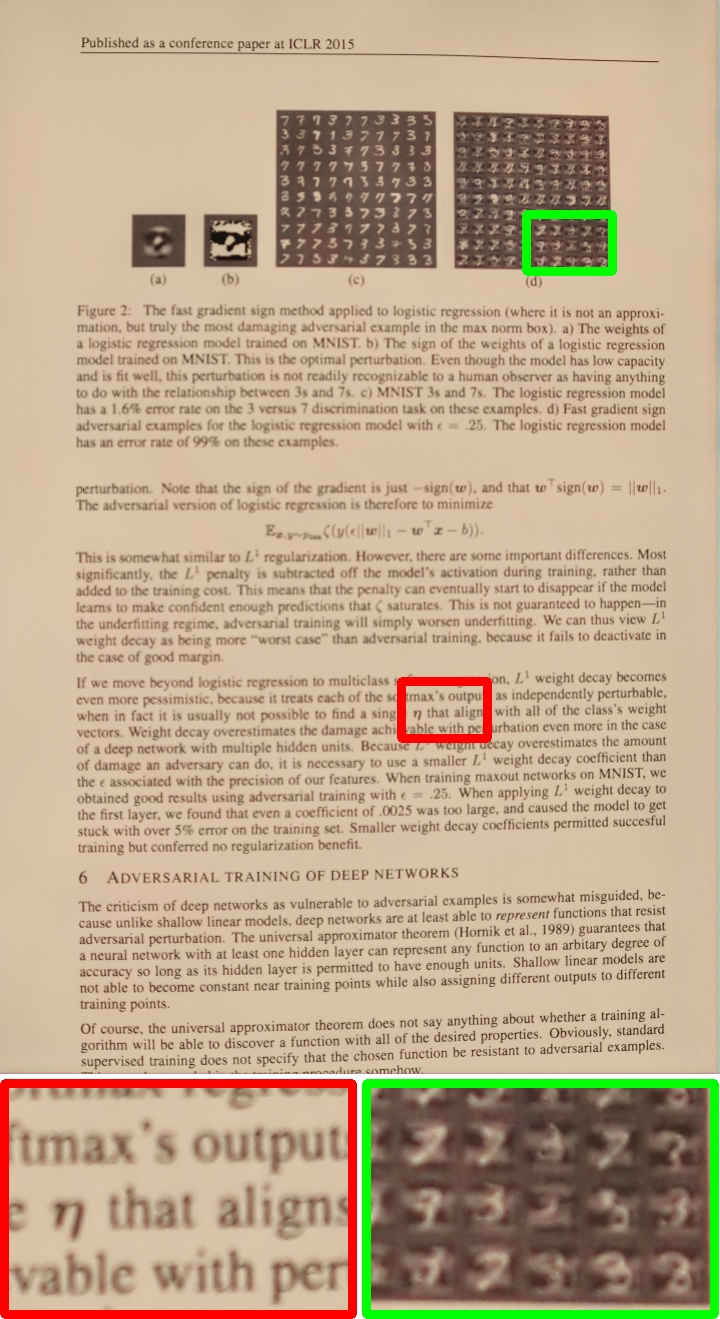}}
            \centerline{Ours}\medskip
        \end{minipage}
        \hfill
        \begin{minipage}[b]{.19\linewidth}
            \centering
            \centerline{\includegraphics[width=\linewidth]{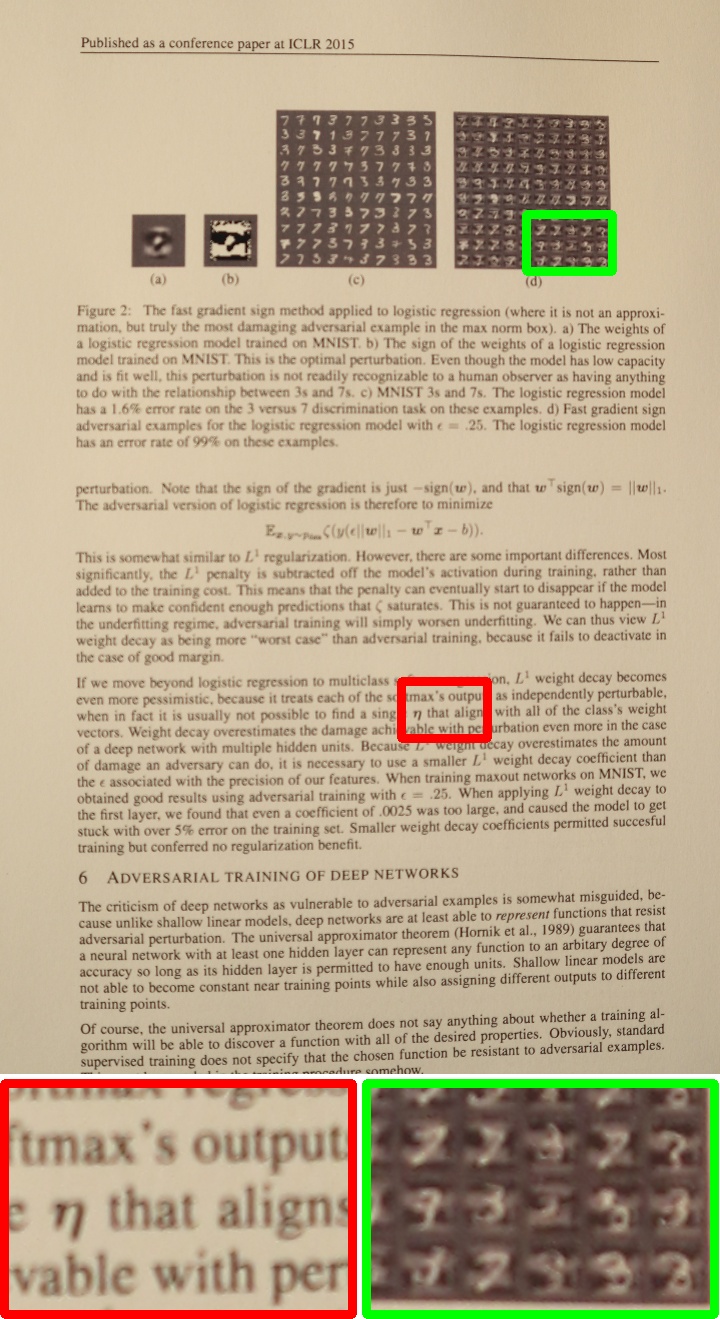}}
            \centerline{Target}\medskip
        \end{minipage}
    \end{minipage}
    \vspace{-1.5em}
    \caption{Qualitative results of the methods comparison in high resolution, the first row represents the results on the Kligler dataset and the second row depicts the results on the SD7K dataset.}
    \label{fig:highres}
\end{figure*}

\subsection{High-Frequency Restoration Module}
\label{sec:high}

After getting the low-frequency features, we also design novel components to recover the high-frequency details of the learned features. In detail, we first upsample $L_3$ and its corresponding result $L_3'$ to match the size of $L_2$, then we learn a contour on the concatenation of $[up(L_3), up(L_3'), L_2]$ via several residual blocks~\cite{he2016deep}, where $up(\cdot)$ means bilinear upsample operation, as illustrated in Fig.~\ref{fig:method}. This is a straightforward method for optimizing global correction compared to normal RGB images because the bands of the high-frequency feature map are very sparse, allowing us to reduce the volume of calculations as discussed in \cite{liang2021high}. This contour will be expanded progressively into a collection of contours to match the remaining high-frequency components. 
As illustrated in Fig.~\ref{fig:method}, given the high-frequency component $L_2$ and the contour $C_{L_2}$, the output $L_2'$ can be written as $L_2' = L_2 \otimes C_{L_2}$, where $\otimes$ represents the pixel-wise multiplication.

\begin{table*}[ht]
\resizebox{\textwidth}{!}{%
\begin{tabular}{l|cccccccccccc|c}
\toprule
 &
  \multicolumn{12}{c|}{SD7K} &
   \\ \cline{2-13}
 &
  \multicolumn{4}{c|}{$512 \times 512$} &
  \multicolumn{4}{c|}{$1024 \times 1024$} &
  \multicolumn{4}{c|}{$2462 \times 3699$~(Full Size)} &
   \\
\multirow{-3}{*}{Method} &
  PSNR$\uparrow$ &
  SSIM$\uparrow$ &
  RMSE$\downarrow$ &
  \multicolumn{1}{c|}{Time(s) $\downarrow$} &
  PSNR$\uparrow$ &
  SSIM$\uparrow$ &
  RMSE$\downarrow$ &
  \multicolumn{1}{c|}{Time(s) $\downarrow$} &
  PSNR$\uparrow$ &
  SSIM$\uparrow$ &
  RMSE$\downarrow$ &
 \multicolumn{1}{c|}{Time(s) $\downarrow$} &
  \multirow{-3}{*}{\shortstack{Param\\(M)}} \\ \hline
Input &
  15.95 &
  0.89 &
  44.09 &
  \multicolumn{1}{c|}{N/A} &
  15.95 &
  0.90 &
  44.09 &
  \multicolumn{1}{c|}{N/A} &
  15.94 &
  0.91 &
  44.14 &
  N/A &
  N/A \\
Wang \etal~\cite{Wang2020ShadowRO} &
  15.31 &
  0.82 &
  47.88 &
  \multicolumn{1}{c|}{$> 10$} &
  15.31 &
  0.85 &
  47.88 &
  \multicolumn{1}{c|}{$> 30$} &
  15.29 &
  0.86 &
  47.95 &
  $> 60$ &
  N/A \\
Wang \etal~\cite{wang2019effective}&
  13.32 &
  0.68 &
  67.48 &
  \multicolumn{1}{c|}{$> 10$} &
  13.32 &
  0.71 &
  67.48 &
  \multicolumn{1}{c|}{$> 30$} &
  13.26 &
  0.74 &
  68.07 &
  $> 60$ &
  N/A \\
Shah \etal~\cite{Shah2018AnIA}&
  9.89 &
  0.71 &
  86.35 &
  \multicolumn{1}{c|}{$> 10$} &
  9.90 &
  0.76 &
  86.35 &
  \multicolumn{1}{c|}{$> 30$} &
  9.88 &
  0.79 &
  86.46 &
  $> 60$ &
  N/A \\
Jung \etal~\cite{jung2018water}&
  19.86 &
  0.92 &
  26.76 &
  \multicolumn{1}{c|}{$> 5$} &
  19.86 &
  0.93 &
  26.76 &
  \multicolumn{1}{c|}{$> 10$} &
  19.82 &
  0.92 &
  26.86 &
  $> 30$ &
  N/A \\
MaskGAN~\cite{hu2019mask} &
  24.82 &
  0.87 &
  15.43 &
  \multicolumn{1}{c|}{0.96} &
  24.82 &
  0.85 &
  15.43 &
  \multicolumn{1}{c|}{2.07} &
  24.67 &
  0.86 &
  15.72 &
  {\color[HTML]{CB0000} \textbf{7.59}} &
  28.29 \\
DHAN~\cite{cun2020towards} &
  25.61 &
  0.85 &
  14.27 &
  \multicolumn{1}{c|}{0.21} &
  25.60* &
  0.83* &
  14.27* &
  \multicolumn{1}{c|}{N/A} &
  25.42* &
  0.85* &
  14.60* &
  N/A &
  27.82 \\
AEFNet~\cite{fu2021auto} &
  24.18 &
  0.95 &
  16.83 &
  \multicolumn{1}{c|}{{\color[HTML]{CB0000} \textbf{0.11}}} &
  23.00* &
  0.90* &
  19.14* &
  \multicolumn{1}{c|}{N/A} &
  22.94* &
  0.90* &
  19.28* &
  N/A &
  391.10 \\
BEDSR-Net~\cite{lin2020bedsr} &
  21.50 &
  0.90 &
  30.52 &
  \multicolumn{1}{c|}{0.85} &
  19.81* &
  0.85* &
  33.03* &
  \multicolumn{1}{c|}{N/A} &
  19.74* &
  0.86* &
  33.18* &
  N/A &
  29.44 \\
ShadowFormer~\cite{guo2023shadowformer} &
  23.71 &
  0.90 &
  17.54 &
  \multicolumn{1}{c|}{0.15} &
  22.69* &
  0.84* &
  19.65* &
  \multicolumn{1}{c|}{N/A} &
  22.64* &
  0.87* &
  19.79* &
  N/A &
  11.35 \\
BMNet~\cite{Zhu_2022_CVPR} &
   24.86 & 
   0.80 & 
   15.59 &
  \multicolumn{1}{c|}{0.18} &
   24.84* &
   0.79* &
   15.67* &
  \multicolumn{1}{c|}{N/A} &
   24.70* &
   0.83* &
   15.96* &
  N/A &
   2.11 \\
Ours &
  {\color[HTML]{CB0000} \textbf{28.69}} &
  {\color[HTML]{CB0000} \textbf{0.97}} &
  {\color[HTML]{CB0000} \textbf{9.98}} &
  \multicolumn{1}{c|}{0.24} &
  {\color[HTML]{CB0000} \textbf{28.68}} &
  {\color[HTML]{CB0000} \textbf{0.97}} &
  {\color[HTML]{CB0000} \textbf{9.98}} &
  \multicolumn{1}{c|}{{\color[HTML]{CB0000} \textbf{0.75}}} &
  {\color[HTML]{CB0000} \textbf{28.67}} &
  {\color[HTML]{CB0000} \textbf{0.96}} &
  {\color[HTML]{CB0000} \textbf{10.00}} &
  7.93 &
  29.34 \\ \bottomrule
\end{tabular}%
}
\vspace{0.5em}
\caption{Quantitative results of the ablation study on three datasets in different resolutions. The best result is highlighted in red and bold. The asterisk (*) indicates that the result is obtained from a low-resolution result~(512$\times$512) and then improved using a super-resolution technique~\cite{ledig2017photo} to address an out-of-memory error.}
\label{table:high}
\end{table*}

After multiplication, the Texture Recovery Module (TRM) is then designed to improve elaboration, inspired by the successful natural shadow removal methods~\cite{cun2020towards, xu2022shadow}, which contain a series of the dilated convolutions and attentive aggregation nodes to recover the detailed features. 

In detail, to increase the reception fields on the high-resolution map, we incorporate dilated convolution~\cite{yu2015multi}. Dilated convolution inserts spaces between the kernel values, effectively increasing the receptive field of the filter without losing resolution or coverage over the input. The amount of dilation is controlled by a dilation rate parameter. For a dilation rate of 1, the layer behaves like a regular convolution. For rates larger than 1, the filter is effectively widened to a larger reception fields. This allows the network to incorporate a wider context and capture features at different scales in a computationally efficient manner compared to using very large filters. 
As for the Attentive Aggregation Node, to re-weight the significance of feature maps, each aggregation node employs a squeeze-and-excitation block~\cite{hu2018squeeze} to reweight the feature. A $3 \times 3$ convolution layer is then employed to compress the features and match the original channels. Finally, a spatial pooling pyramid~(SPP)~\cite{he2015spatial} is introduced at the conclusion of the TRM to facilitate the remixing of multi-context features. After getting the learned high-frequency of each layer, we use the method introduced in Sec.~\ref{sec:lp} to recover the image and calculate the supervision.

\subsection{Loss functions}
\label{sec:loss}
For training, we utilize the loss functions of $\mathbf{Smooth L_1}$~\cite{girshick2015fast} $ \mathcal{L}_{l1}^{smooth} $ and $\mathbf{SSIM}$~\cite{wang2004image} $ \mathcal{L}_{SSIM}$ between the target image and the reconstructed image from LP, both two losses are helpful to preserve the details of the image, which can be represented as: $ \mathcal{L}_{total} = \mathcal{L}_{l1}^{smooth} + \lambda * \mathcal{L}_{SSIM}$, 
where we set the weight of $\mathcal{L}_{SSIM}$ $\lambda$ to 0.4 empirically.

\begin{figure*}[ht]
    \begin{minipage}[b]{1.0\linewidth}
        \begin{minipage}[b]{.24\linewidth}
            \centering
            \centerline{\includegraphics[width=\linewidth]{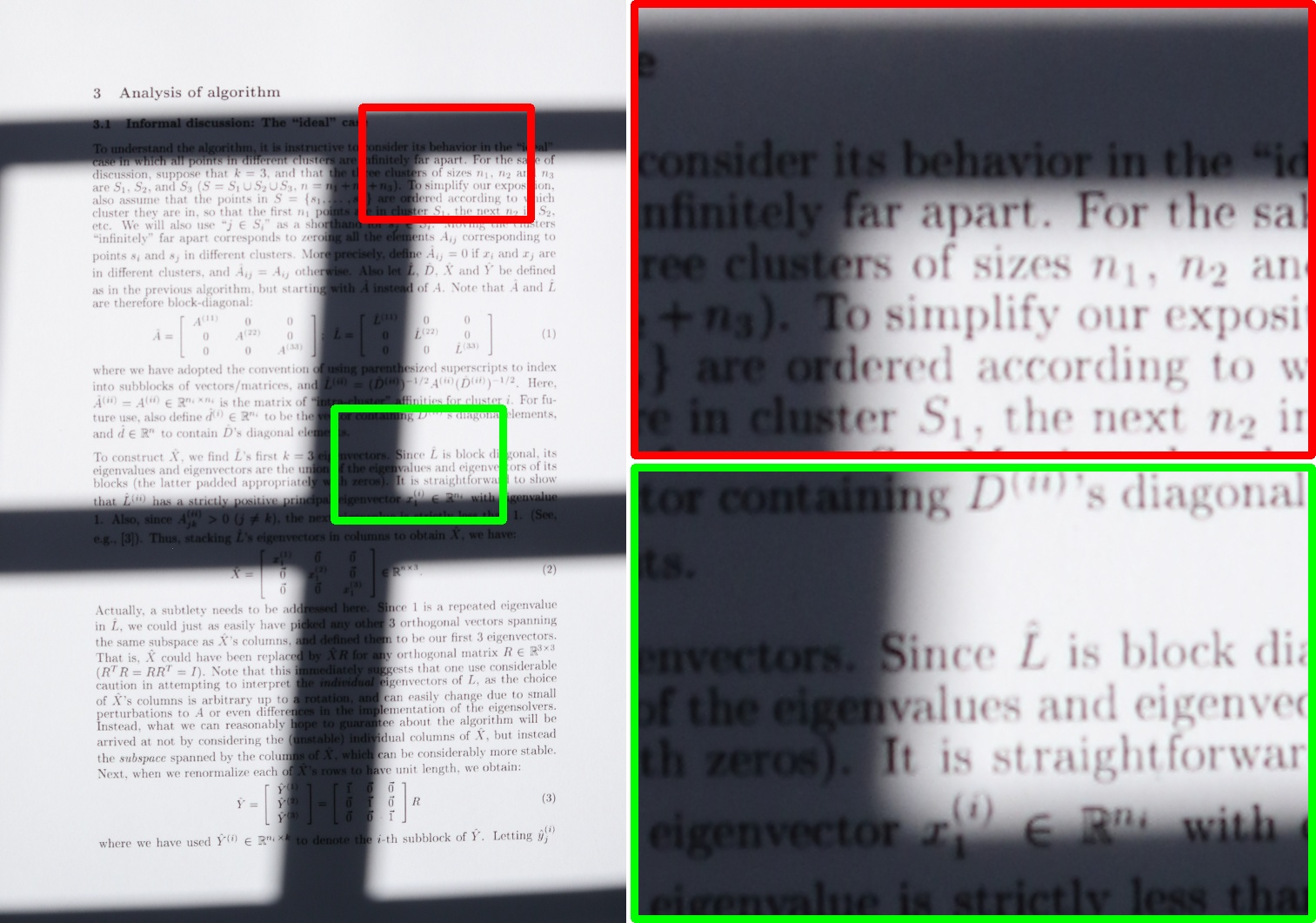}}
            \centerline{Input}\medskip
        \end{minipage}
        \hfill
        \begin{minipage}[b]{.24\linewidth}
            \centering
            \centerline{\includegraphics[width=\linewidth]{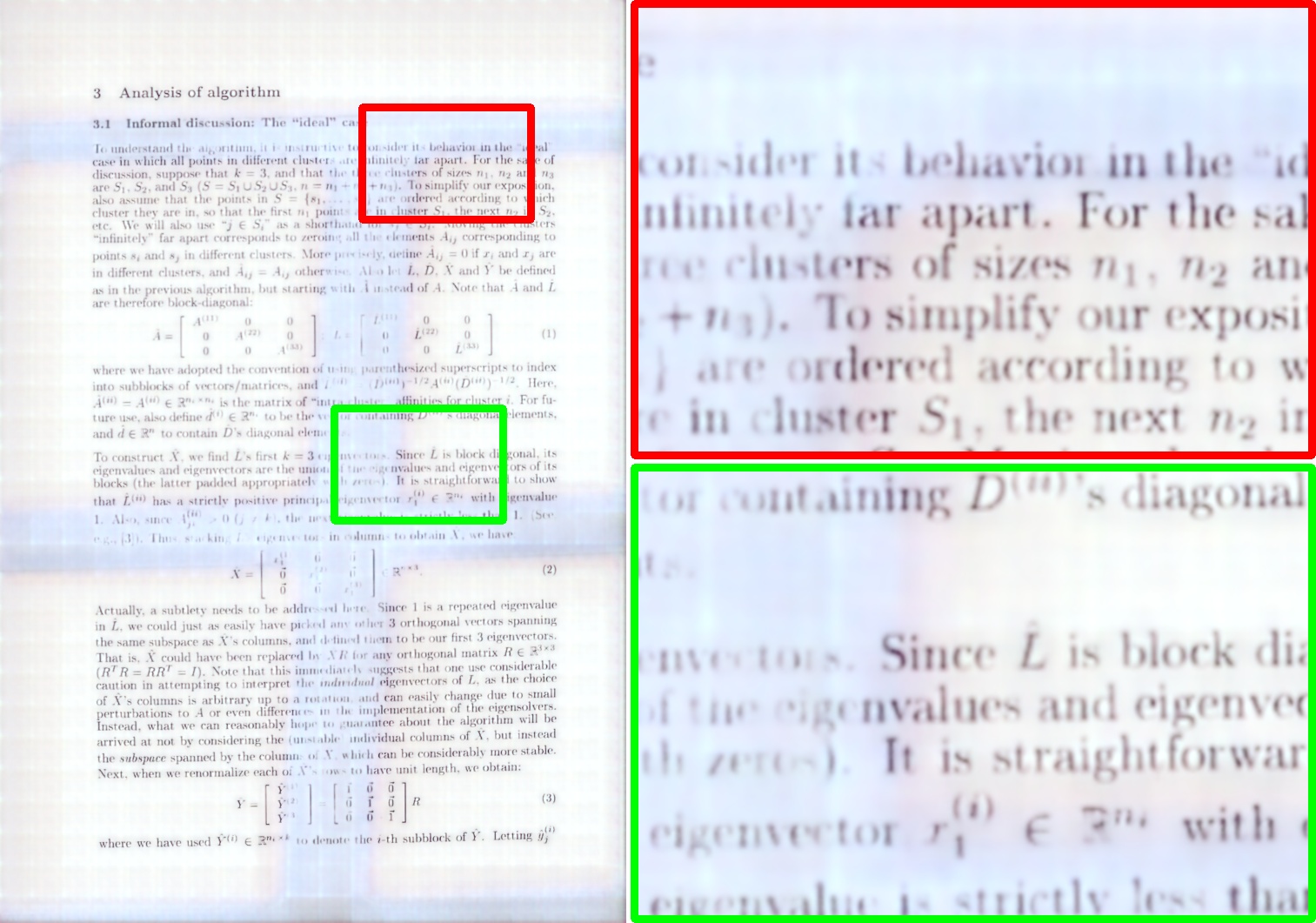}}
            \centerline{Ours w/ Depth=3}\medskip
        \end{minipage}
        \hfill
        \begin{minipage}[b]{.24\linewidth}
            \centering
            \centerline{\includegraphics[width=\linewidth]{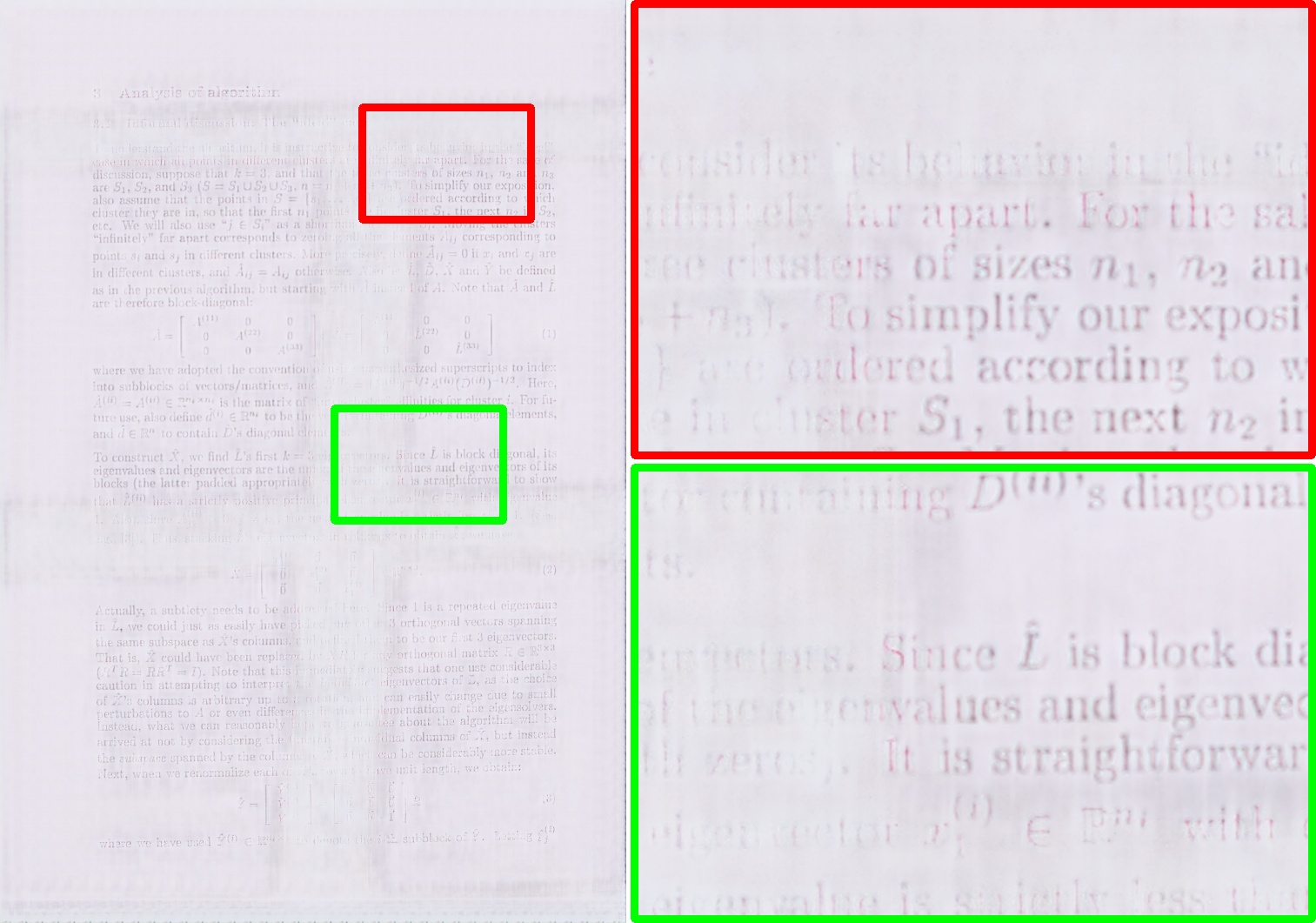}}
            \centerline{Ours w/o TRM}\medskip
        \end{minipage}
        \hfill
        \begin{minipage}[b]{.24\linewidth}
            \centering
            \centerline{\includegraphics[width=\linewidth]{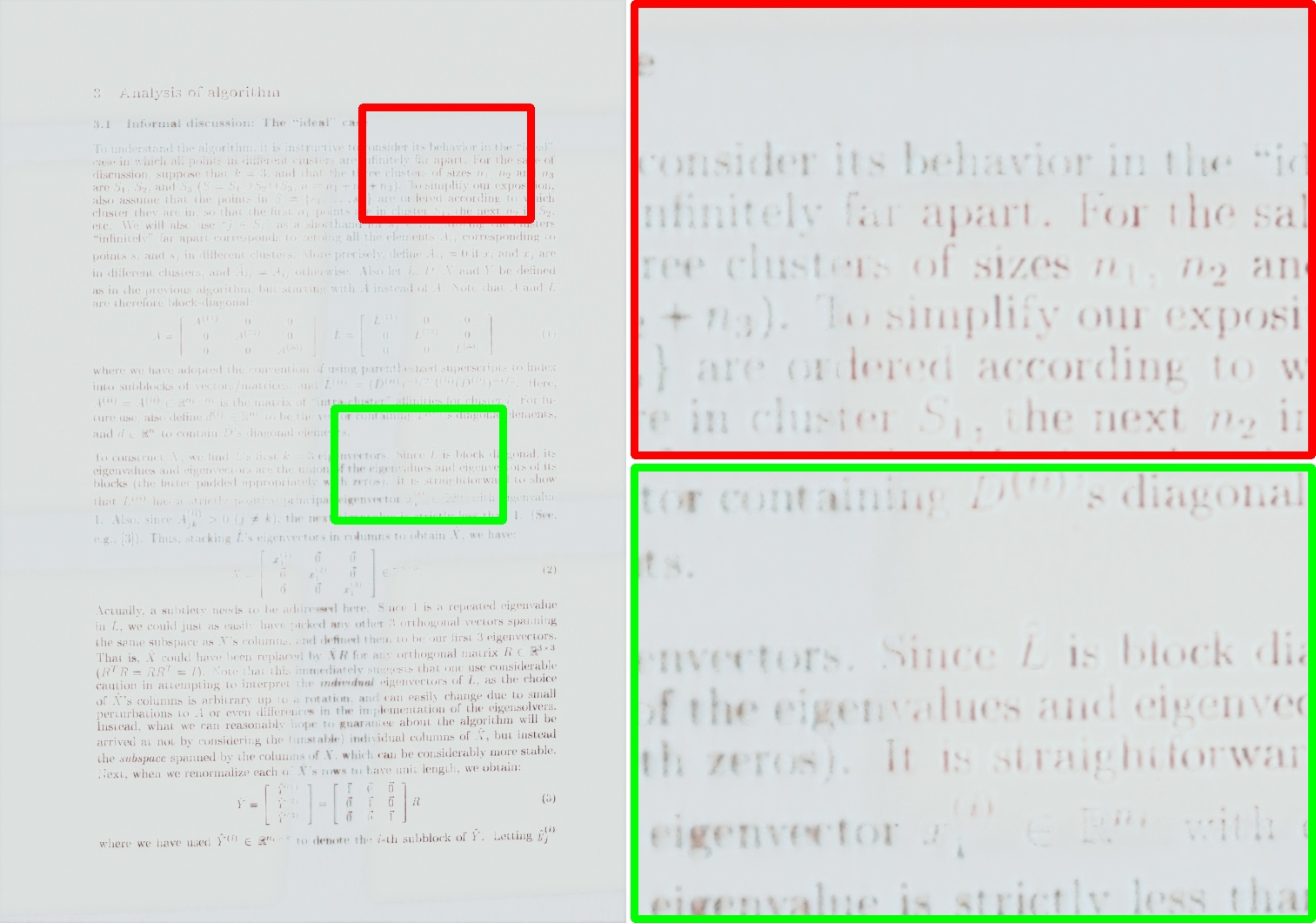}}
            \centerline{Ours w/o DFE}\medskip
        \end{minipage}
    \end{minipage}
    \begin{minipage}[b]{1.0\linewidth}
        \begin{minipage}[b]{.24\linewidth}
            \centering
            \centerline{\includegraphics[width=\linewidth]{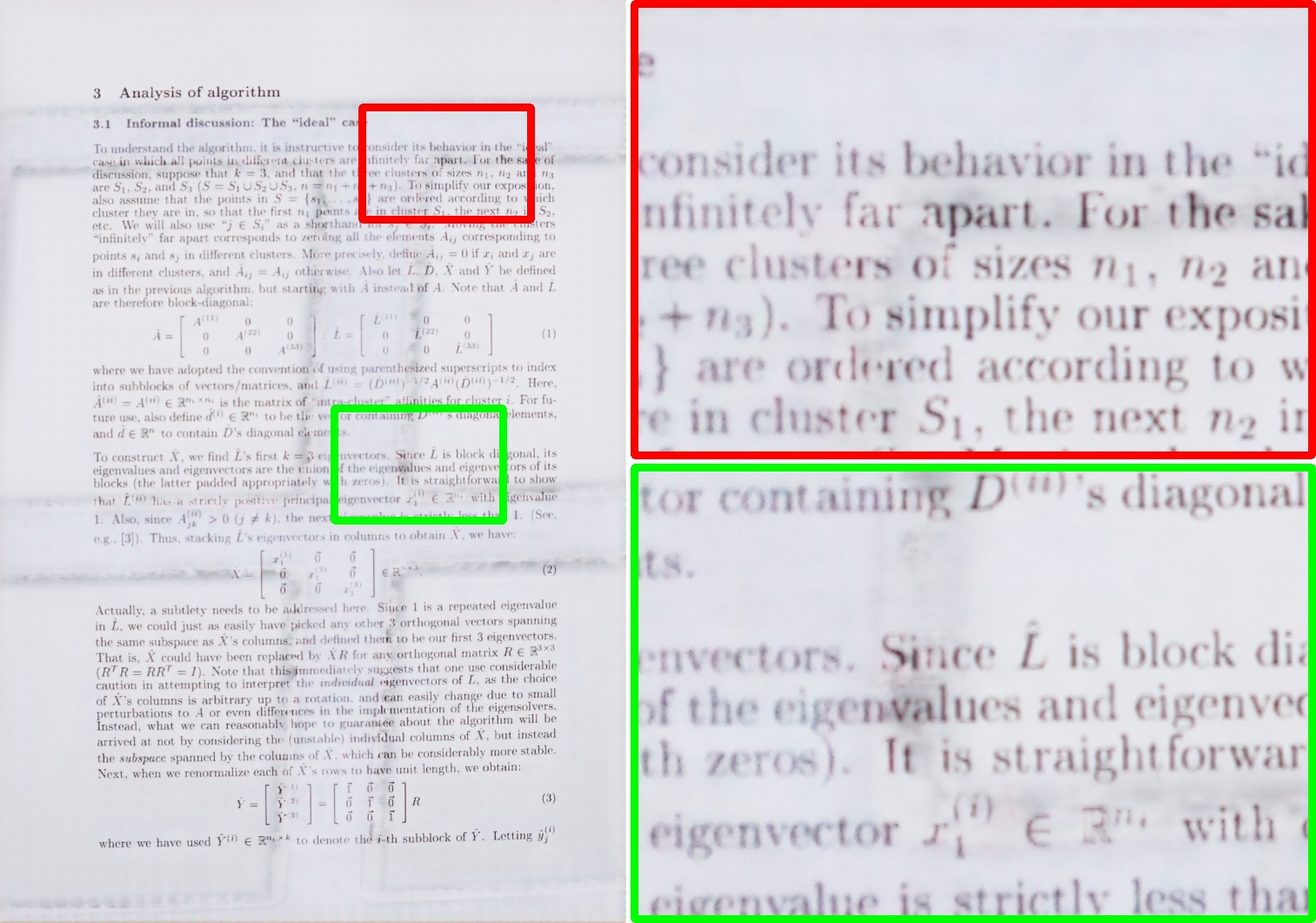}}
            \centerline{Ours w/o Synthesis}\medskip
        \end{minipage}
        \hfill
        \begin{minipage}[b]{.24\linewidth}
            \centering
            \centerline{\includegraphics[width=\linewidth]{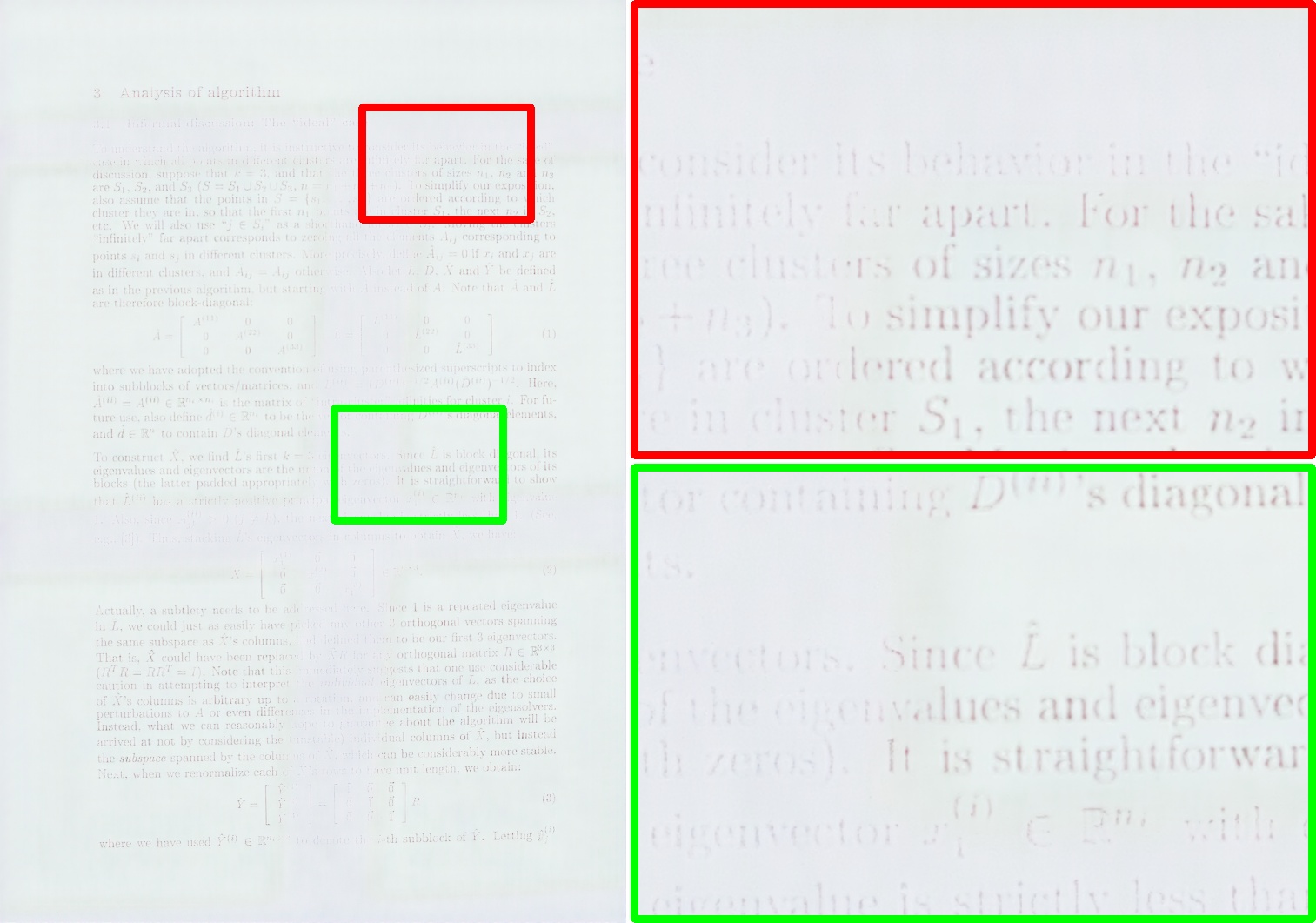}}
            \centerline{Ours w/o DAT}\medskip
        \end{minipage}
        \hfill
        \begin{minipage}[b]{.24\linewidth}
            \centering
            \centerline{\includegraphics[width=\linewidth]{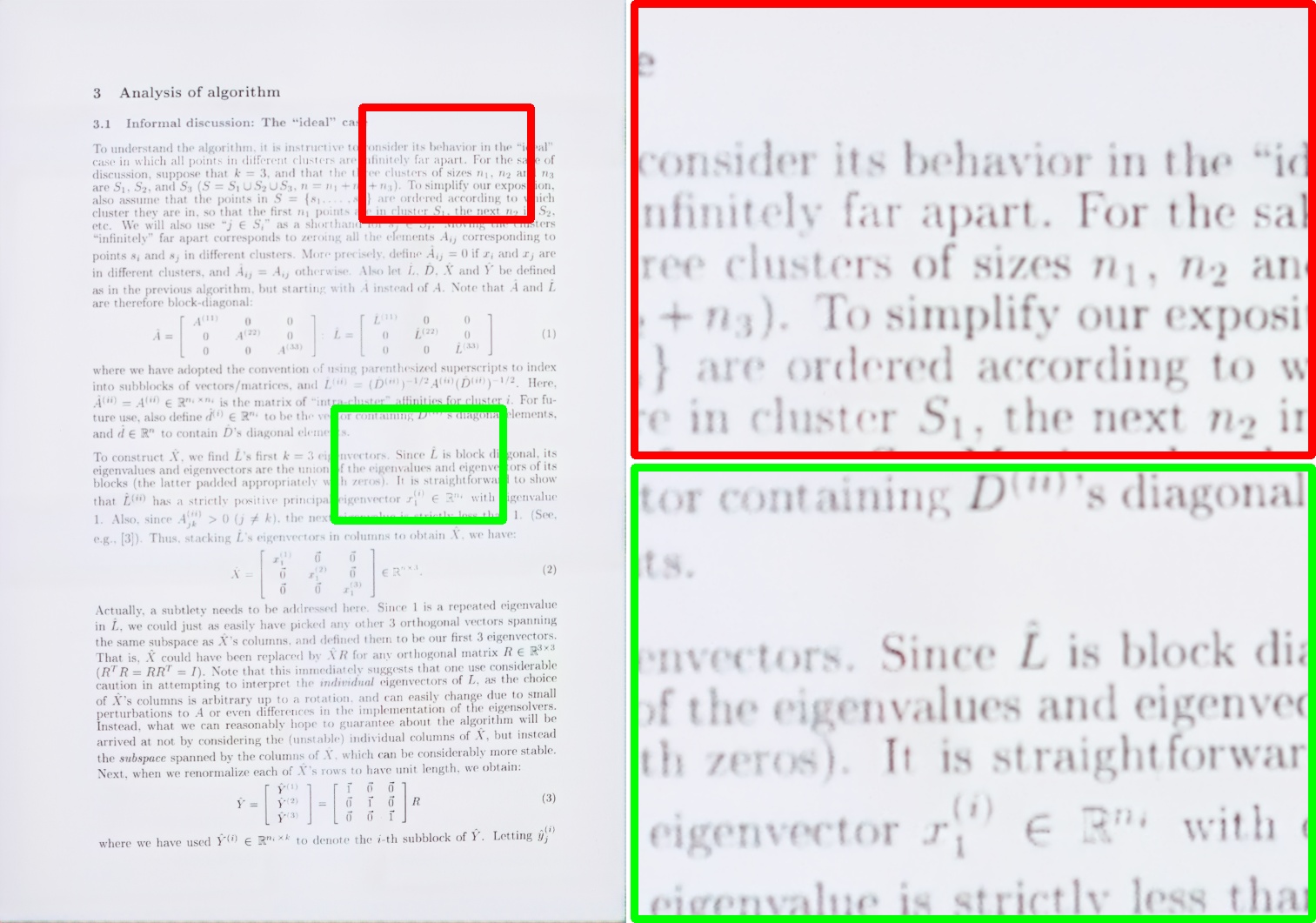}}
            \centerline{Ours}\medskip
        \end{minipage}
        \hfill
        \begin{minipage}[b]{.24\linewidth}
            \centering
            \centerline{\includegraphics[width=\linewidth]{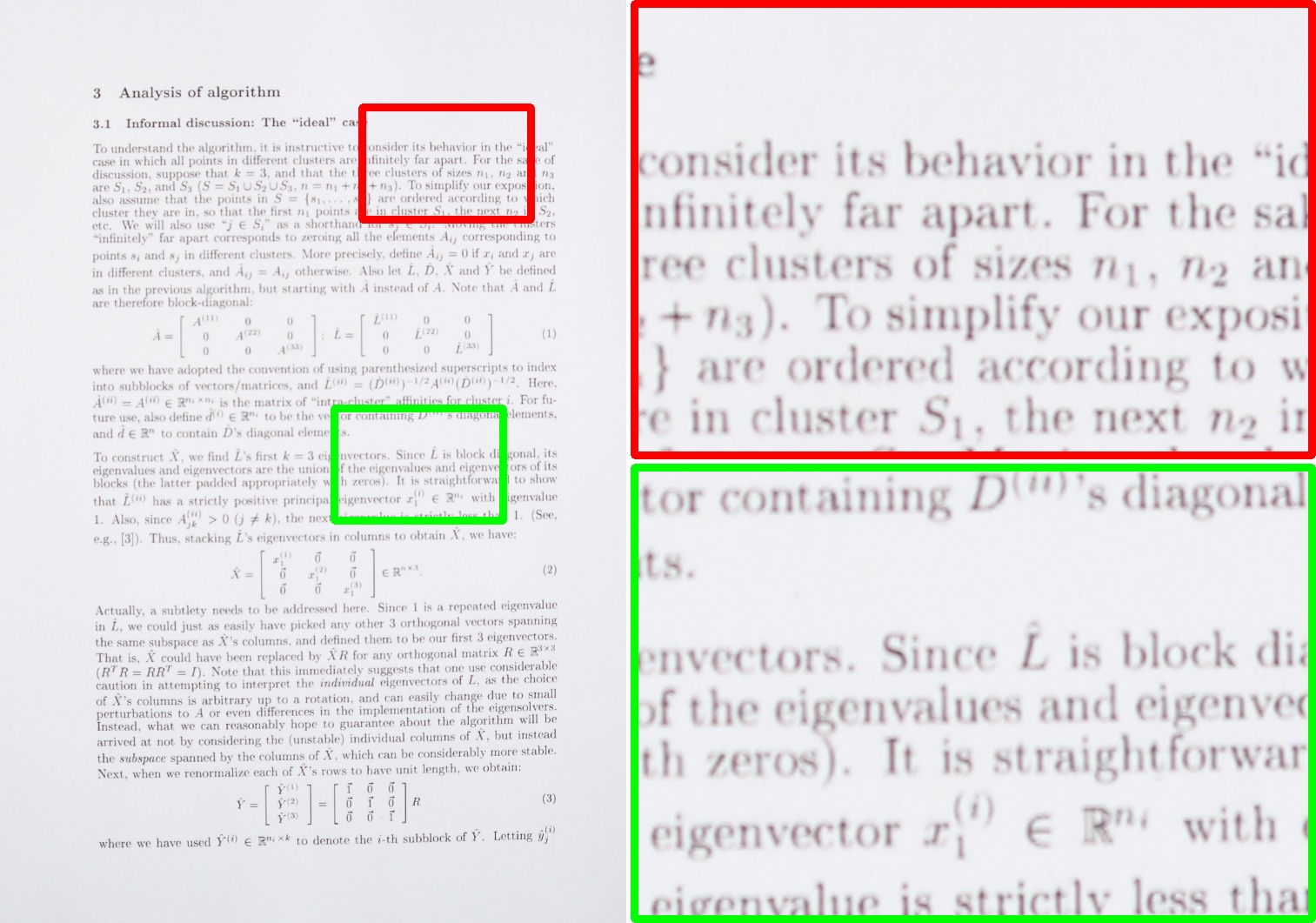}}
            \centerline{Target}\medskip
        \end{minipage}
    \end{minipage}
    \caption{Qualitative results of the ablation studies on different modules of the proposed FSENet.}
    \label{fig:ablation}
\end{figure*}

\begin{table*}[ht]
\centering
\begin{tabular}{l|ccc|ccc|ccc}
\toprule
                                 & \multicolumn{3}{c|}{Jung (Original)}                                                                                 & \multicolumn{3}{c|}{Kliger (Original)}                                                                               & \multicolumn{3}{c}{SD7K (Original)}                                                                   \\ \cline{2-10} 
\multirow{-2}{*}{Ablation Study} & \multicolumn{1}{c|}{PSNR$\uparrow$} & \multicolumn{1}{c|}{SSIM$\uparrow$} & RMSE$\downarrow$              & \multicolumn{1}{c|}{PSNR$\uparrow$} & \multicolumn{1}{c|}{SSIM$\uparrow$} & RMSE$\downarrow$              & \multicolumn{1}{c|}{PSNR$\uparrow$} & \multicolumn{1}{c|}{SSIM$\uparrow$} & RMSE$\downarrow$ \\ \hline
Input & 12.93 & 0.79 & 61.34 & 13.25 & 0.81 & 56.75 & 15.94 & 0.91 & 44.14 \\

Ours w/ Depth=3                          & 20.97                              & 0.83                               & 23.19                        & 26.33                              & 0.91                               & 13.66                        & 20.83                              & 0.93                               & 24.02           \\
Ours w/o TRM          & 19.50                              & 0.81                               & 27.60                        & 27.10                              & 0.92                               & 12.78                        & 18.44                              & 0.86                               & 35.92          \\
Ours w/o DFE               & 21.90                              & 0.84                               & 21.02                        & 25.16                              & 0.92                              & 14.89                        & 19.70                              & 0.88                               & 29.07           \\

Ours w/o DAT          & 22.68                              & 0.85                               & 19.28                        & 22.70                              & 0.88                               & 20.19                        & 19.03                              & 0.87                               & 32.94           \\
Ours w/o Synthesis                  & 22.21                              & 0.85                               & 21.33                        & 24.79                              & 0.92                               & 15.54                        & 22.94                              & 0.93                               & 21.93           \\
Ours Full                     & {\color[HTML]{CB0000} \textbf{23.60}}       & {\color[HTML]{CB0000} \textbf{0.85}}        & {\color[HTML]{CB0000} \textbf{17.56}} & {\color[HTML]{CB0000} \textbf{28.98}}       & {\color[HTML]{CB0000} \textbf{0.93}}        & {\color[HTML]{CB0000} \textbf{10.60}} & {\color[HTML]{CB0000} \textbf{28.67}} & {\color[HTML]{CB0000} \textbf{0.96}} &  {\color[HTML]{CB0000} \textbf{10.00}} \\ \bottomrule
\end{tabular}
\vspace{0.5em}
\caption{Quantitative results of the ablation study on three datasets in high-resolution. The best result is highlighted in red and bold.}
\label{table:ablation}
\end{table*}

\section{Experiments}


\subsection{Implementation Details} 
We evaluate the performance of different network structures with different resolutions on the proposed SD7K dataset, where the batch size is set to 4. We also give more experiments on the low-resolution of 512$\times$512 of Jung~\cite{jung2018water}, Kligler~\cite{kligler2018document}, and SD7K for fair comparison following previous methods, we give more details in the supplementary materials.  To further utilize the dataset, we also incorporate a simple shadow synthesis pipeline to further increase the diversity of the training process. 
We investigate several potential options, including alpha composition~\cite{shade1998layered} and Perlin noise generation~\cite{hart2001perlin} for comparison, and we empirically choose alpha composition to synthesize shadows through the shadow-free image and the mask on the fly via random alpha values in the range between $[0.2, 0.7]$.

As for the evaluation metrics, we employ three widely-used metrics: Root Mean Square Error~(RMSE), Peak Signal-to-Noise Ratio~(PSNR), and Structural Similarity~(SSIM). While PSNR and SSIM are commonly used in image restoration evaluation~\cite{chen2022simple, s2am, s2crnet}, RMSE is the most important and frequently used metric specifically for shadow removal~\cite{cun2020towards, guo2023shadowformer}.

\subsection{Comparisons with State-of-the-Art Methods}
As presented in Tab.~\ref{table:high}, we compared FSENet with both traditional and learning-based models of various resolutions using publicly available implementations of these methods, where the proposed method outperforms all other methodologies in most metrics.

Furthermore, our analysis reveals that traditional methods, such as~\cite{jung2018water,Shah2018AnIA}, rely on physics-based assumptions that are not always correct, resulting in unstable performance across different datasets as shown in Fig.~\ref{fig:highres}. Deep learning models, on the other hand, are highly dependent on the dataset magnitude, and their performance is relatively consistent across different scenarios.

We also find that some deep learning models that are effective in low-resolution shadow removal do not scale well to high-resolution documents. They either compromise the aspect ratio of the document or lead to out-of-memory errors. While traditional methods can still remove shadows, they are time-consuming, requiring pixel-by-pixel calculations of illumination difference~\cite{Shah2018AnIA} and background estimation~\cite{wang2019effective}. At present, only Mask-ShadowGAN~\cite{hu2019mask} and the proposed FSENet can effectively remove document shadows in high-resolution scenarios. Despite having more parameters and requiring more time to produce results, FSENet outperforms other methods quantitatively.

\subsection{Ablation Studies}
As presented in Tab.~\ref{table:ablation} and Fig.~\ref{fig:ablation}, we conduct our ablation studies at the original high-resolution documents, examining the effectiveness of several components in our algorithm. These components include the depth of the Laplacian Pyramid, the efficiency of TRM, DFE, and DAT modules, and the effectiveness of our shadow synthesis pipeline. 

We first evaluate the depth of the proposed Laplacian Pyramid. From Tab.~\ref{table:ablation}, with the depth of the Laplacian Pyramid increases~(Our w/ Depth=3), the magnitude of the low-frequency components decreases, affecting the low-frequency deshadowing and the reconstructed outcome. Consequently, the model's performance progressively declines as the depth increases. We then evaluate the performance of the proposed DFE and DAT modules, which are also responsible for low-frequency deshading, primarily influencing the pixel-wise variation of the final result. 
Although each module individually contributes to enhancing shadow elimination performance, their combined network produces the best results.
As for the high-frequency modules, the TRM module handles high-frequency components such as contours and edges, thus contributing significantly to the overall efficacy of our algorithm. Finally, we also evaluate the importance of the proposed shadow synthesis pipeline, where w/o shadow synthesis reduces the performance a lot, especially on the two small datasets.

\subsection{Limitations}
Although our model is capable of removing shadows in high-resolution images, the current network is not able to run on real-time or on edge devices, such as mobile phones, due to the larger parameters and heavy GPU computations. In the future, we will explore different approaches to further reduce the required parameters, making it usable on smart devices.	

\section{Conclusion}
In this paper, we tackle the problem of removing shadows from high-resolution documents. To achieve this goal, we consider the contribution of the dataset and the network. As for the dataset, we gather a high-resolution dataset SD7K, which is comprised of over 7000 triplets of real-world document images with distinct characteristics under various illumination conditions. This dataset is an order of magnitude greater than previous datasets and allows us to validate our approach in a more diverse and comprehensive manner. As for the network structure, we propose FSENet, a frequency-aware method to remove shadows in higher resolution via frequency decomposition. 
We employ LP and a series of meticulously designed attention blocks, enabling each component to be learned separately. Powered by our dataset and network, our method achieves a new state-of-the-art for document shadow removal. We believe both the network structure and the dataset will benefit the research community on this topic.

\noindent\textbf{Acknowledgements.}
This work is supported in part by the University of Macau under Grant MYRG2022-00190-FST, in part by the Science and Technology Development Fund, Macau SAR, under Grant 0034/2019/AMJ, Grant 0087/2020/A2 and Grant 0049/2021/A.

{\small
\bibliographystyle{ieee_fullname}
\bibliography{egbib}
}

\clearpage
\section{The Experiments on the Resolution of 512}
We also evaluate the performance on three datasets: Jung~\cite{jung2018water}, Kligler~\cite{kligler2018document}, and the proposed SD7K, in a relatively low resolution of 512 $\times$ 512 where the divisions between training and testing are shown in Table~\ref{table:eval}. We train each dataset separately to satisfy the settings of the previous works. The batch size is set to 1 for the other two datasets due to the limited samples. 

\begin{table}[h]
\centering
\begin{tabular}{lcc}
\toprule
Datasets  & \# of Training  & \# of Testing  \\ \hline
Jung~\cite{jung2018water}    & 67               & 20              \\
Kligler~\cite{kligler2018document} & 272              & 28              \\
SD7K    & 6479             & 760             \\ \bottomrule
\end{tabular}
\vspace{0.5em}
\caption{The settings of the training and testing.}
\label{table:eval}
\end{table}

\begin{figure*}[ht]
    \begin{minipage}[b]{1.0\linewidth}
        \begin{minipage}[b]{.24\linewidth}
            \centering
            \centerline{\includegraphics[width=\linewidth]{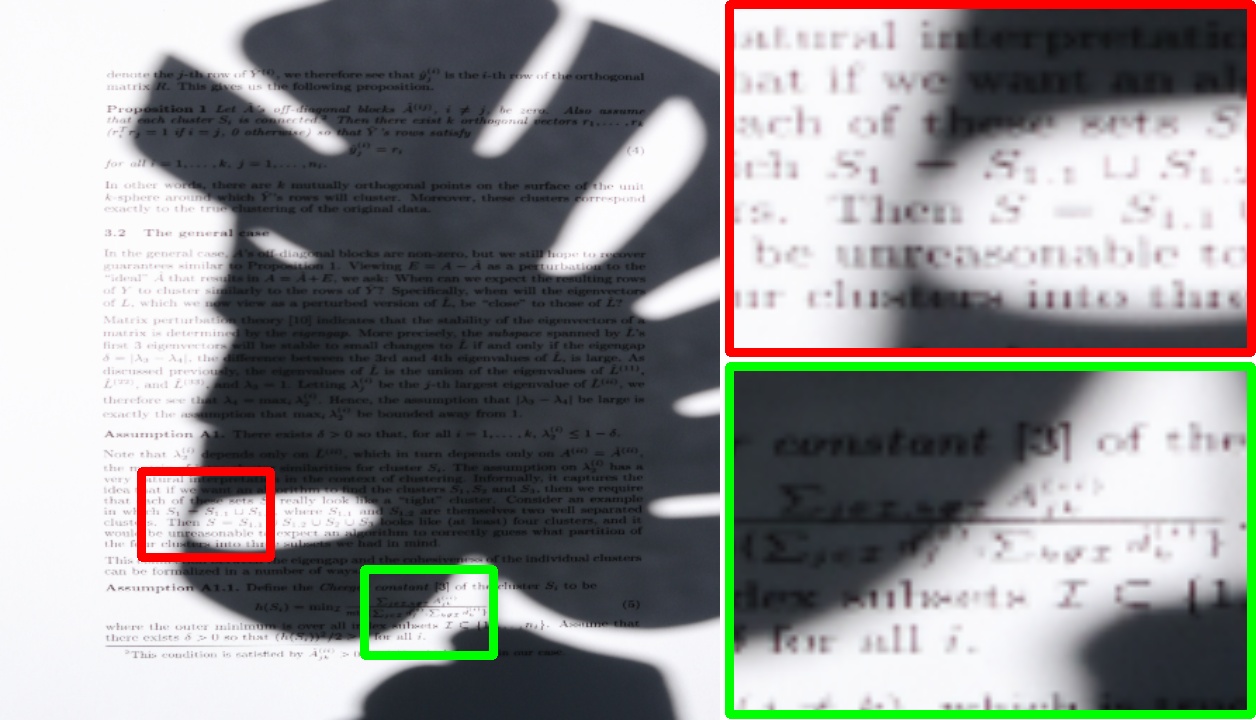}}
            \centerline{Input}\medskip
        \end{minipage}
        \hfill
        \begin{minipage}[b]{.24\linewidth}
            \centering
            \centerline{\includegraphics[width=\linewidth]{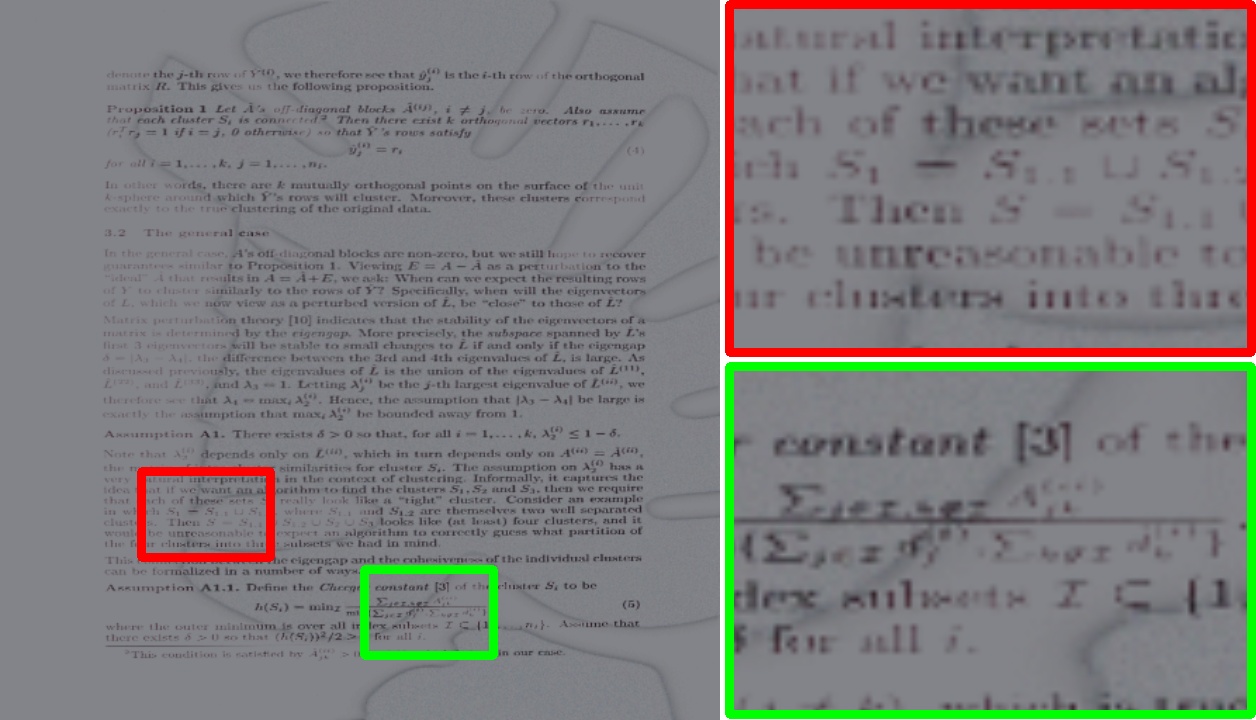
            }}
            \centerline{Wang \etal~\cite{Wang2020ShadowRO}}\medskip
        \end{minipage}
        \hfill
        \begin{minipage}[b]{.24\linewidth}
            \centering
            \centerline{\includegraphics[width=\linewidth]{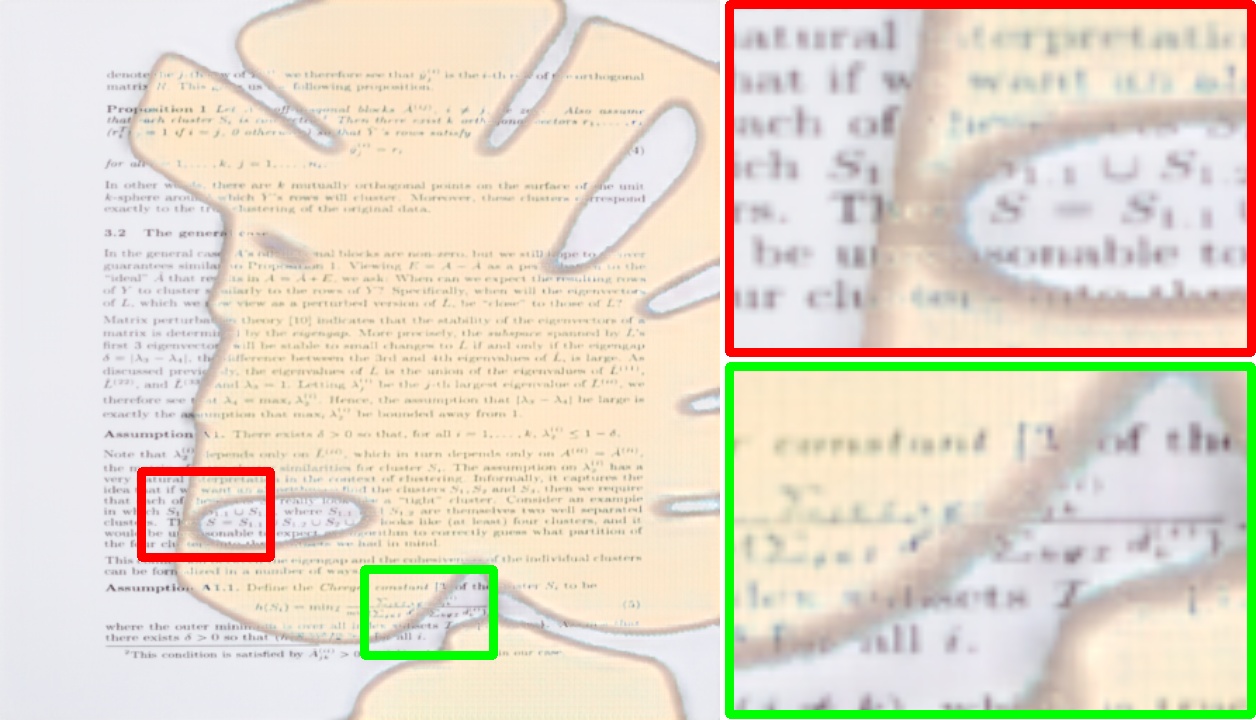}}
            \centerline{SP+M+I Net~\cite{le2021physics}}\medskip
        \end{minipage}
        \hfill
        \begin{minipage}[b]{.24\linewidth}
            \centering
            \centerline{\includegraphics[width=\linewidth]{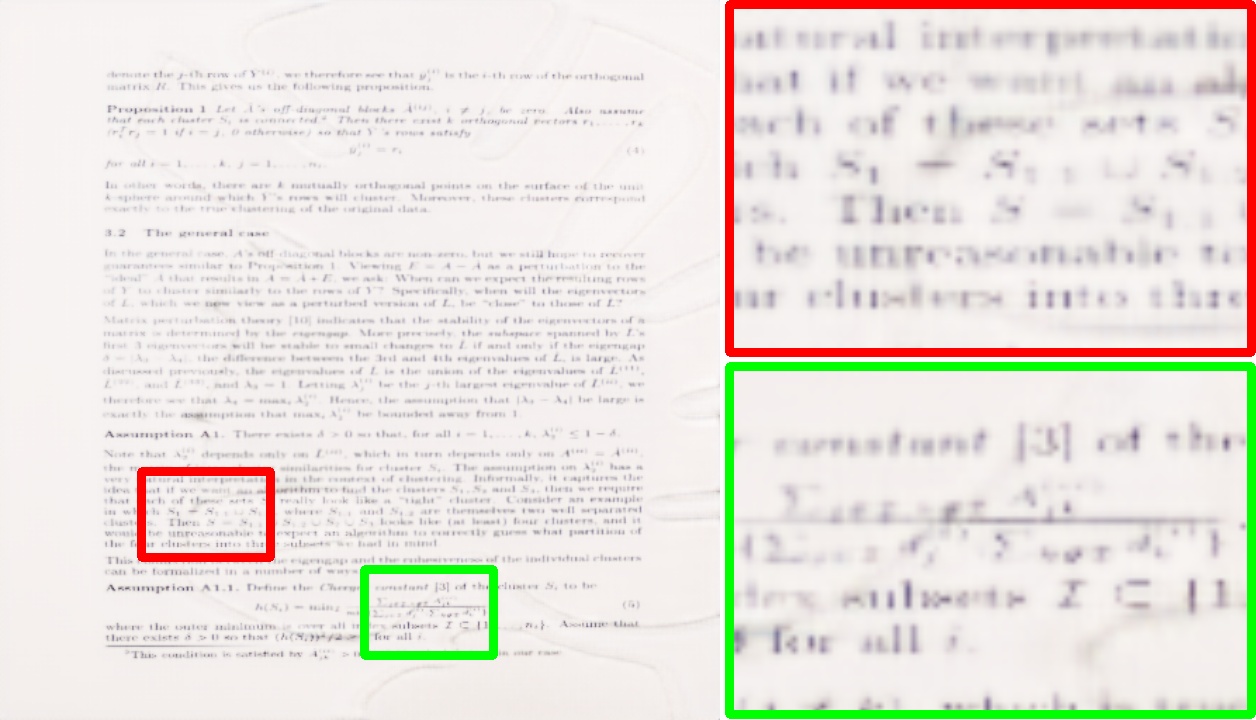}}
            \centerline{BEDSR-Net~\cite{lin2020bedsr}}\medskip
        \end{minipage}
    \end{minipage}
    \begin{minipage}[b]{1.0\linewidth}
        \begin{minipage}[b]{.24\linewidth}
            \centering
            \centerline{\includegraphics[width=\linewidth]{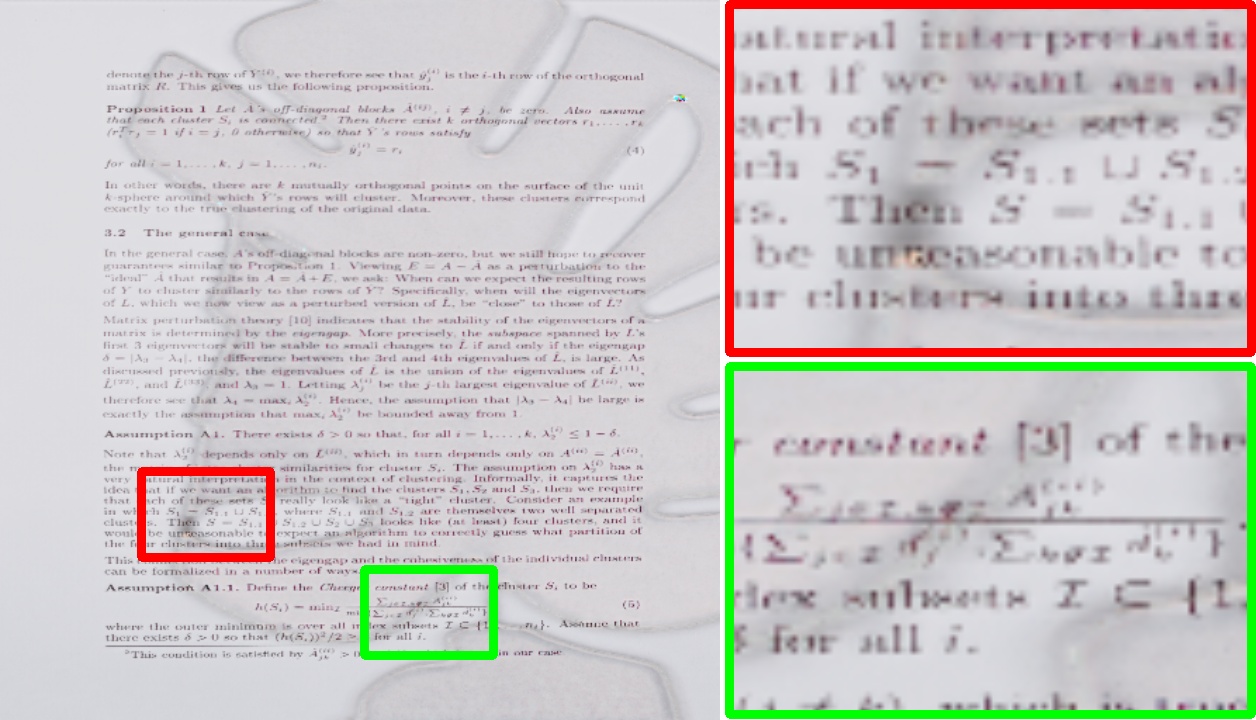}}
            \centerline{SG-ShadowNet~\cite{wan2022style}}\medskip
        \end{minipage}
        \hfill
        \begin{minipage}[b]{.24\linewidth}
            \centering
            \centerline{\includegraphics[width=\linewidth]{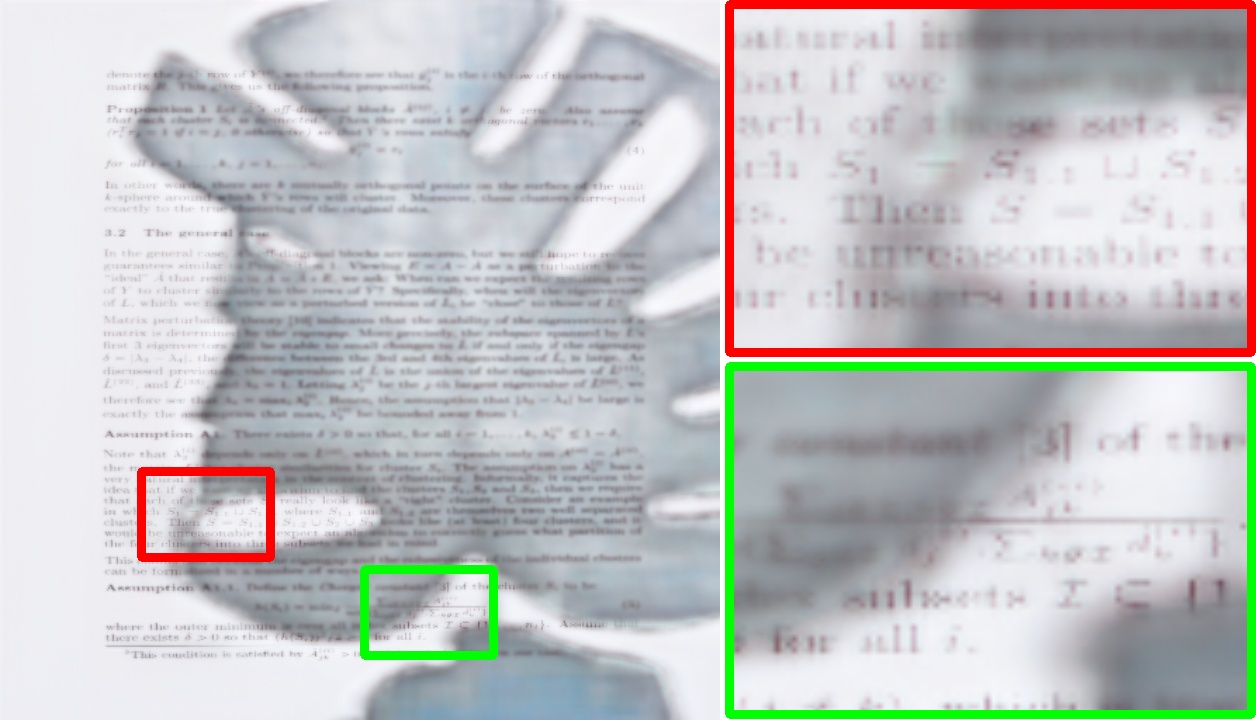}}
            \centerline{ShadowFormer~\cite{guo2023shadowformer}}\medskip
        \end{minipage}
        \hfill
        \begin{minipage}[b]{.24\linewidth}
            \centering
            \centerline{\includegraphics[width=\linewidth]{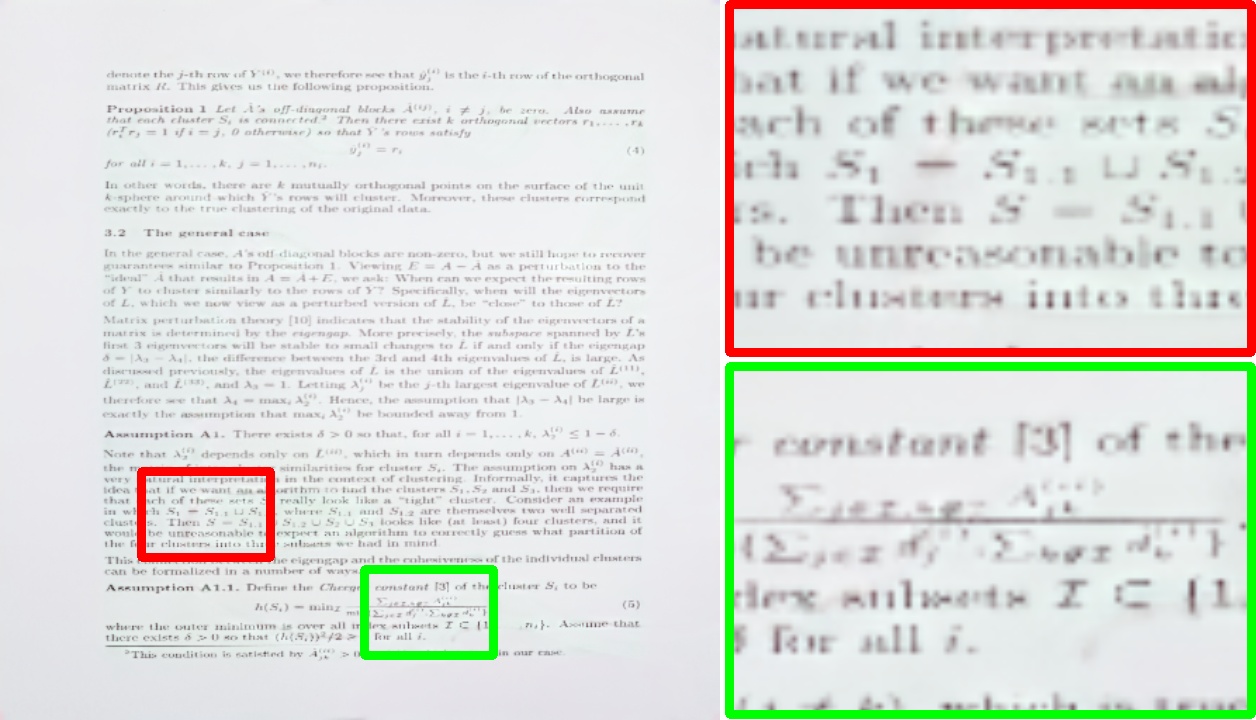}}
            \centerline{Ours}\medskip
        \end{minipage}
        \hfill
        \begin{minipage}[b]{.24\linewidth}
            \centering
            \centerline{\includegraphics[width=\linewidth]{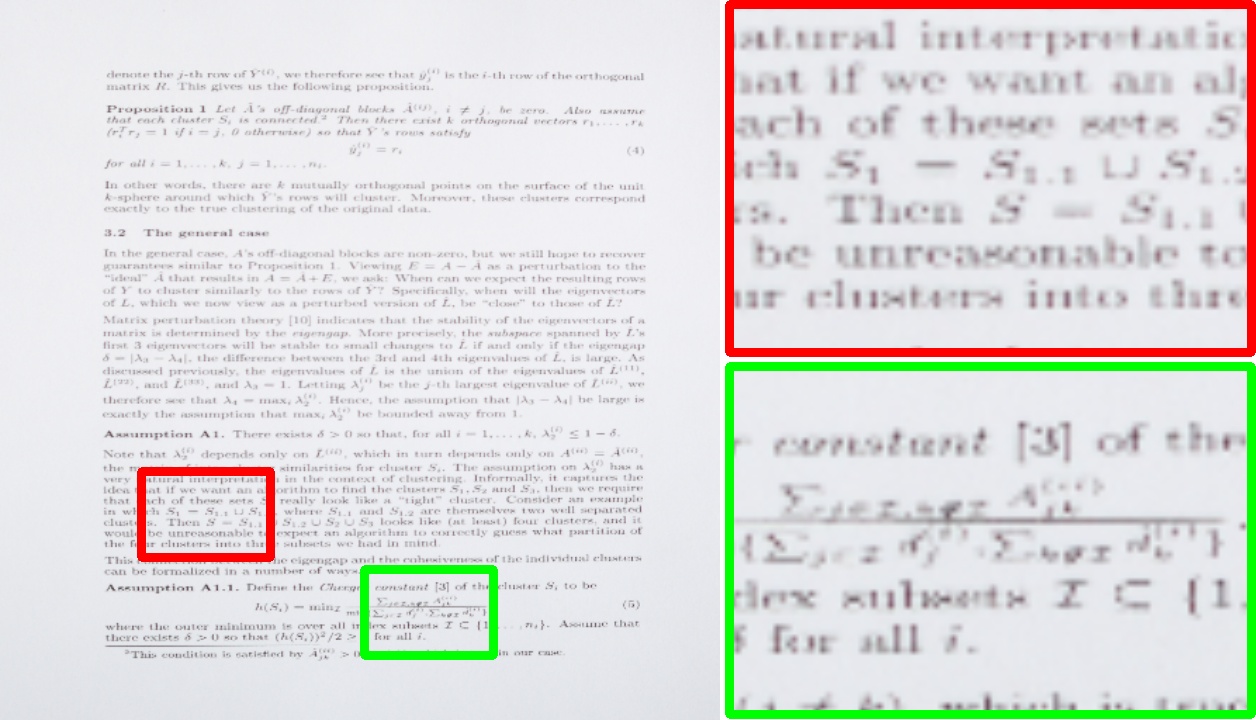}}
            \centerline{Target}\medskip
        \end{minipage}
    \end{minipage}
    \vspace{-2em}
    \caption{Qualitative results of the methods comparison in low resolution on SD7K dataset.}
    \label{fig:lowres}
\end{figure*}

\begin{figure*}[ht]
    \begin{minipage}[b]{1.0\linewidth}
        \begin{minipage}[b]{.24\linewidth}
            \centering
            \centerline{\includegraphics[width=\linewidth]{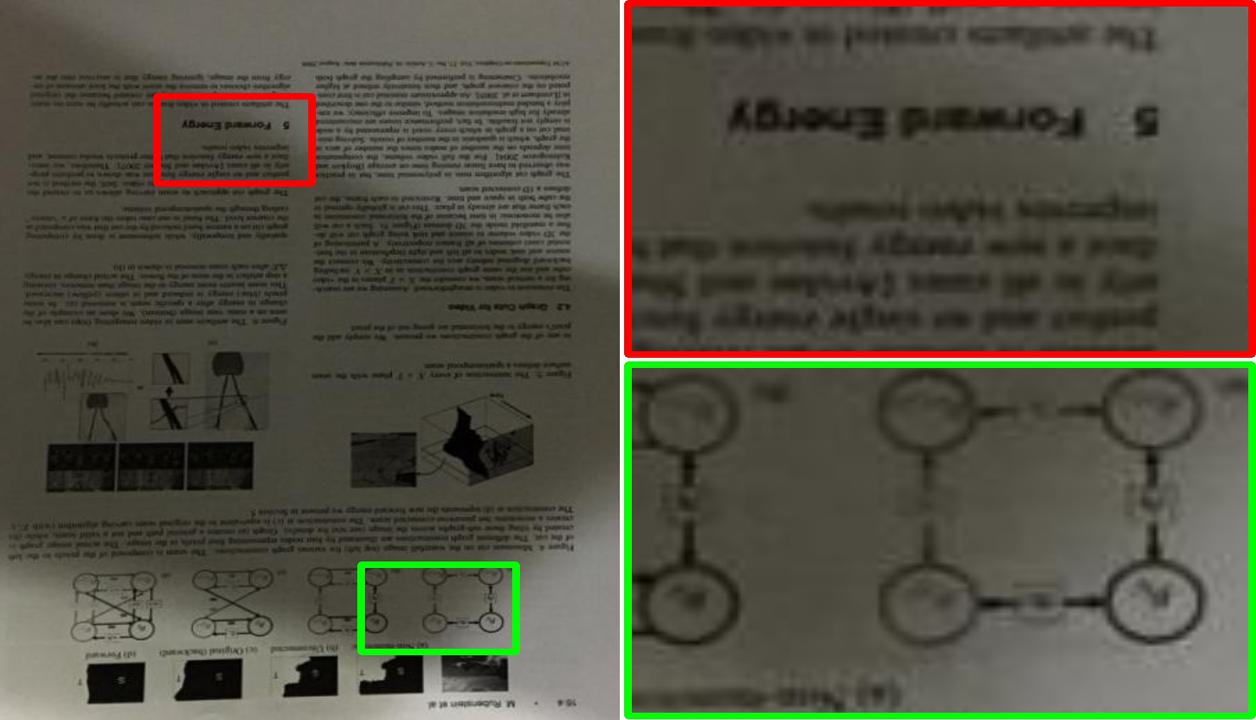}}
            \centerline{Input}\medskip
        \end{minipage}
        \hfill
        \begin{minipage}[b]{.24\linewidth}
            \centering
            \centerline{\includegraphics[width=\linewidth]{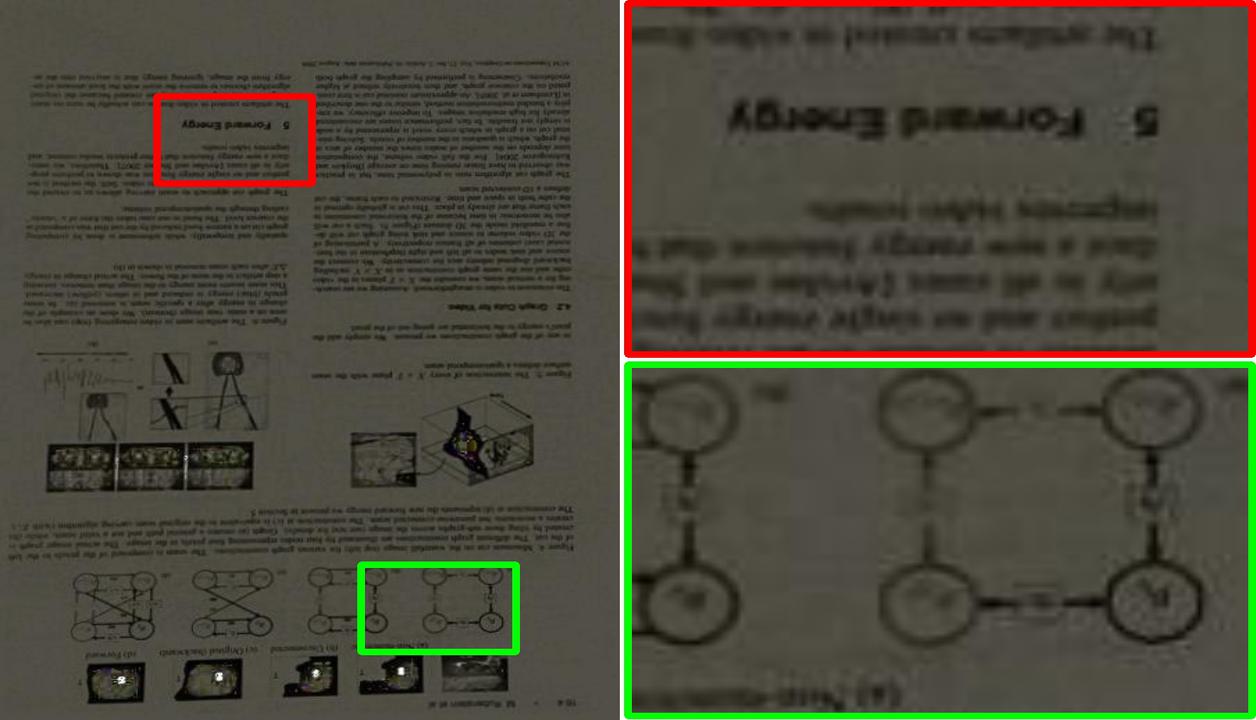
            }}
            \centerline{Wang \etal~\cite{Wang2020ShadowRO}}\medskip
        \end{minipage}
        \hfill
        \begin{minipage}[b]{.24\linewidth}
            \centering
            \centerline{\includegraphics[width=\linewidth]{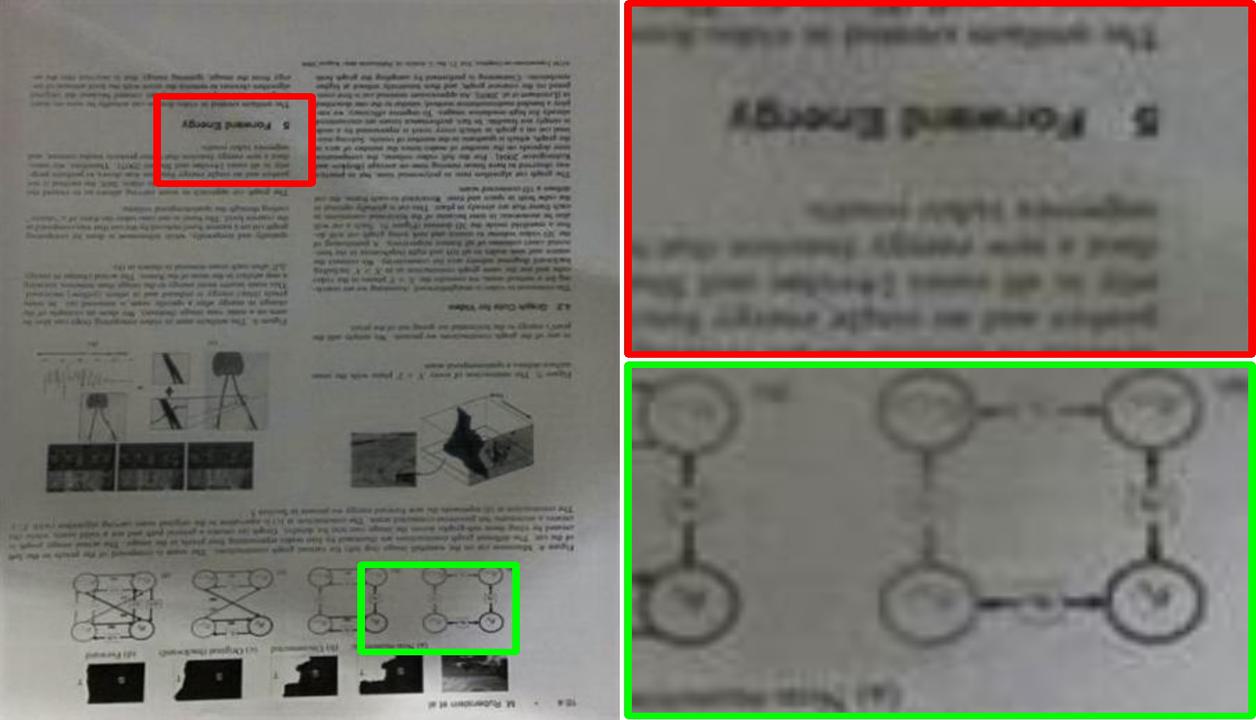}}
            \centerline{SP+M+I Net~\cite{le2021physics}}\medskip
        \end{minipage}
        \hfill
        \begin{minipage}[b]{.24\linewidth}
            \centering
            \centerline{\includegraphics[width=\linewidth]{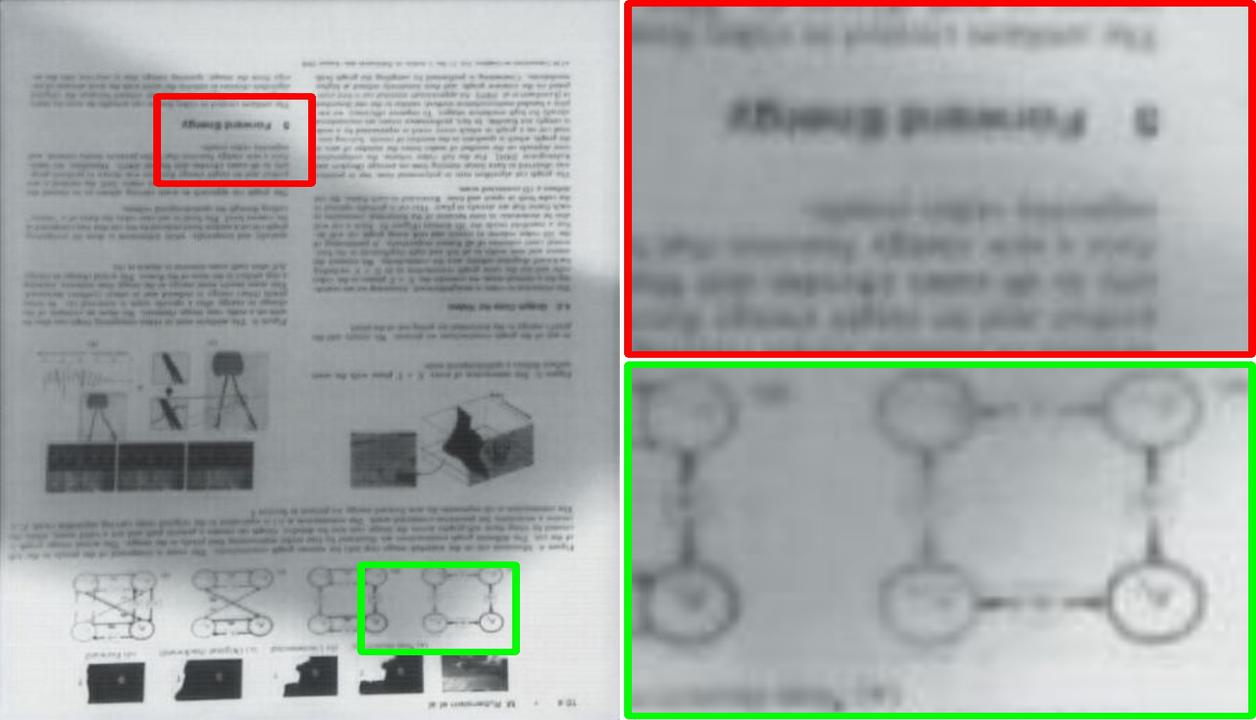}}
            \centerline{BEDSR-Net~\cite{lin2020bedsr}}\medskip
        \end{minipage}
    \end{minipage}
    \begin{minipage}[b]{1.0\linewidth}
        \begin{minipage}[b]{.24\linewidth}
            \centering
            \centerline{\includegraphics[width=\linewidth]{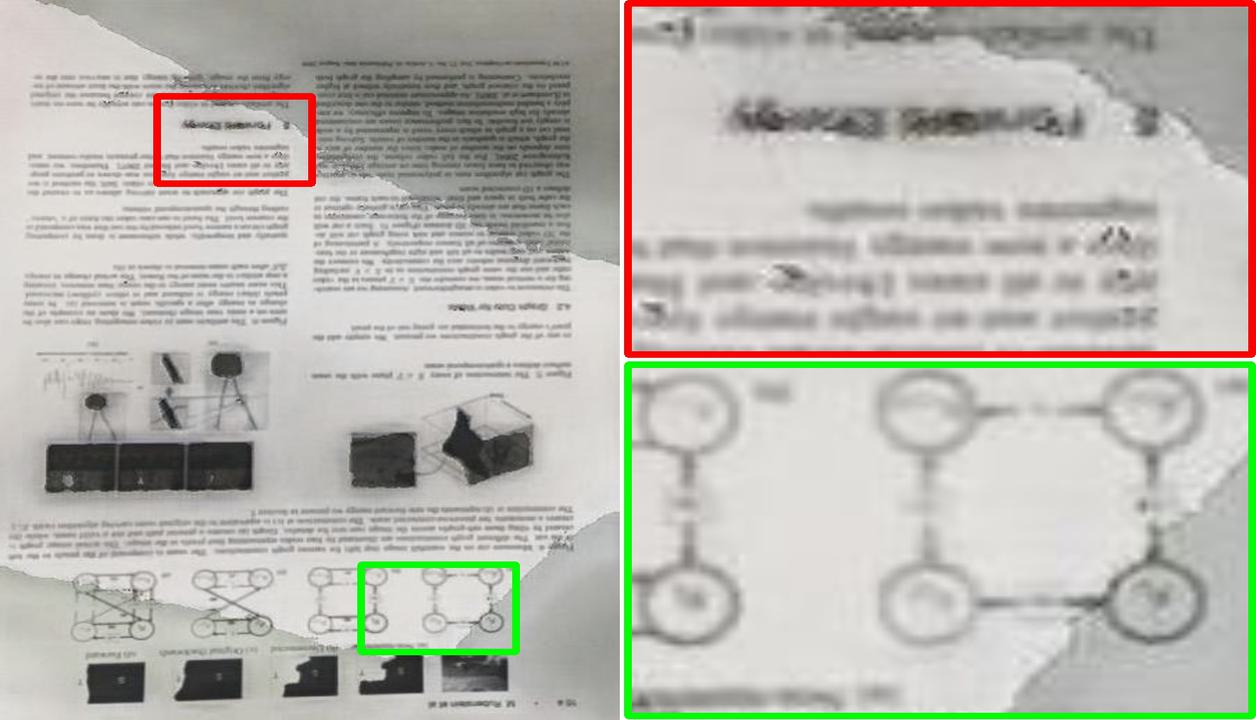}}
            \centerline{SG-ShadowNet~\cite{wan2022style}}\medskip
        \end{minipage}
        \hfill
        \begin{minipage}[b]{.24\linewidth}
            \centering
            \centerline{\includegraphics[width=\linewidth]{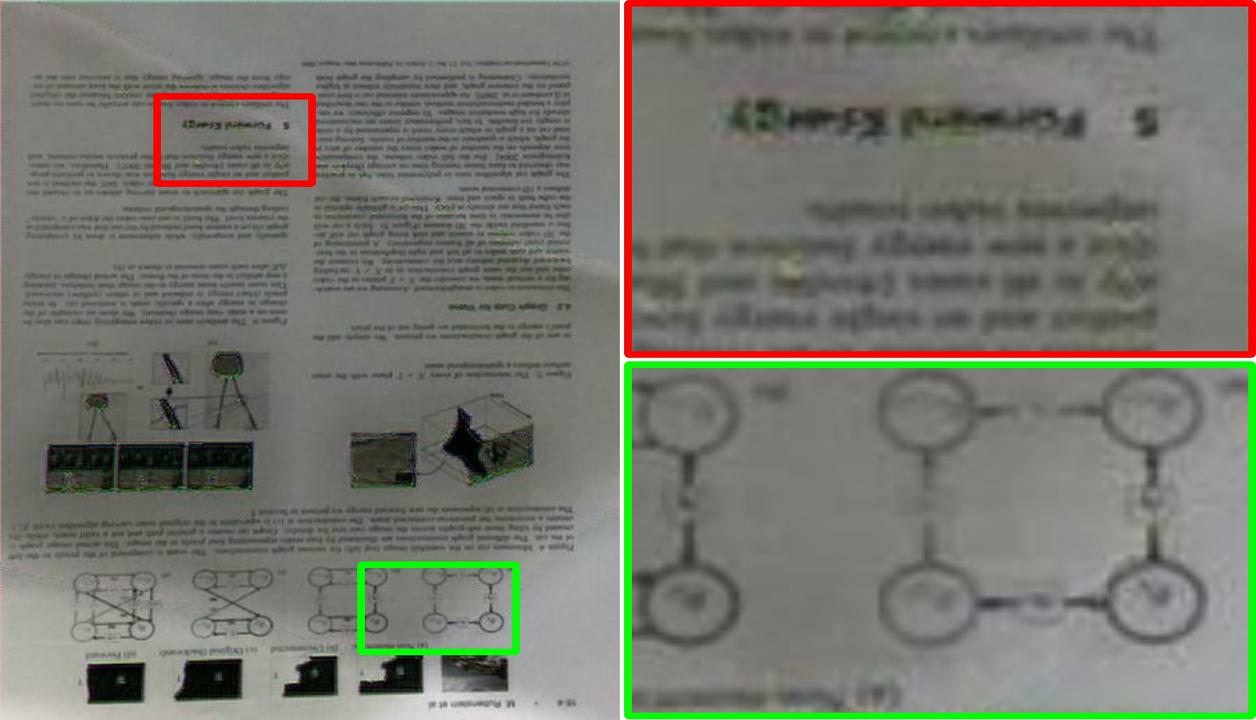}}
            \centerline{ShadowFormer~\cite{guo2023shadowformer}}\medskip
        \end{minipage}
        \hfill
        \begin{minipage}[b]{.24\linewidth}
            \centering
            \centerline{\includegraphics[width=\linewidth]{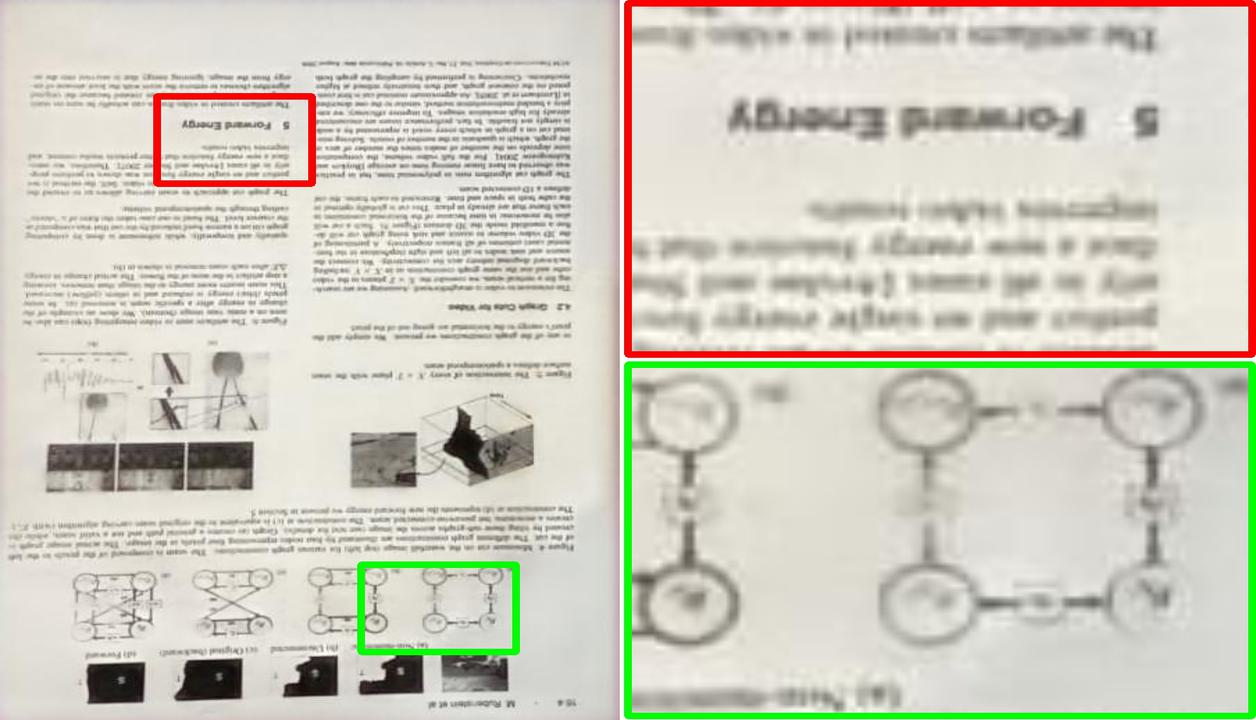}}
            \centerline{Ours}\medskip
        \end{minipage}
        \hfill
        \begin{minipage}[b]{.24\linewidth}
            \centering
            \centerline{\includegraphics[width=\linewidth]{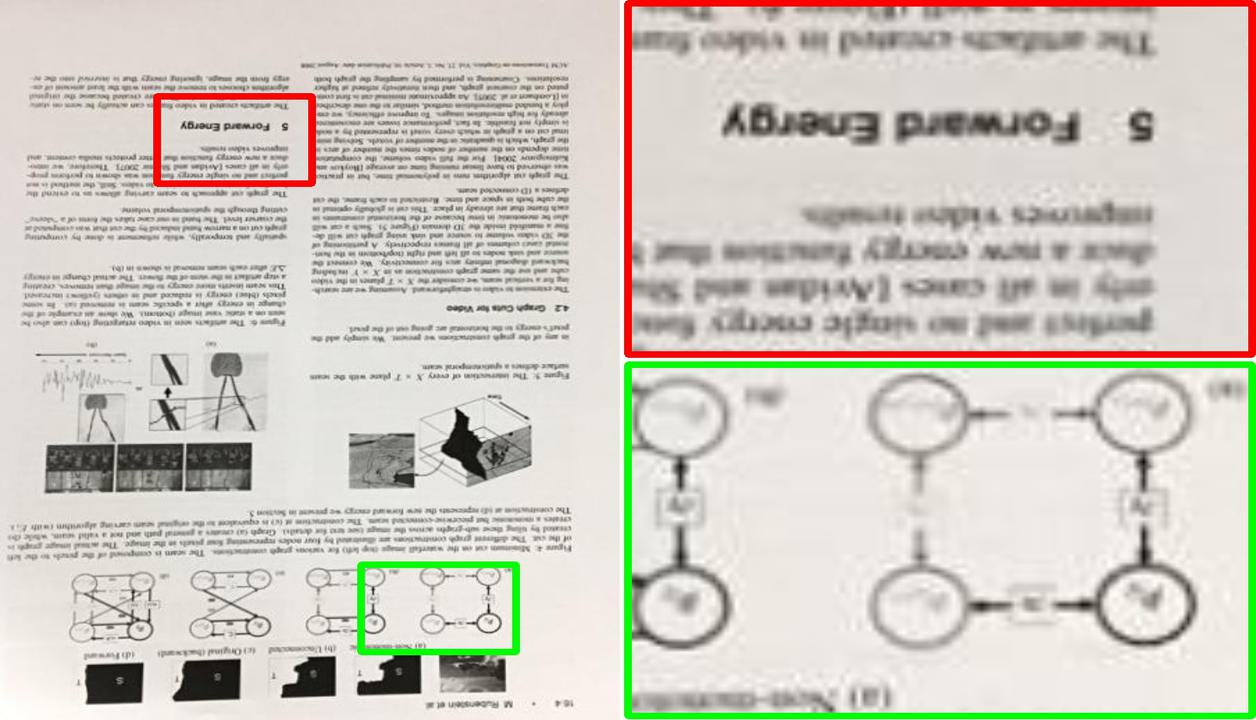}}
            \centerline{Target}\medskip
        \end{minipage}
    \end{minipage}
    \vspace{-2em}
    \caption{Qualitative results of the methods comparison in low resolution on Jung dataset.}
    \label{fig:lowres0}
\end{figure*}

 We give the quantitative evaluation in Table~\ref{table:low} and the visual results are available in Figure~\ref{fig:lowres} and Figure~\ref{fig:lowres0}.
As shown in Table~\ref{table:low}, our proposed FSENet outperforms all other methods under the low-resolution data setting. In the Figure~\ref{fig:lowres} and Figure~\ref{fig:lowres0}, we can clearly observe that Wang \etal~\cite{Wang2020ShadowRO}, SP+M+I Net~\cite{le2021physics}, SG-ShadowNet~\cite{wan2022style} and ShadowFormer~\cite{guo2023shadowformer} exhibit the phenomenon of incomplete shadow removal. Meanwhile, \cite{Wang2020ShadowRO} shows a large difference from the target in white balance. Despite BEDSR-Net~\cite{lin2020bedsr} performing relatively well in both white balance and shadow removal, a close examination of the finer details reveals that BEDSR-Net yields blurry results with missing texture details. Simultaneously, BEDSR-Net fails in Figure~\ref{fig:lowres0} in terms of shadow removal. In comparison, our result achieves the best performance in white balance, shadow removal, and detail restoration.

\section{More High-Resolution Results on SD7K}
\begin{figure*}[ht]
    \begin{minipage}[b]{1.0\linewidth}
        \begin{minipage}[b]{.19\linewidth}
            \centering
            \centerline{\includegraphics[width=\linewidth, height=1.6\linewidth]{figs/6_exp/highres0/mag_input.jpg}}
        \end{minipage}
        \hfill
        \begin{minipage}[b]{.19\linewidth}
            \centering
            \centerline{\includegraphics[width=\linewidth, height=1.6\linewidth]{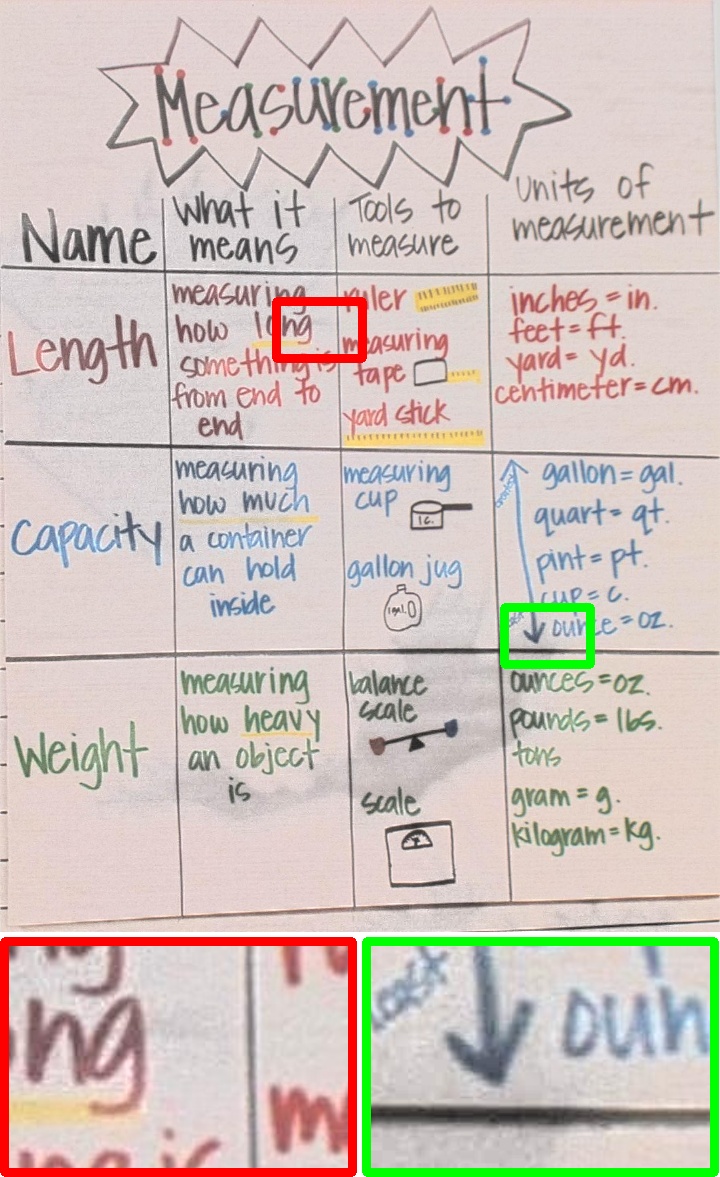
            }}
        \end{minipage}
        \hfill
        \begin{minipage}[b]{.19\linewidth}
            \centering
            \centerline{\includegraphics[width=\linewidth, height=1.6\linewidth]{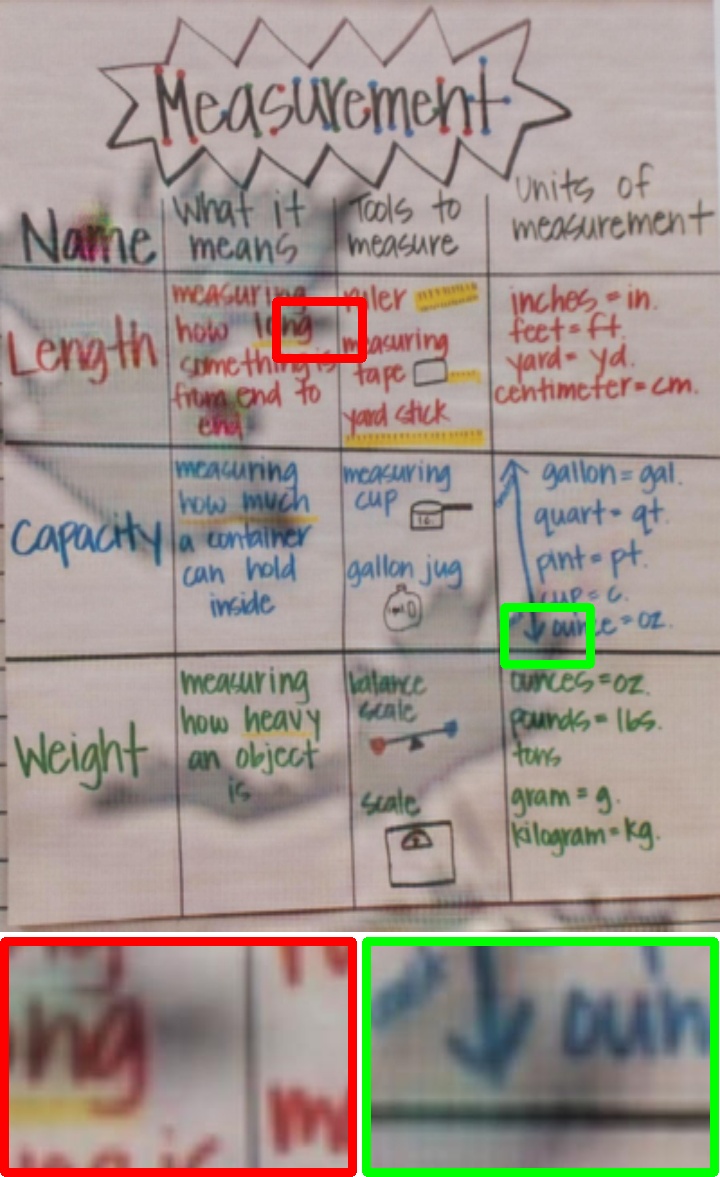}}
        \end{minipage}
        \hfill
        \begin{minipage}[b]{.19\linewidth}
            \centering
            \centerline{\includegraphics[width=\linewidth, height=1.6\linewidth]{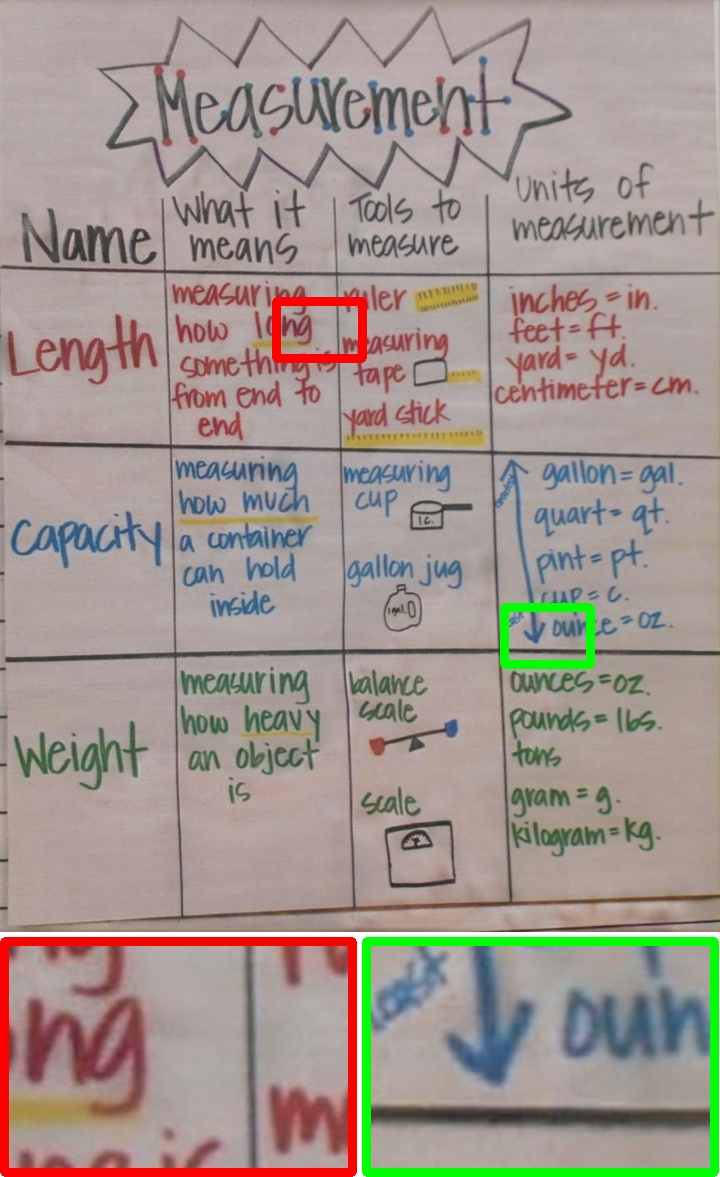}}
        \end{minipage}
        \hfill
        \begin{minipage}[b]{.19\linewidth}
            \centering
            \centerline{\includegraphics[width=\linewidth, height=1.6\linewidth]{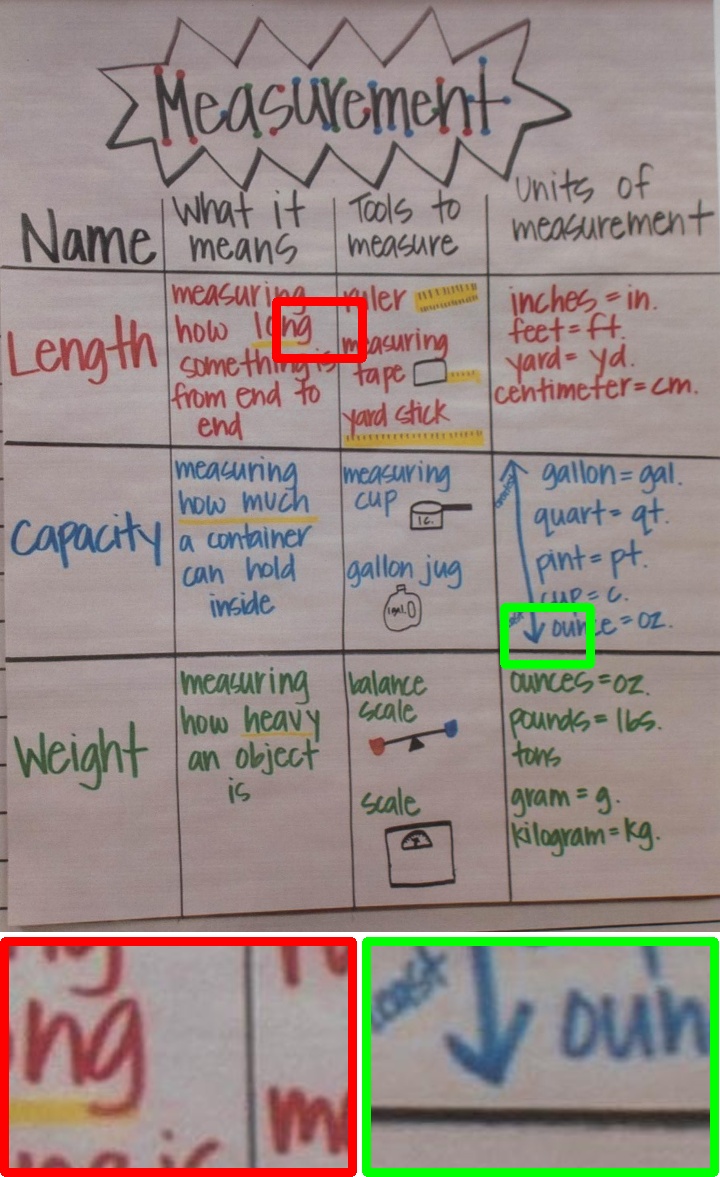}}
        \end{minipage}

        \vspace{0.15cm}
        
        \begin{minipage}[b]{.19\linewidth}
            \centering
            \centerline{\includegraphics[width=\linewidth, height=1.6\linewidth]{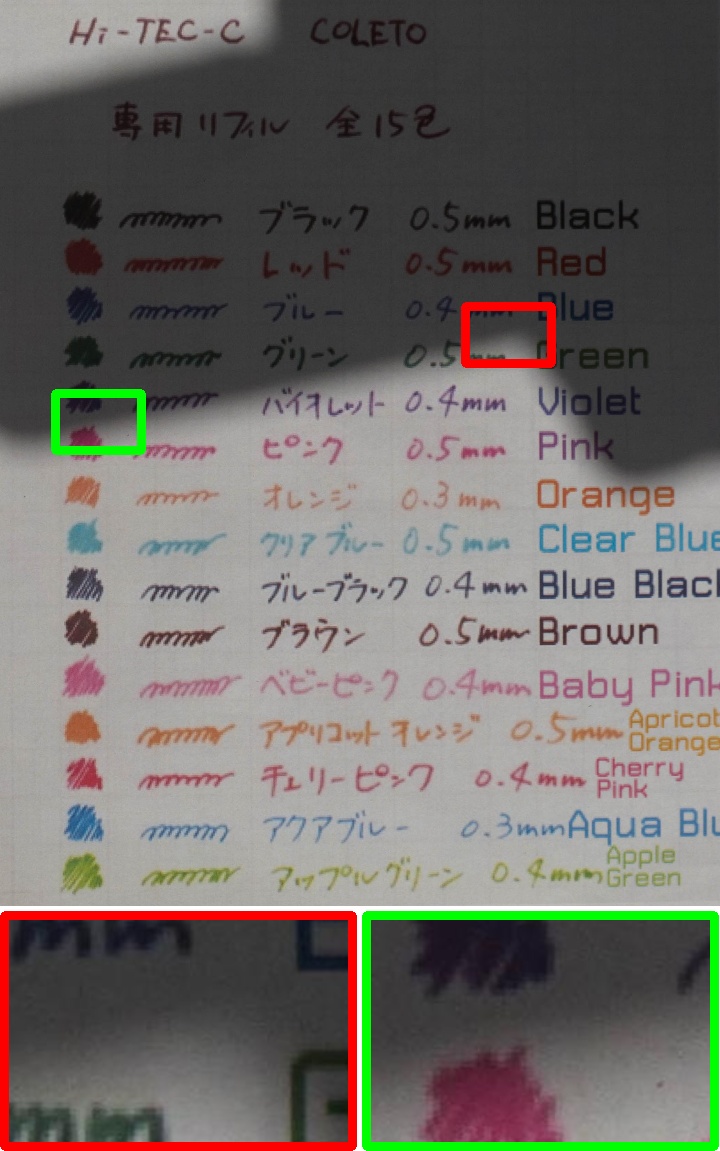}}
        \end{minipage}
        \hfill
        \begin{minipage}[b]{.19\linewidth}
            \centering
            \centerline{\includegraphics[width=\linewidth, height=1.6\linewidth]{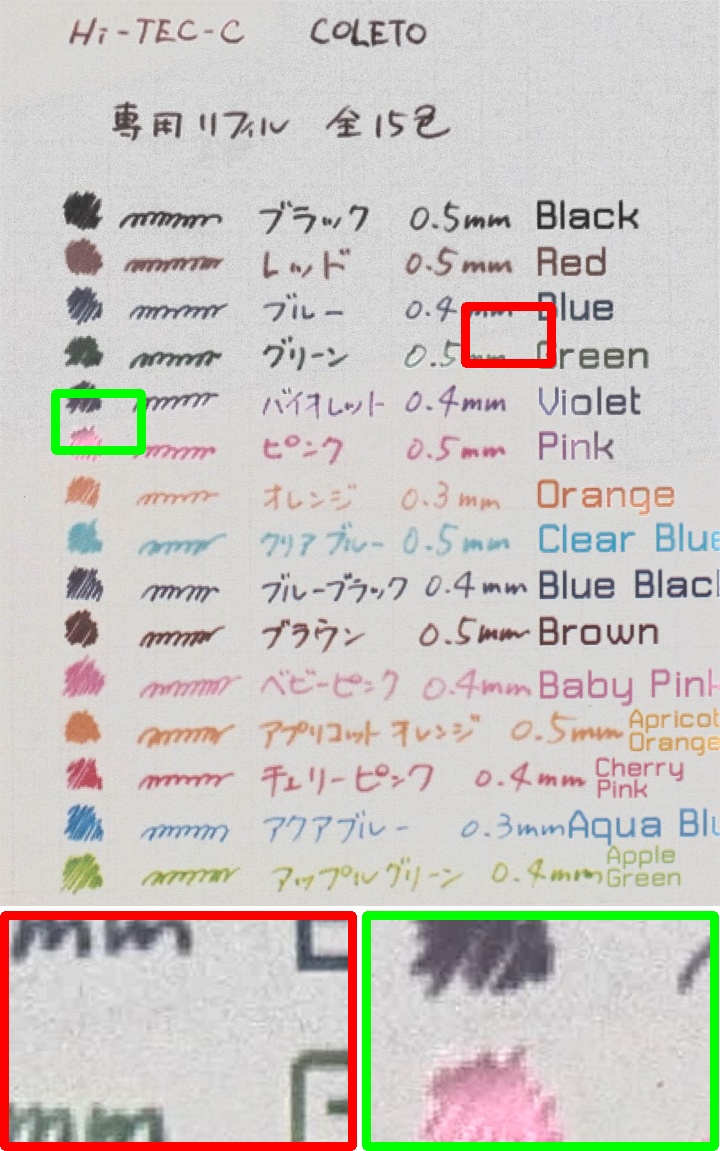
            }}
        \end{minipage}
        \hfill
        \begin{minipage}[b]{.19\linewidth}
            \centering
            \centerline{\includegraphics[width=\linewidth, height=1.6\linewidth]{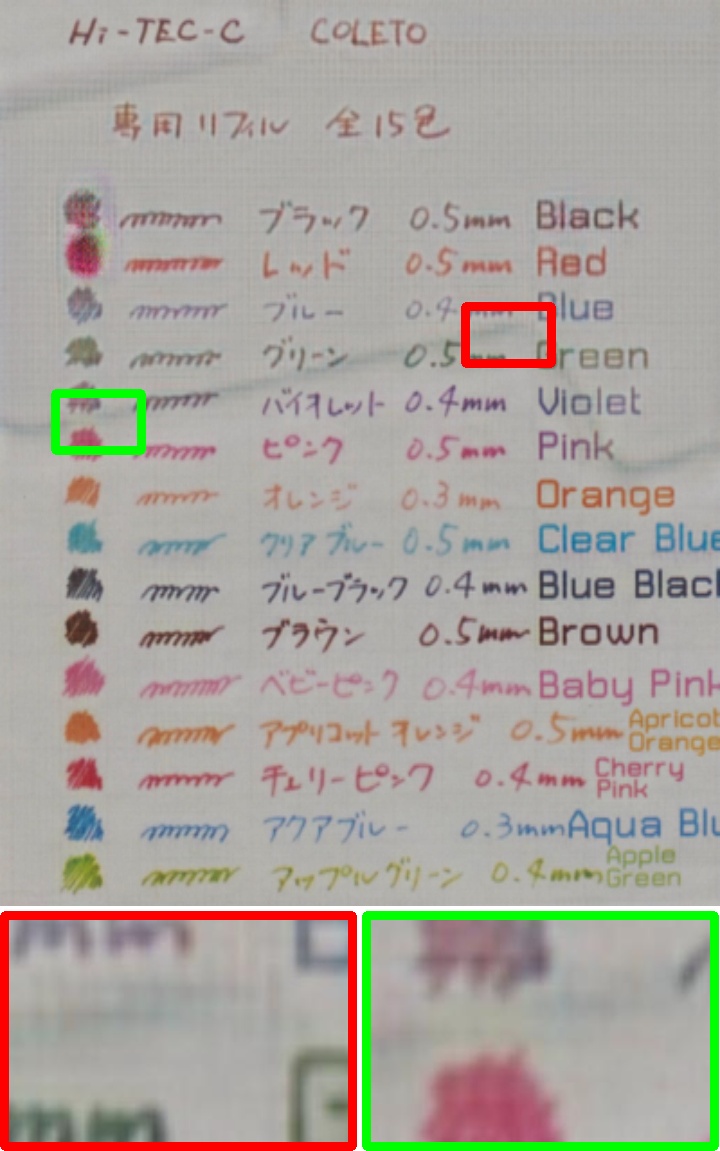}}
        \end{minipage}
        \hfill
        \begin{minipage}[b]{.19\linewidth}
            \centering
            \centerline{\includegraphics[width=\linewidth, height=1.6\linewidth]{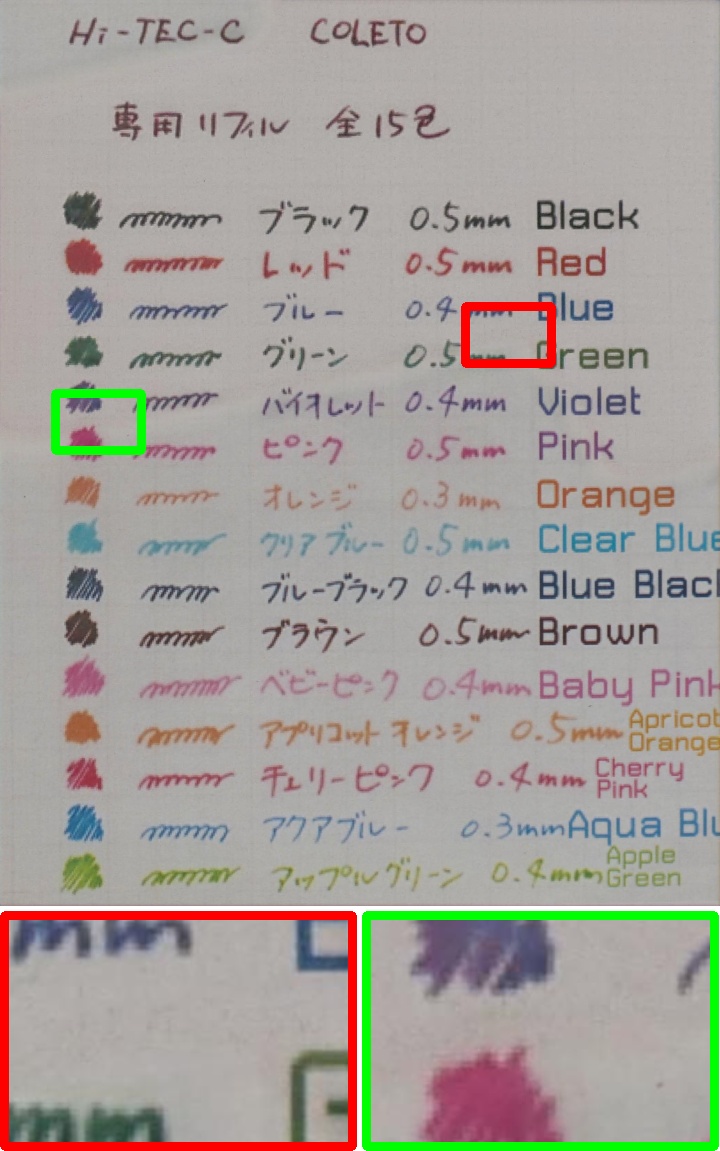}}
        \end{minipage}
        \hfill
        \begin{minipage}[b]{.19\linewidth}
            \centering
            \centerline{\includegraphics[width=\linewidth, height=1.6\linewidth]{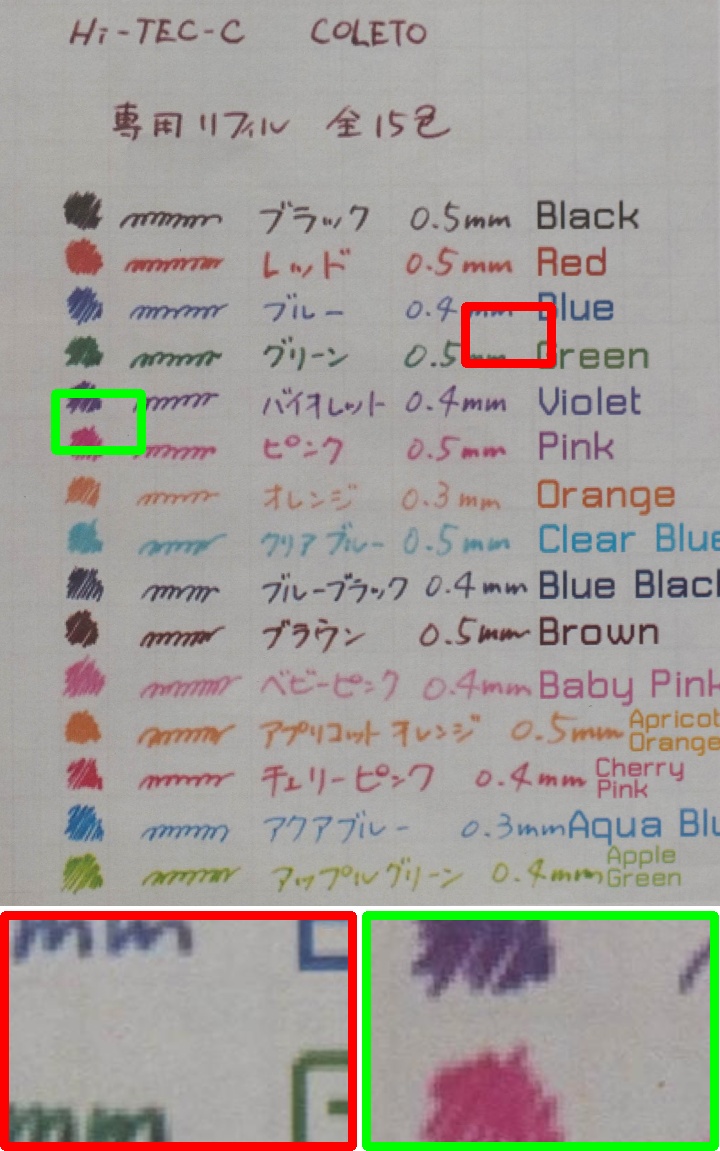}}
        \end{minipage}

        \vspace{0.15cm}
        
        \begin{minipage}[b]{.19\linewidth}
            \centering
            \centerline{\includegraphics[width=\linewidth, height=1.6\linewidth]{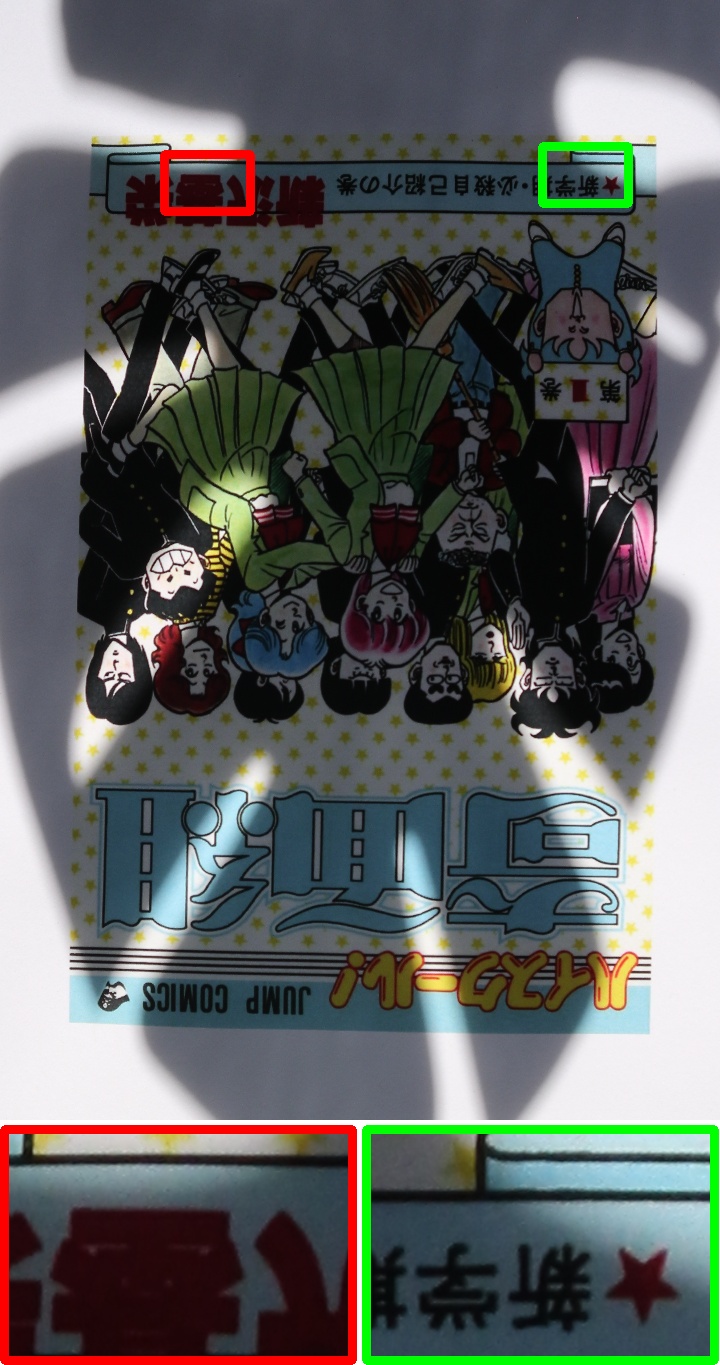}}
        \end{minipage}
        \hfill
        \begin{minipage}[b]{.19\linewidth}
            \centering
            \centerline{\includegraphics[width=\linewidth, height=1.6\linewidth]{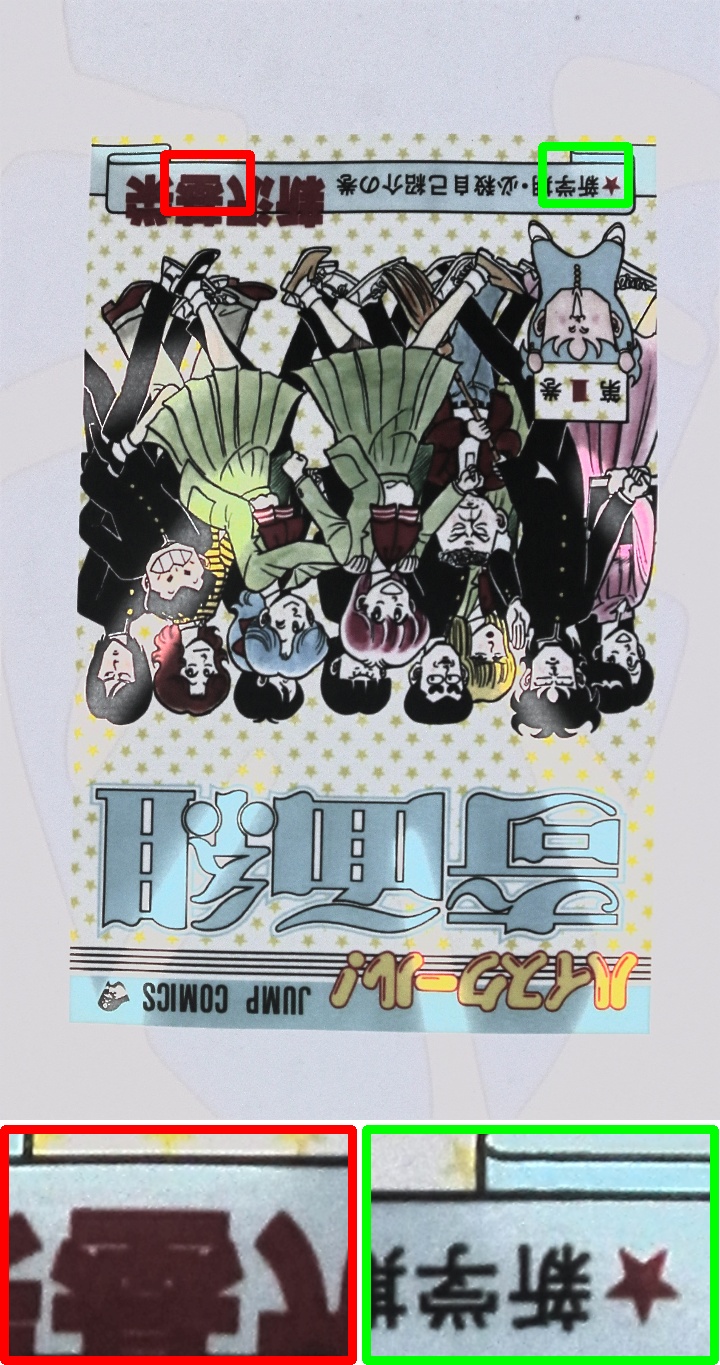
            }}
        \end{minipage}
        \hfill
        \begin{minipage}[b]{.19\linewidth}
            \centering
            \centerline{\includegraphics[width=\linewidth, height=1.6\linewidth]{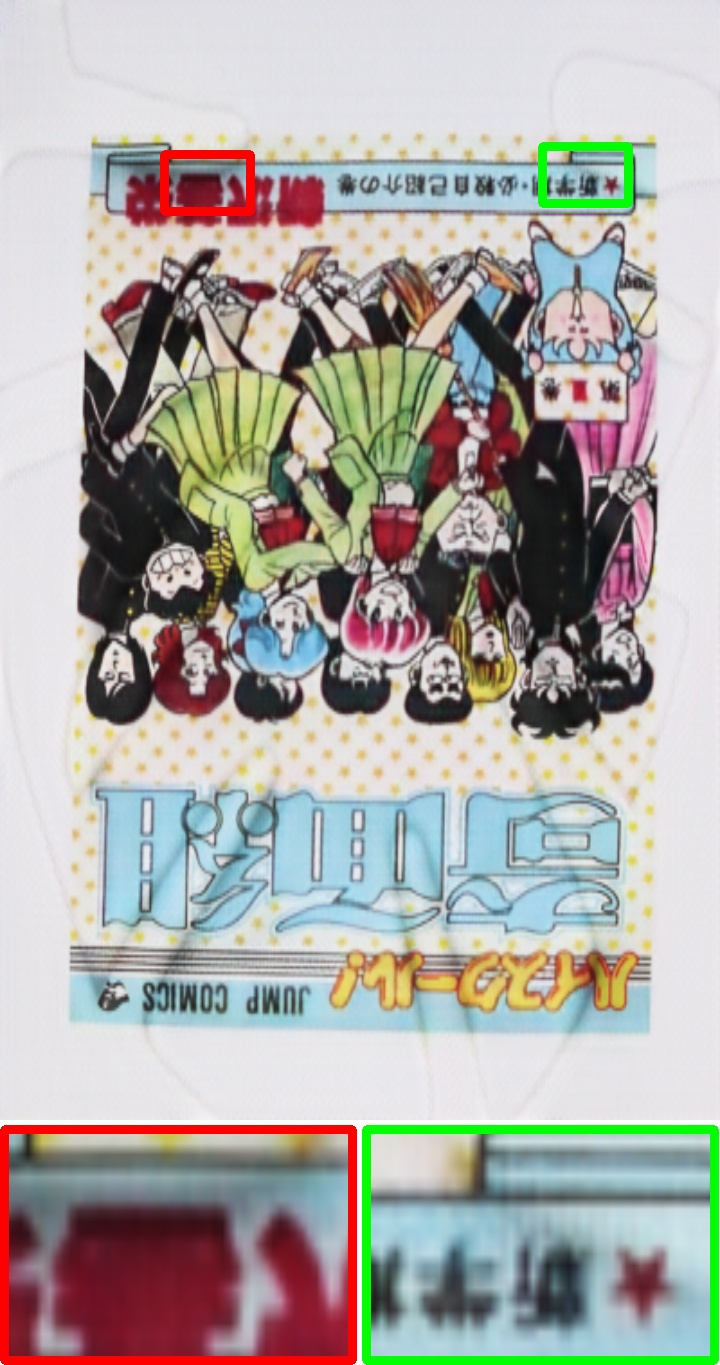}}
        \end{minipage}
        \hfill
        \begin{minipage}[b]{.19\linewidth}
            \centering
            \centerline{\includegraphics[width=\linewidth, height=1.6\linewidth]{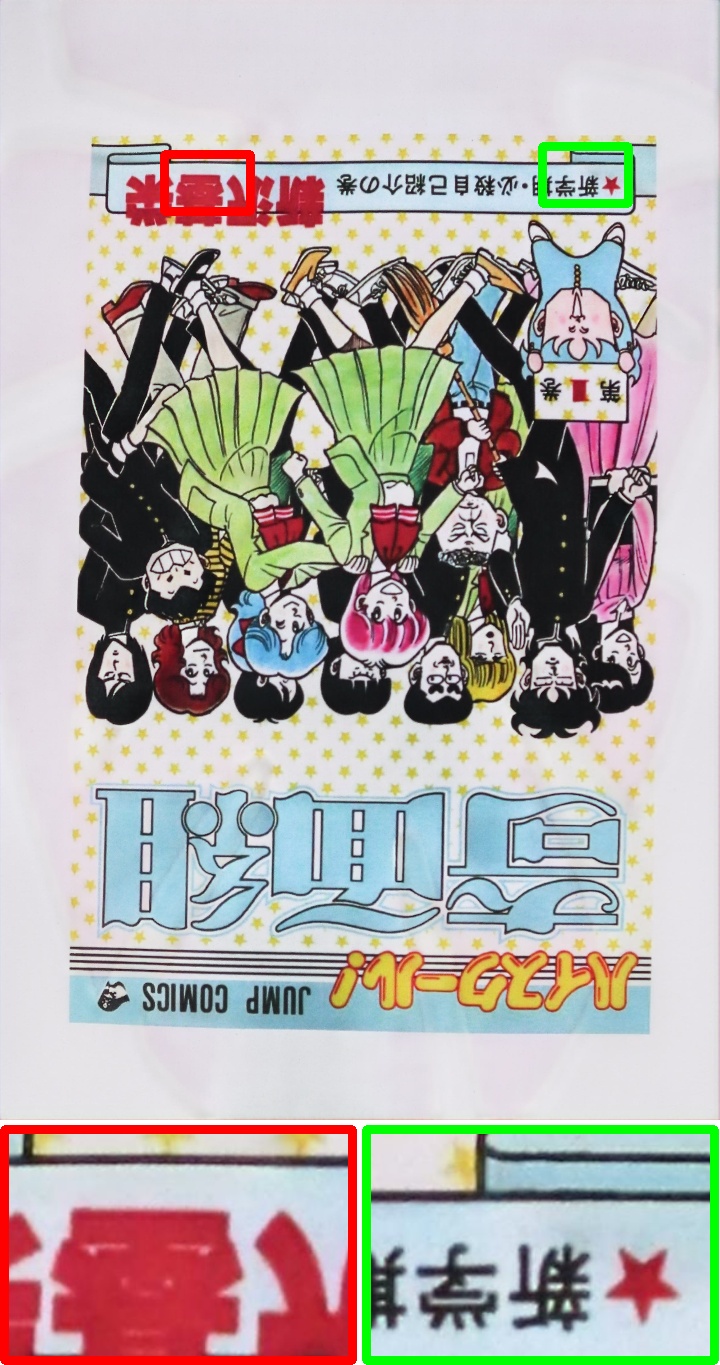}}
        \end{minipage}
        \hfill
        \begin{minipage}[b]{.19\linewidth}
            \centering
            \centerline{\includegraphics[width=\linewidth, height=1.6\linewidth]{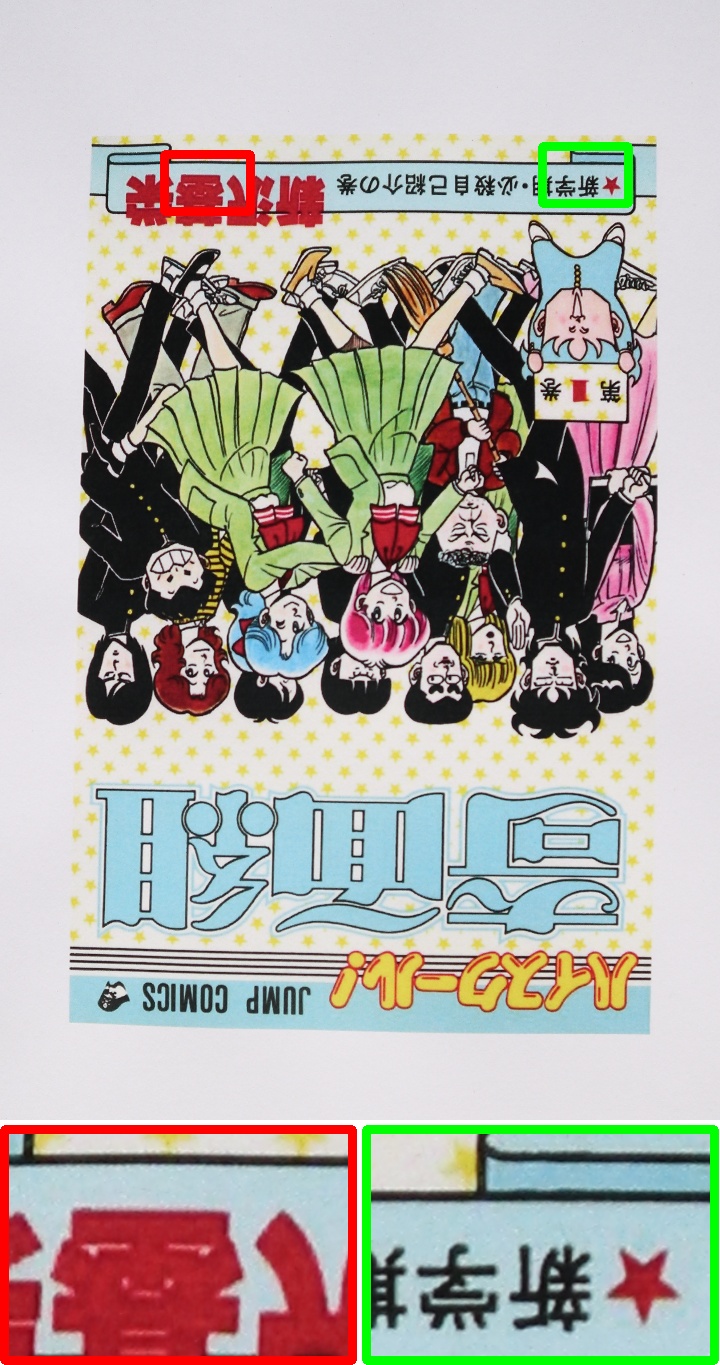}}
        \end{minipage}

        \vspace{0.15cm}
        
        \begin{minipage}[b]{.19\linewidth}
            \centering
            \centerline{\includegraphics[width=\linewidth, height=1.6\linewidth]{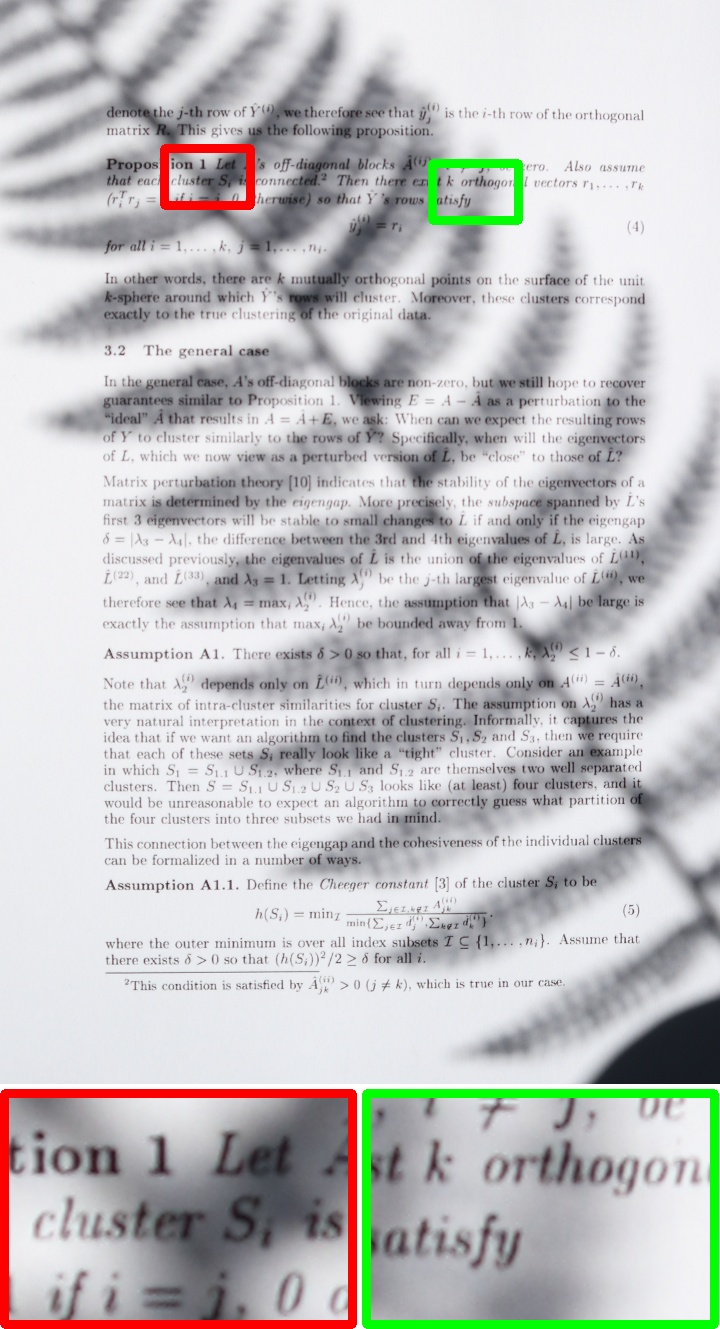}}
            \centerline{Input}\medskip
        \end{minipage}
        \hfill
        \begin{minipage}[b]{.19\linewidth}
            \centering
            \centerline{\includegraphics[width=\linewidth, height=1.6\linewidth]{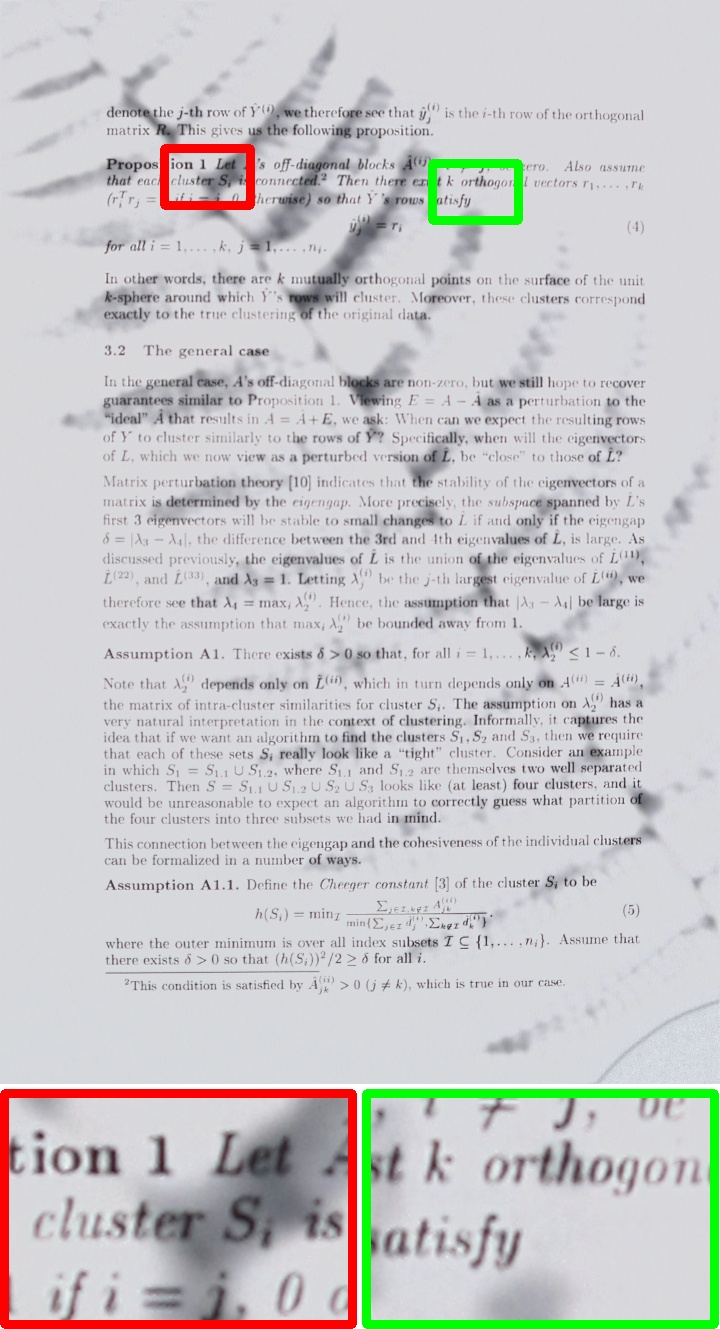
            }}
            \centerline{Jung \etal~\cite{jung2018water}}\medskip
        \end{minipage}
        \hfill
        \begin{minipage}[b]{.19\linewidth}
            \centering
            \centerline{\includegraphics[width=\linewidth, height=1.6\linewidth]{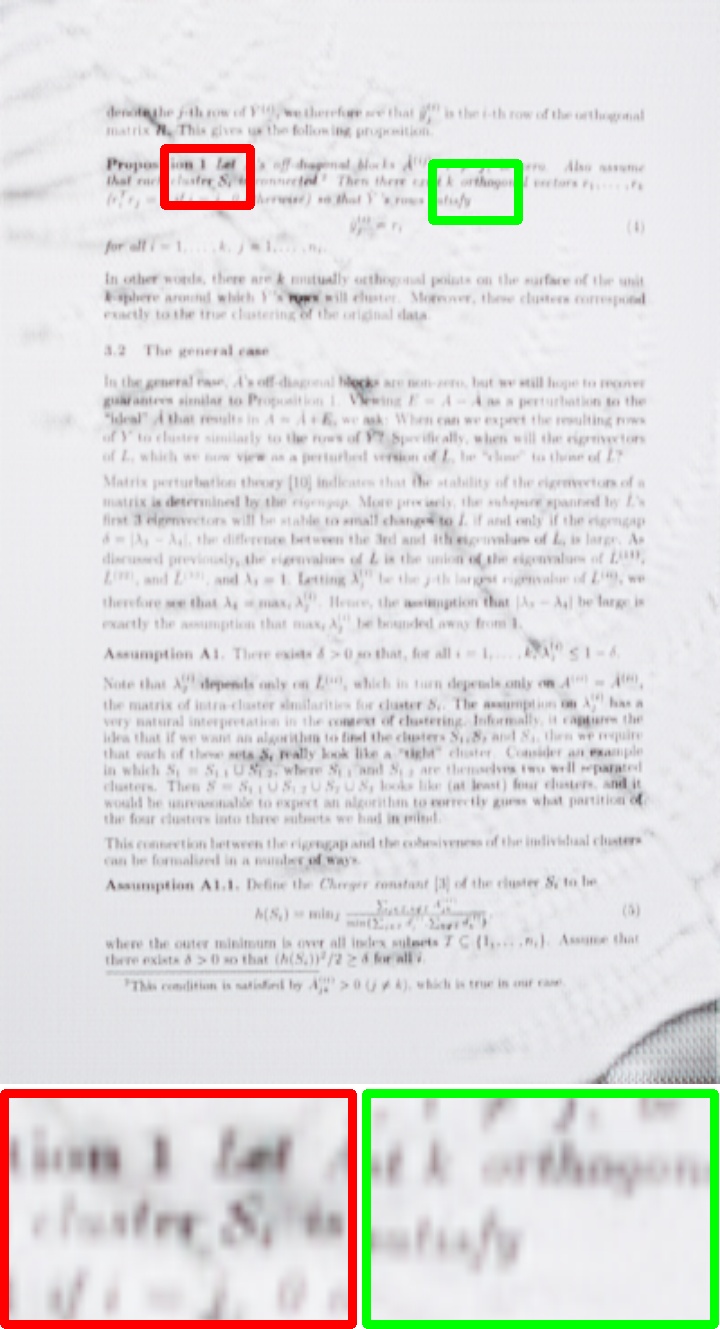}}
            \centerline{MaskShadowGAN~\cite{hu2019mask}}\medskip
        \end{minipage}
        \hfill
        \begin{minipage}[b]{.19\linewidth}
            \centering
            \centerline{\includegraphics[width=\linewidth, height=1.6\linewidth]{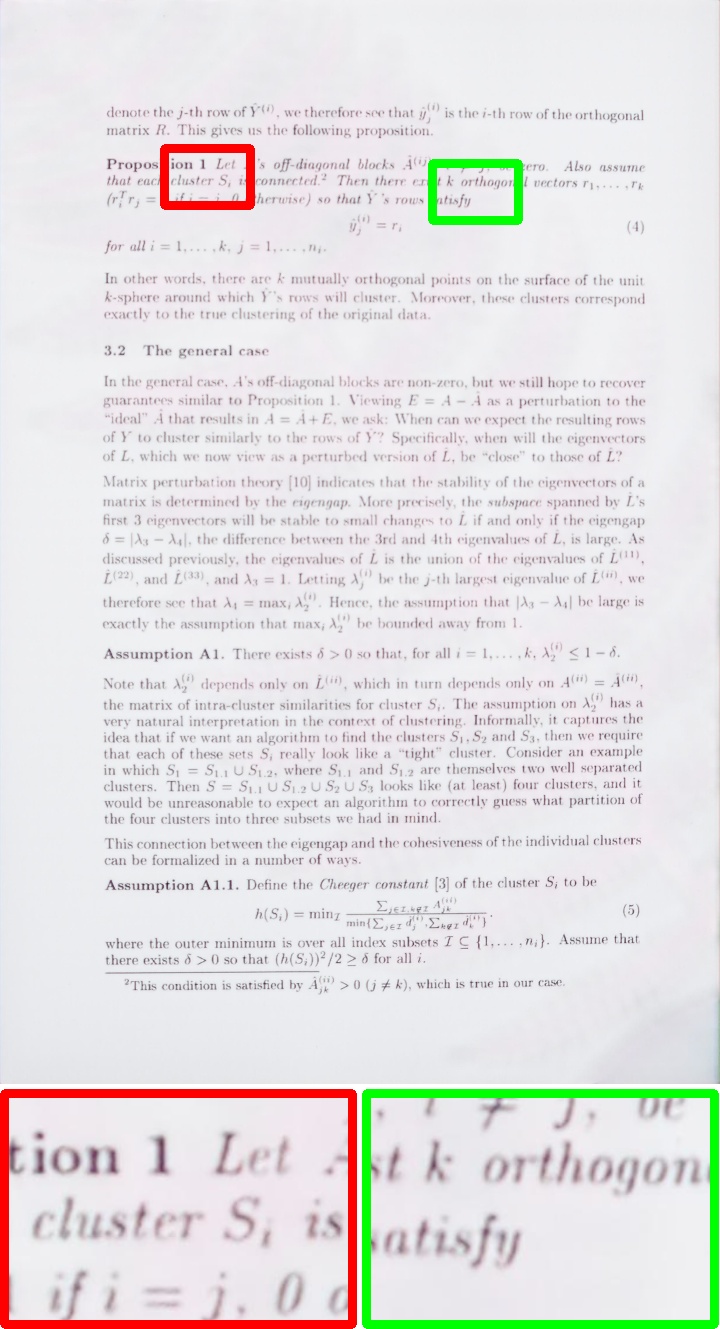}}
            \centerline{Ours}\medskip
        \end{minipage}
        \hfill
        \begin{minipage}[b]{.19\linewidth}
            \centering
            \centerline{\includegraphics[width=\linewidth, height=1.6\linewidth]{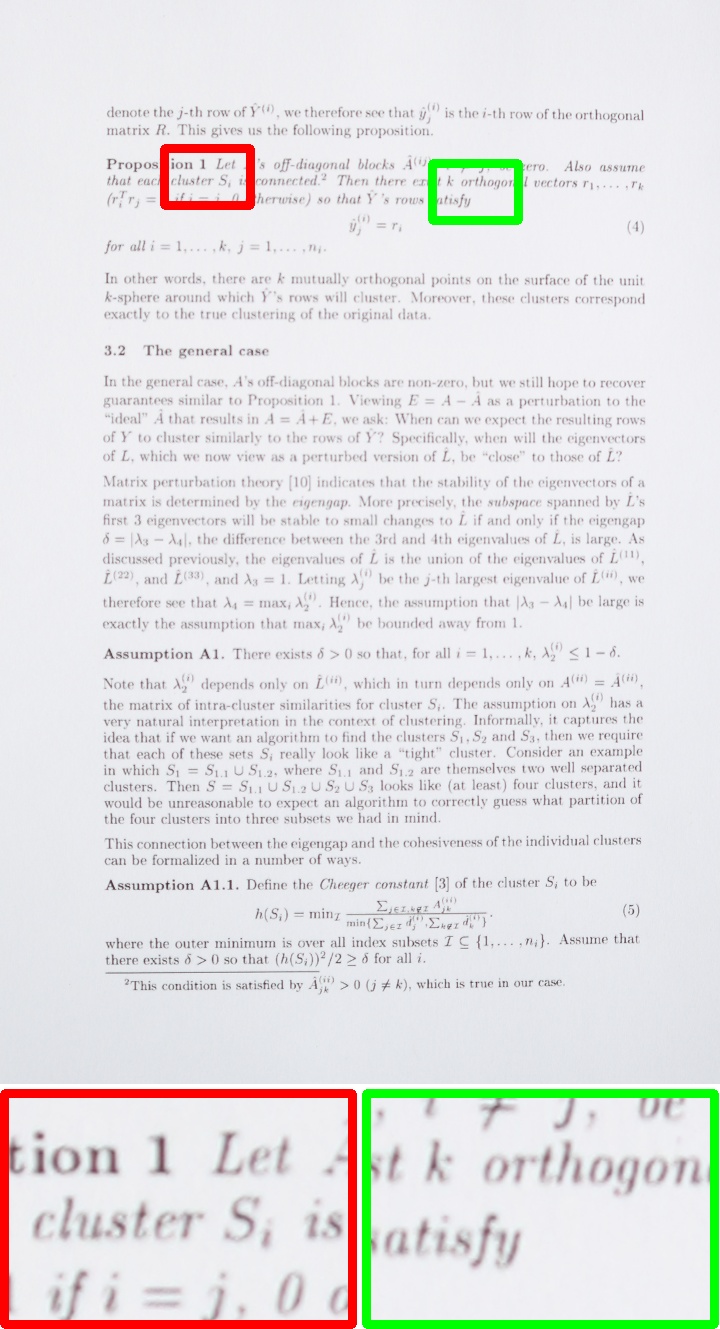}}
            \centerline{Target}\medskip
        \end{minipage}
    \end{minipage}
    \caption{Qualitative results of the methods comparison in high resolution, the first two rows represent the results on Kligler dataset and the last two rows depict the results on SD7K dataset.}
    \label{fig:highres_supp}
\end{figure*}

We give more visual comparisons on the high-resolution inputs in Figure~\ref{fig:highres_supp}. The results in the first two rows come from Kligler dataset, and the results in the last two rows come from SD7K. It can be observed that the shadow removal results from Jung \etal~\cite{jung2018water} are unstable, and the white balance often differs from the target. Meanwhile, shadow removal with MaskShadowGAN~\cite{hu2019mask} is always incomplete, with the original shadow regions prone to leaving stains and causing texts to become unreadable. In contrast, our method performs relatively well in both white balancing and shadow removal.

\subsection{The Detailed Structure of DFE and TAA Blocks}
\begin{figure}[t]
    \includegraphics[width=0.5\textwidth]{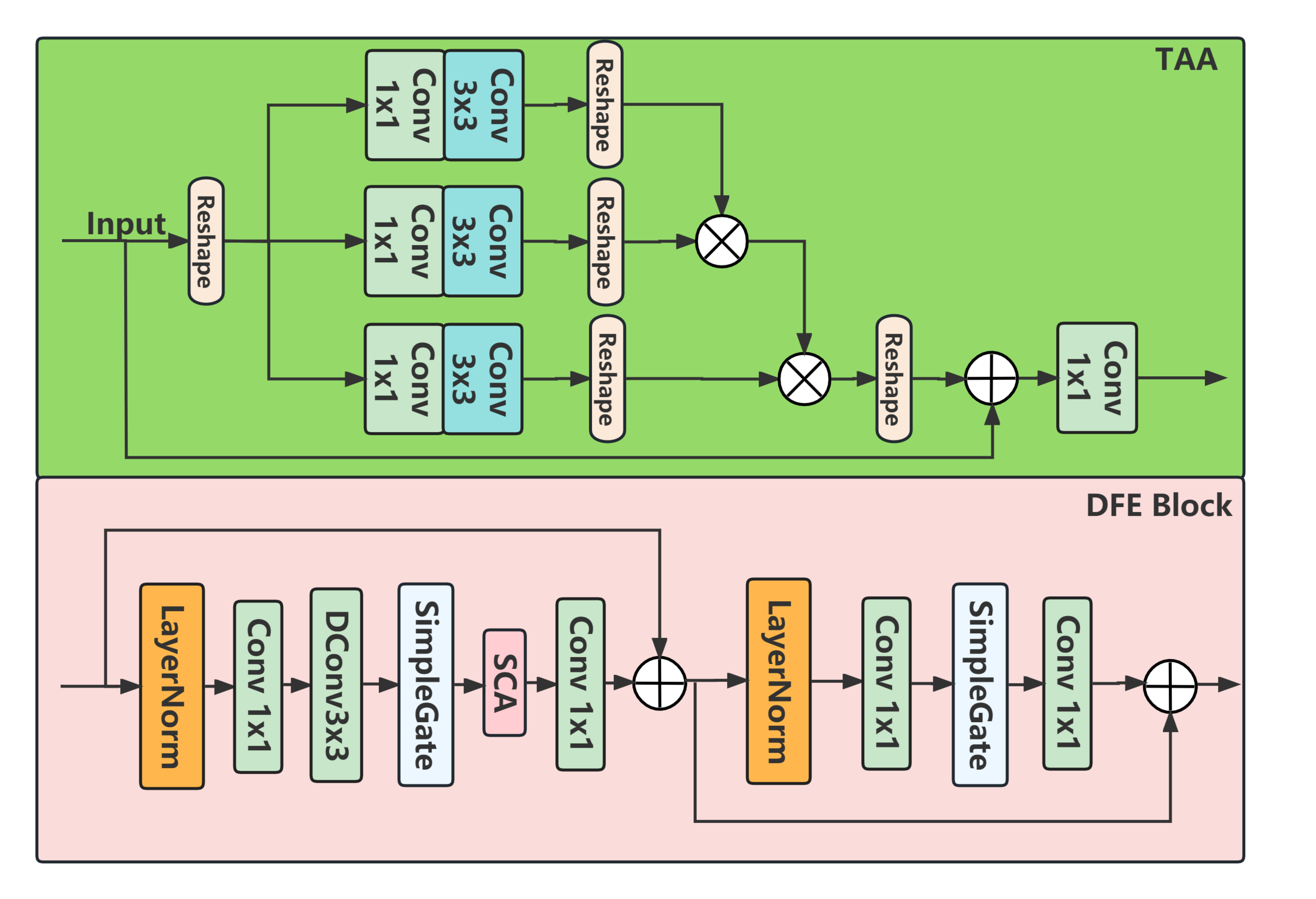}
    \caption{
    The network structure of our DFE and TAA. 
    }
    \label{fig:dfe_taa}
\end{figure}

In Sec. 4 of the main paper, we show the introduction of the TAA and DFE by text description, here, we give a detailed network structure to better understand our method as shown in Figure~\ref{fig:dfe_taa}.

\begin{table*}[ht]
\centering
\begin{tabular}{l|ccc|ccc|ccc}
\toprule
& \multicolumn{3}{c|}{Jung ($512 \times 512$)}                                                                                 & \multicolumn{3}{c|}{Kligler ($512 \times 512$)}                                                                               & \multicolumn{3}{c}{SD7K ($512\times512$)}                                                                       \\ \cline{2-10} 
\multirow{-2}{*}{Method} & \multicolumn{1}{c|}{PSNR$\uparrow$} & \multicolumn{1}{c|}{SSIM$\uparrow$} & RMSE$\downarrow$              & \multicolumn{1}{c|}{PSNR$\uparrow$} & \multicolumn{1}{c|}{SSIM$\uparrow$} & RMSE$\downarrow$              & \multicolumn{1}{c|}{PSNR$\uparrow$} & \multicolumn{1}{c|}{SSIM$\uparrow$} & RMSE$\downarrow$     \\ \hline
Input & 13.01 & 0.82 & 60.85 & 13.26 & 0.80 & 56.73 & 15.95 & 0.89 & 44.09 \\
Wang \etal~\cite{Wang2020ShadowRO} & 11.17 & 0.78 & 73.27 & 15.73 & 0.82 & 44.04 & 15.31 & 0.82 & 47.88 \\
Wang \etal~\cite{wang2019effective} & 9.11 & 0.71 & 90.99 & 15.38 & 0.72 & 48.03 & 13.32 & 0.68 & 67.48 \\
Shah \etal~\cite{Shah2018AnIA} & 14.69 & 0.80 & 47.97 & 8.36 & 0.70 & 97.86 & 9.89 & 0.71 & 86.35 \\
Jung \etal~\cite{jung2018water} & 22.77 & 0.88 & 19.13 & 14.30 & 0.84 & 49.91 & 19.86 & 0.92 & 26.76 \\
AEFNet~\cite{fu2021auto} & 23.52 & 0.85 & 19.44 & 19.53 & 0.89 & 27.72 & 24.18 & 0.95 & 16.83 \\
BEDSR-Net~\cite{lin2020bedsr} & 21.51 & 0.85 & 22.58 & 22.31 & 0.75 & 20.86 & 21.50 & 0.90 & 30.52 \\
DHAN~\cite{cun2020towards} & 20.58 & 0.82 & 25.95 & 25.66 & 0.84 & 15.49 & 25.61 & 0.85 & 14.27 \\
LG-ShadowNet~\cite{liu2021shadow} & 19.99 & 0.84 & 27.69 & 26.29 & 0.87 & 14.38 & 24.88 & 0.86 & 16.77 \\
Mask-ShadowGAN~\cite{hu2019mask} & 19.41 & 0.82 & 29.16 & 25.79 & 0.87 & 15.01 & 24.82 & 0.87 & 15.43 \\
SG-ShadowNet~\cite{wan2022style} & 22.90 & 0.86 & 19.20 & 25.34 & 0.91 & 14.96 & 28.22 & 0.96 & 10.61\\
SP+M Net~\cite{Le_2020_ECCV} & 20.04 & 0.84 & 29.13 & 18.85 & 0.86 & 29.57 & 18.84 & 0.91 & 30.94 \\
SP+M+I Net~\cite{le2021physics} & 23.55 & 0.86 & 20.44 & 27.63 & 0.88 & 12.61 & 30.42 & 0.95 & 8.36\\
ST-CGAN~\cite{wang2018stacked} & 13.93 & 0.33 & 52.51 & 12.17 & 0.44 & 63.22 & 12.87 & 0.32 & 61.17\\
ShadowFormer~\cite{guo2023shadowformer} & 20.61 & 0.85 & 26.96 & 17.26 & 0.77 & 36.31 & 23.71 & 0.90 & 17.54 \\
BMNet~\cite{Zhu_2022_CVPR} & 23.12 & 0.80 & 18.19 & 26.15 & 0.84 & 14.17 & 24.86 & 0.80 & 15.59 \\
Ours                     & {\color[HTML]{CB0000} \textbf{24.36}}       & {\color[HTML]{CB0000} \textbf{0.88}}        & {\color[HTML]{CB0000} \textbf{16.04}} & {\color[HTML]{CB0000} \textbf{29.11}}       & {\color[HTML]{CB0000} \textbf{0.94}}        & {\color[HTML]{CB0000} \textbf{10.49}} & {\color[HTML]{CB0000} \textbf{31.83}} & {\color[HTML]{CB0000} \textbf{0.97}} & {\color[HTML]{CB0000} \textbf{7.41}} \\ \bottomrule
\end{tabular}
\vspace{0.5em}
\caption{Quantitative results of the models' comparison on three datasets in low-resolution. The best result is highlighted in red and bold.}
\label{table:low}
\end{table*}

\end{document}